\documentclass[a4paper,fleqn]{cas-sc}

\usepackage[square, comma, sort&compress, numbers]{natbib}

\usepackage[linesnumbered,ruled]{algorithm2e}
\usepackage{rotating}
\usepackage{url}
\def\tsc#1{\csdef{#1}{\textsc{\lowercase{#1}}\xspace}}
\tsc{WGM}
\tsc{QE}
\tsc{EP}
\tsc{PMS}
\tsc{BEC}
\tsc{DE}

\begin{document}
\let\WriteBookmarks\relax
\def\floatpagepagefraction{1}
\def\textpagefraction{.001}

\shorttitle{Efficient Multiplayer Battle Game Optimizer for Adversarial Robust Neural Architecture Search}

\shortauthors{Zhong R. et al.}

\title [mode = title]{Efficient Multiplayer Battle Game Optimizer for Adversarial Robust Neural Architecture Search}            

\author[1]{Rui Zhong}[orcid=0000-0003-4605-5579]
\ead{rui.zhong.u5@elms.hokudai.ac.jp}
\credit{Conceptualization, Methodology, Investigation, Writing – original draft, Writing – review \& editing, and Funding acquisition.}
\affiliation[1]{organization={Graduate School of Information Science and Technology, Hokkaido University}, city={Sapporo}, country={Japan}}

\author[2]{Yuefeng Xu}[]
\ead{xyf20070623@gmail.com}
\credit{Methodology, Formal Analysis, and Writing – review \& editing.}
\affiliation[2]{organization={School of Engineering, University of Fukui}, city={Fukui}, country={Japan}}

\author[3]{Chao Zhang}[orcid=0000-0002-0845-9217]
\ead{zhang@u-fukui.ac.jp}
\credit{Conceptualization and Writing – review \& editing.}
\affiliation[3]{organization={Department of Engineering, University of Fukui}, city={Fukui}, country={Japan}}

\author[4]{Jun Yu}[orcid=0000-0001-5029-0294]
\cormark[1]
\ead{yujun@ie.niigata-u.ac.jp}
\credit{Investigation, Methodology, Formal Analysis, and Writing – review \& editing.}
\affiliation[4]{organization={Institute of Science and Technology, Niigata University}, city={Niigata}, country={Japan}}

\cortext[cor1]{Corresponding author}

\begin{abstract}
This paper introduces a novel metaheuristic algorithm, known as the efficient multiplayer battle game optimizer (EMBGO), specifically designed for addressing complex numerical optimization tasks. The motivation behind this research stems from the need to rectify identified shortcomings in the original MBGO, particularly in search operators during the movement phase, as revealed through ablation experiments. EMBGO mitigates these limitations by integrating the movement and battle phases to simplify the original optimization framework and improve search efficiency. Besides, two efficient search operators: differential mutation and Lévy flight are introduced to increase the diversity of the population. To evaluate the performance of EMBGO comprehensively and fairly, numerical experiments are conducted on benchmark functions such as CEC2017, CEC2020, and CEC2022, as well as engineering problems. Twelve well-established MA approaches serve as competitor algorithms for comparison. Furthermore, we apply the proposed EMBGO to the complex adversarial robust neural architecture search (ARNAS) tasks and explore its robustness and scalability. The experimental results and statistical analyses confirm the efficiency and effectiveness of EMBGO across various optimization tasks. As a potential optimization technique, EMBGO holds promise for diverse applications in real-world problems and deep learning scenarios. The source code of EMBGO is made available in \url{https://github.com/RuiZhong961230/EMBGO}.
\end{abstract}

\begin{highlights}
\item We quantitatively analyze the deficiency of the primary multiplayer battle game optimizer (MBGO) and propose an efficient MBGO (EMBGO) by integrating the movement and the battle phases and introducing novel differential mutation and Lévy flight.
\item Comprehensive numerical experiments on CEC2017, CEC2020, and CEC2022 benchmark functions and eight engineering problems are conducted compared with twelve well-known metaheuristic algorithms to investigate the performance of EMBGO. 
\item We further apply the proposed EMBGO to solve adversarial robust neural architecture search (ARNAS) tasks. The experimental results and statistical analyses confirm the efficiency and effectiveness of EMBGO.
\end{highlights}


\maketitle

\section{Introduction} \label{sec:1}
Metaheuristic algorithms (MA) have become widely embraced in optimization, especially when traditional methods face challenges posed by the complexity, non-differentiation, non-convexity, multi-modality, and nonlinearity of objective functions and constraints \cite{Zhong:23_2}. Functioning as stochastic optimization methodologies, MAs frequently derive inspiration from various sources, including the social behaviors of animals \cite{Sayed:24, Guo:24}, natural phenomena \cite{Zhang:23, Rui:24}, chemistry \cite{Siamak:21, Shehadeh:23}, and the laws of physics \cite{Basset:22, Sujoy:24}. While a comprehensive review of the rich history of the MA community is beyond the scope of this paper, interested readers can delve into \cite{Velasco:23, Rajwar:23, Mohammadi:23} for in-depth exploration. Furthermore, the No Free Lunch Theorem \cite{Wolpert:97} significantly contributes to the thriving of the MA community. Formally, the No Free Lunch Theorem for optimization can be expressed as follows: Any pair of optimization algorithms exhibit identical average performance across all conceivable problems. When an algorithm demonstrates strong performance within a specific problem class, it inherently implies a corresponding weakness when applied to other classes of problems. This trade-off is pivotal for maintaining consistent average performance across all conceivable problem sets. It underscores the substantial reliance of optimization algorithm performance on the distinct characteristics and structure of the problem, emphasizing the imperative for the development of problem-specific optimizers \cite{Zhong:23}.

As one of the latest metaheuristic algorithms, the multiplayer battle game optimizer (MBGO) \cite{Xu:23} draws inspiration from the common patterns across different multiplayer battle royale games. It introduces two distinctive search phases: the movement and the battle phases. In the movement phase, MBGO utilizes the concept of the safe zone to guide individuals toward potential locations (i.e., areas with superior fitness). Subsequently, during the battle phase, a series of search strategies are employed to emulate the behaviors of game players. Empirical results on CEC2017 and CEC2020 benchmark functions, as well as engineering optimization problems, validate the efficiency and scalability of MBGO across various optimization tasks when compared with eight popular MAs.

However, the original MBGO still manifests certain shortcomings, including unbalanced exploitation and exploration, premature convergence, and stagnation when trapped in local optima. To specifically address the issue of unbalanced exploitation and exploration, we conduct pre-experiments to assess the efficiency of search operators in both the movement and battle phases independently. Subsequently, inefficient operators in the movement phase are replaced with more effective ones, such as differential mutation and Lévy flight, with the aim of accelerating the optimization convergence. Our proposed efficient multiplayer battle game optimizer (EMBGO) undergoes comprehensive numerical experiments on CEC2017, CEC2020, and CEC2022 benchmark functions, as well as engineering optimization problems, to thoroughly evaluate its performance. Furthermore, the application in adversarial robust neural architecture search (ARNAS) tasks serves to demonstrate the scalability and robustness of our proposed EMBGO. To summarize, the contributions of this paper can be outlined as follows:

\begin{itemize}
  \item We conduct a numerical analysis to identify the shortcomings related to unbalanced search phases in the original MBGO. To address this deficiency, we integrate the movement and the battle phase and introduce efficient differential mutation and Lévy flight to improve the performance of MBGO.
  \item We systematically and impartially carry out numerical experiments on benchmark functions, including CEC2017, CEC2020, and CEC2022 benchmark functions, along with eight engineering problems. These experiments involve a comparative assessment with twelve well-known MA approaches to evaluate the performance of EMBGO.
  \item We extend our experiments to include ARNAS tasks to investigate the performance of EMBGO in various optimization scenarios. The experimental results and statistical analyses confirm the efficiency and effectiveness of our proposed EMBGO.
\end{itemize}

The remainder of this paper is structured as follows: Section \ref{sec:2} provides an introduction to the original MBGO and ARNAS. In Section \ref{sec:3}, we conduct preliminary pre-experiments on CEC2020 benchmark functions to elucidate the limitations of MBGO. Section \ref{sec:4} details our proposed EMBGO. Numerical experiments and statistical analyses are presented in Section \ref{sec:5}. The performance of EMBGO is further discussed in Section \ref{sec:6}. Finally, Section \ref{sec:7} concludes the paper.

\section{Related works} \label{sec:2}

\subsection{Multiplayer battle game optimizer (MBGO)} \label{sec:2.1}
Inspired by the behaviors and interactions observed in multiplayer battle games, MBGO roughly divides the optimization process into two distinctive phases: the movement and the battle phases. Each of these phases integrates carefully designed search operators that mimic the strategic behaviors of players engaged in a battle game.

\begin{figure}[htbp]
    \centering
    \includegraphics[width=13cm]{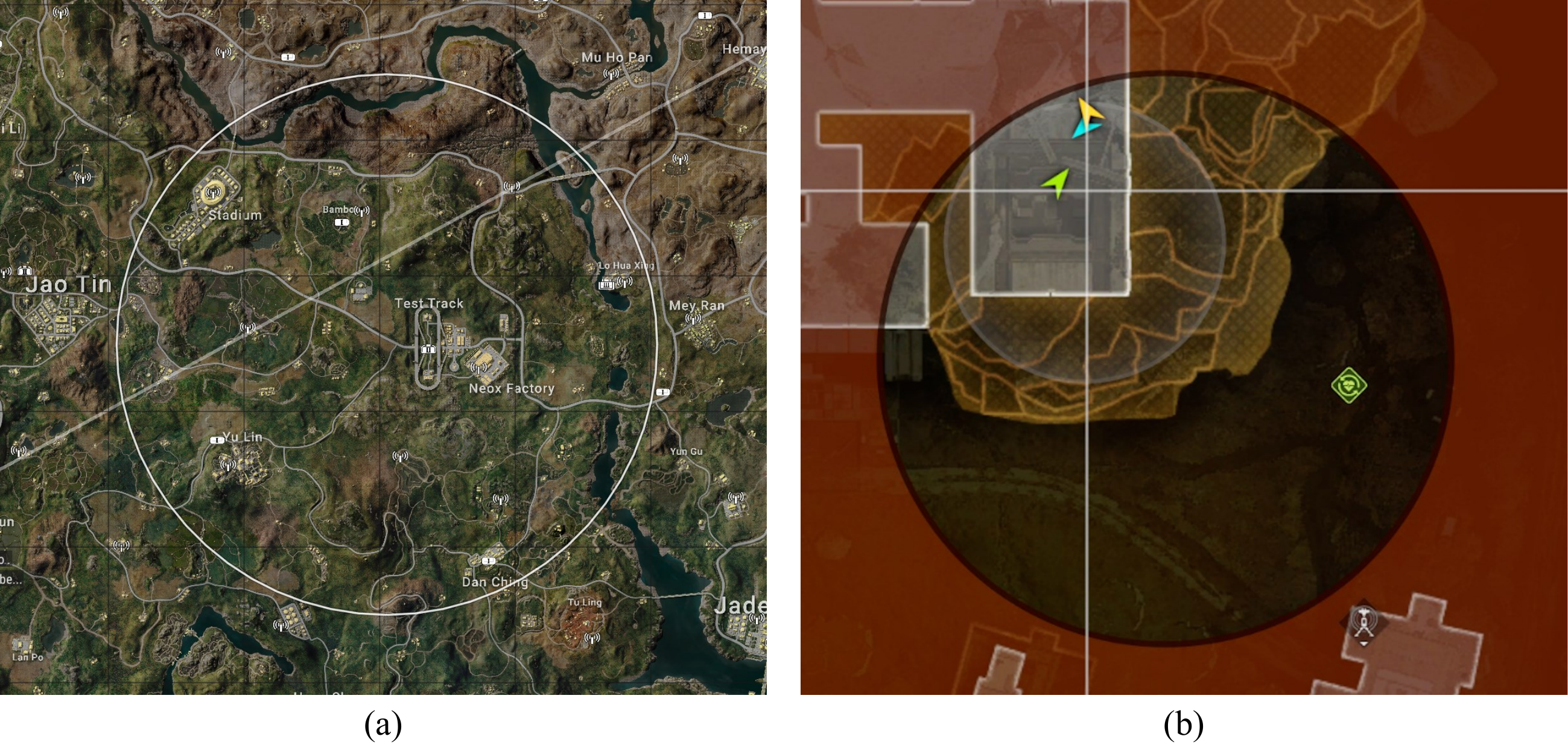}
    \caption{Safe zones from representative multiplayer battle games. (a). The safe zone from PUBG. (b). The safe zone from APEX.}
    \label{fig:2.1.1}
\end{figure}

\textbf{Movement phase}: In the movement phase, players' movements are predominantly influenced by the safe zone, a prevalent game mechanic in many multiplayer battle games. A visualization of this safe zone concept is provided in Figure \ref{fig:2.1.1}.
Players must navigate towards the safe zone to avoid damage from the extreme environments. Motivated by this unique game mechanism, MBGO establishes the current best individual $X_{best}$ as the center of the safe zone, and the Euclidean distance between $X_{best}$ and the current worst individual $X_{worst}$ serves as the baseline radius. To introduce a level of randomness, a random value $\delta \sim U(0.8, 1.2)$ is employed to amplify the radius. Eq. (\ref{eq:2.1.1}) defines the radius as follows.
\begin{equation}
    \label{eq:2.1.1}
    \begin{aligned}
        R = (\Vert X_{best} - X_{worst} + \epsilon\Vert)\cdot \delta
    \end{aligned}
\end{equation}
where $\Vert \cdot \Vert$ represents the Euclidean distance, and $\epsilon$ is a small value to prevent the radius from becoming zero. For individuals located within the safe zone, MBGO utilizes both local information and information from the current best individual $X_{best}$ to construct offspring individual, as expressed in Eq. (\ref{eq:2.1.2}). 
\begin{equation}
    \label{eq:2.1.2}
    \begin{aligned}
        X_{new} = X_i + X_{best} \cdot \sin(2\pi r)
    \end{aligned}
\end{equation}
where $r$ is a random number within the range of (0, 1). For individuals positioned outside the safe zone, Eq. (\ref{eq:2.1.3}) is utilized to speed up individuals' movement towards potential areas.
\begin{equation}
    \label{eq:2.1.3}
    \begin{aligned}
        X^k_{new} = \begin{cases}
		  X^k_{i} + \theta, \ if \  r < 0.5 \\
            X^k_{i} + (X^k_{best} - X^k_{i}) \cdot r, \ otherwise
	    \end{cases}
    \end{aligned}
\end{equation}
where $\theta$ is a random value following the standard normal distribution $N(0, 1)$. More precisely, for each independent dimension $k$, each scheme in Eq. (\ref{eq:2.1.3}) has an equal probability of being chosen for constructing the offspring individual. Importantly, the movement phase includes an inherent greedy selection mechanism. This implies that after constructing a new offspring individual, the selection operator is triggered, permitting the superior offspring individual to survive.

\textbf{Battle phase}: The battle phase emulates diverse player behaviors when encountering random enemies in the game. While player behaviors may exhibit variability during the game, the ultimate objective remains survival until the end of the game and the elimination of enemies to the greatest extent possible. MBGO simplifies real-game observations and imposes idealized constraints. Each individual is assumed to face only one random enemy, with the fitness value representing the player's capacity. Eqs. (\ref{eq:2.1.4}) and (\ref{eq:2.1.5}) provide mathematical models for confronting stronger and weaker enemies, respectively.
\begin{equation}
    \label{eq:2.1.4}
    \begin{aligned}
        X^k_{new} = \begin{cases}
		  X^k_{i} + \mathbf{dir}^k \cdot r, \ if \  r < 0.5 \\
            X^k_{enemy} + \mathbf{dir}^k \cdot r, \ otherwise
	    \end{cases}
    \end{aligned}
\end{equation}
\begin{equation}
    \label{eq:2.1.5}
    \begin{aligned}
        X_{new} = X_{i} + \mathbf{dir} \cdot \cos(2\pi r)
    \end{aligned}
\end{equation}
where $\mathbf{dir}$ represents the differential vector between the $i^{th}$ individual $X_{i}$ and the random selected enemy $X_{enemy}$, as expressed in Eq. (\ref{eq:2.1.6}).
\begin{equation}
    \label{eq:2.1.6}
    \begin{aligned}
        \mathbf{dir} = \begin{cases}
		  X_{i} - X_{enemy} \ if \ X_i \ has \ a \ better \ fitness \ value \\
            X_{enemy} - X_{i}, \ otherwise
	    \end{cases}
    \end{aligned}
\end{equation}

In summary, the pseudocode of MBGO is presented in Algorithm \ref{alg:2.1.1}.
\begin{algorithm}
	\label{alg:2.1.1}
	\DontPrintSemicolon
	\SetAlgoLined
	\KwIn {Population size: $N$, Dimension: $D$, Maximum iteration: $T_{max}$}
	\KwOut {Optimum: $X_{best}$}
	\SetKwFunction{FMBGO}{\textbf{MBGO}}
	\SetKwProg{Fn}{Function}{:}{}
	\Fn{\FMBGO{$N, D, T_{max}$}}{
		Initialize the population randomly \;
        $X_{best} \gets \textbf{best}(R)$ \;
        $t \gets 0$ \;
        \While {$t < T_{max}$} {
            $\bullet$ Movement phase \;
            \For {$i=0 \ to \ N$} {
                Determine the safe zone using Eq. (\ref{eq:2.1.1}) \;
                \If {$X_i$ is within the safe zone} {
                    Construct $X_{new}$ using Eq. (\ref{eq:2.1.2}) \;
                } \Else {
                    Construct $X_{new}$ using Eq. (\ref{eq:2.1.3}) \;
                }
                Embedded greedy selection \;
            }
            $\bullet$ Battle phase \;
            \For {$i=0 \ to \ N$} {
                Select a random enemy $X_{enemy}$ for $X_{i}$ \;
                \If {$X_{enemy}$ has a better fitness value} {
                    Construct $X_{new}$ using Eq. (\ref{eq:2.1.4}) \;
                } \Else {
                    Construct $X_{new}$ using Eq. (\ref{eq:2.1.5}) \;
                }
                Embedded greedy selection \;
            }
            $X_{best} \gets \textbf{best}(R)$ \;
            $t \gets t+1$ \;
        }
        \textbf{return} $X_{best}$ \;
        }
\caption{MBGO \cite{Xu:23}}
\end{algorithm}

\subsection{Adversarial robust neural architecture search (ARNAS)}  \label{sec:2.2}
Neural Architecture Search (NAS) is a technique within the realm of deep learning that automates the process of designing and selecting optimal neural network architectures for a given task. Traditional methods of designing neural networks often involve human experts manually crafting architectures based on their domain knowledge and intuition. However, with the increasing complexity of neural networks, designing effective architectures for every task manually has become increasingly challenging. The objective of NAS is to explore the extensive design space of neural networks and discover architectures that excel in specific tasks, such as image classification or natural language processing. 

Since the pioneering work that introduced reinforcement learning to NAS by Zoph and Le \cite{Zoph:17}, interest and research in NAS have rapidly expanded. Representative techniques encompass a variety of approaches, including random search \cite{Tom:19, Li:20, Xie:20}, reinforcement learning \cite{Hieu:18, Ramakanth:19, Cai:19, Gabriel:20, Guo:20}, evolutionary approaches \cite{Han:20, Chu:21, Chen:21, Xia:21, Tong:22}, and gradient predictor-based approaches \cite{Dong:19, Yao:19, Liu:19, Jiang:19, Wang:23}. For more comprehensive surveys on this topic, please refer to \cite{Ren:21, Matt:24}.

ARNAS aims to uncover high-quality neural network architectures capable of maintaining robust performance when exposed to adversarial data samples. Neural network models are often vulnerable to perturbations in input data, resulting in erroneous decisions with high confidence and hindering their practical deployment \cite{Ramtin:20}. Similar to the search process without adversarial data samples (i.e., the clean case), identifying architectures robust against adversarial perturbations typically involves a laborious trial-and-error process. Moreover, evaluating a network's robustness is significantly more resource-intensive than evaluating its clean accuracy, adding an additional layer of complexity to the optimization process. This challenge has garnered substantial attention in recent years: Chaitanya et al. \cite{Chaitanya:21} conducted a thorough investigation into the ARNAS task, addressing adversarial robustness through complex topology analysis without adversarial training. In the context of small-scale attacks, architectures generated through NAS demonstrate greater robustness when applied to small datasets and simple tasks compared to manually crafted architectures. However, as the dataset size or task complexity increases, expert-designed architectures tend to exhibit more robust performance than their NAS-based counterparts. Xie et al. \cite{Xie:23} proposed G-NAS, a novel robust neural architecture search framework tailored for graph neural networks (GNNs). This approach introduces graph structure mask operations into the search space, creating a robust search space for the message-passing mechanism in GNNs. The space encompasses various defensive operation candidates, enabling the G-NAS framework to search for GNNs with enhanced robustness. Additionally, a robustness metric is defined to guide the search process, facilitating the identification of robust architectures. G-RNA provides insights into GNN robustness from an architectural perspective and efficiently searches for optimal adversarially robust GNNs. Jung et al. \cite{Jung:23} designed a standard ARNAS benchmark based on NAS-Bench-201, a well-established and well-studied NAS benchmark suite in image classification. The benchmark incorporates four representative adversarial attack methods: the fast gradient sign method (FGSM), projected gradient descent (PGD), adaptive PGD (APGD), and Square Attack. Robustness measurements, based on Jacobian and Hessian matrices, were conducted to investigate the robustness and predictability of the benchmark.

\section{Numerical analysis of EMBGO on CEC2017 benchmark functions} \label{sec:3}
The objective of this pre-experiment is to evaluate the efficiency of individual components within MBGO. In this context, we systematically analyze the movement and battle phases, conducting optimization experiments on 30-D CEC2017 benchmark functions with 30 trial runs. The population size and maximum fitness evaluations (FEs) are set to 100 and 30,000, respectively. The parameters in MBGO are held constant by default. To monitor the convergence status, we employ the performance indicator of population diversity \cite{Dirk:20}, defined in Eq. (\ref{eq:3.1}).
\begin{equation}
    \label{eq:3.1}
    \begin{aligned}
        PD = \frac{1}{N\times D}\sum^N_{i=1} \sum^D_{j=1} \frac{\vert X_{ij} - X_{mean,j} \vert}{UB_j - LB_j}
    \end{aligned}
\end{equation}
where $N$ represents the population size, and $D$ corresponds to the dimension size. The symbol $X_{mean}$ designates the centroid of the population, while $LB$ and $UB$ denote the lower and upper bounds of the search space, respectively. This metric serves to characterize the distribution of the population, providing insights into the evolutionary status during the optimization process. The Holm multiple comparison test \cite{Holm:79} is employed to determine the significance between every pair of compared algorithms, and the fitness value of the best-performing algorithm is highlighted in bold.

\begin{figure}[htbp]
    \centering
    \includegraphics[width=16cm]{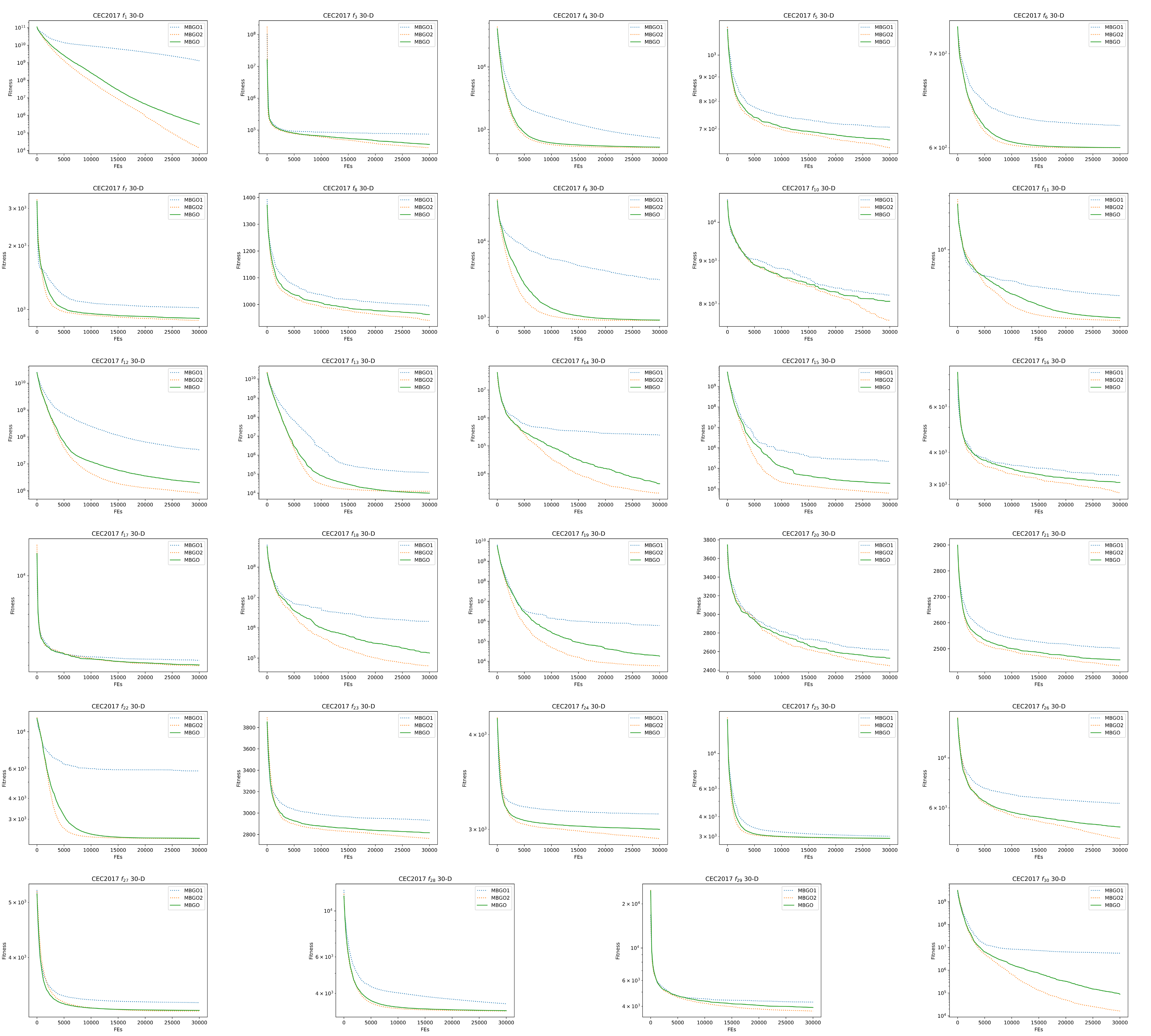}
    \caption{Convergence curves of the fitness value on 30-D CEC2017 benchmark functions.}
    \label{fig:3.1}
\end{figure}

\begin{figure}[htbp]
    \centering
    \includegraphics[width=16cm]{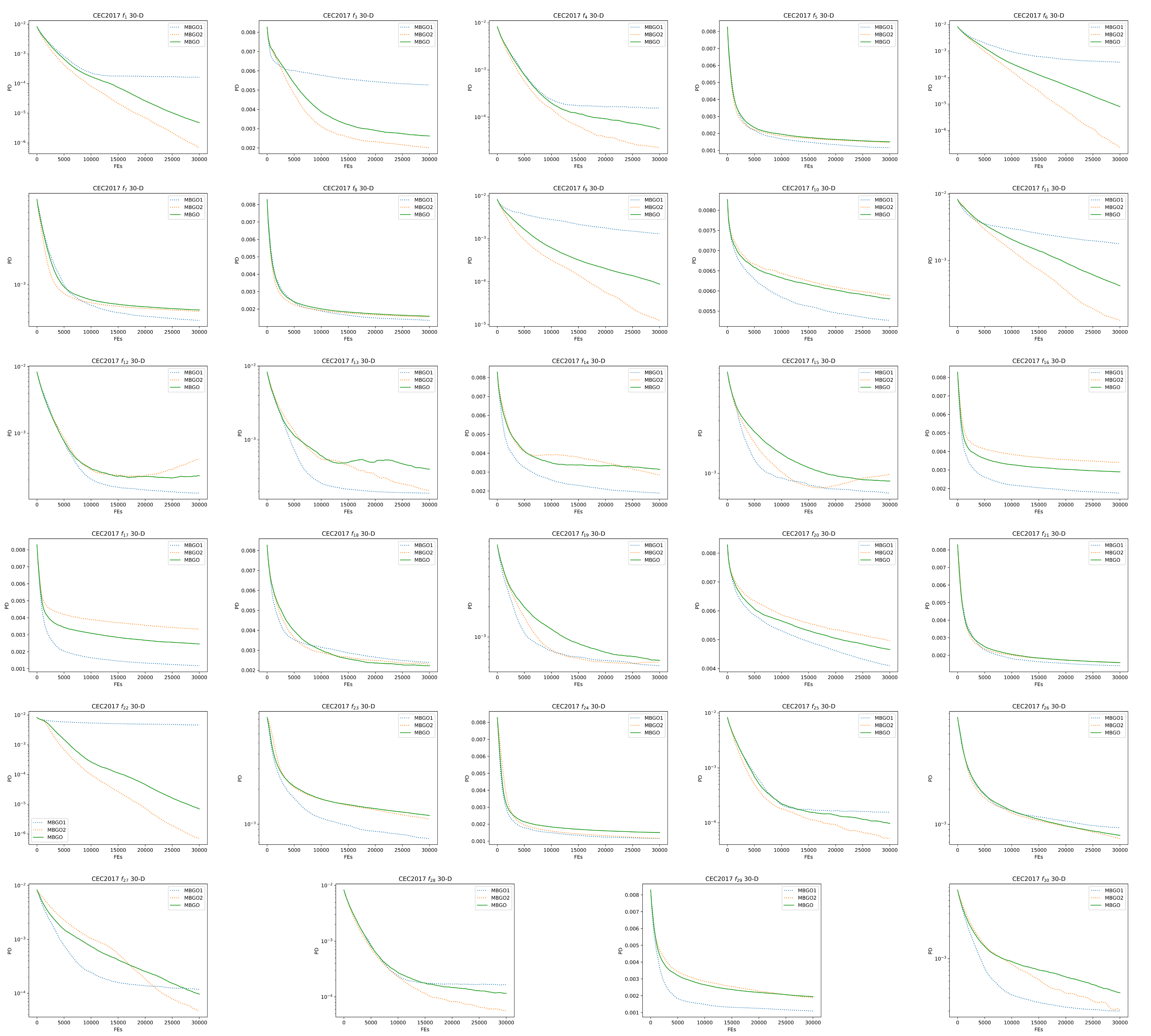}
    \caption{Convergence curves the population diversity on 30-D CEC2017 benchmark functions.}
    \label{fig:3.2}
\end{figure}

The experimental results and statistical analyses are summarized in Table \ref{tbl:3.1}. Convergence curves of the fitness value and population diversity are depicted in Figures \ref{fig:3.1} and \ref{fig:3.2}.
\begin{table}[htbp]
	\scriptsize
	\centering
	\renewcommand\arraystretch{1.5}
	\caption{The experimental results and statistical analyses on 30-D CEC2017 benchmark functions. MBGO1: MBGO with only movement phase; MBGO2: MBGO with only battle phase. Notably, $f_2$ is deleted in CEC2017 benchmark functions and $f_{30}$ is adopted as compensation.}
	\label{tbl:3.1}
		\begin{tabular}{cccccccccccc}
			\toprule
		      \multirow{2}{*}{Func} & \multicolumn{3}{c}{MBGO1} & \multicolumn{3}{c}{MBGO2} & \multicolumn{3}{c}{MBGO} \\
		      \cmidrule(r){2-4} \cmidrule(r){5-7} \cmidrule(r){8-10} 
		      & mean & std & PD & mean & std & PD & mean & std & PD \\
			\midrule
			$f_1$ & 1.307e+09 $+$ & 1.285e+09 & 5.404e-06 & \textbf{1.440e+04} $-$ & 1.801e+04 & 2.280e-08 & 3.143e+05 & 2.102e+05 & 1.597e-07 \\
			$f_3$ & 7.484e+04 $+$ & 1.281e+04 & 1.755e-04 & \textbf{2.806e+04} $-$ & 5.298e+03 & 6.702e-05 & 3.557e+04 & 7.999e+03 & 8.725e-05 \\
			$f_4$ & 7.312e+02 $+$ & 1.545e+02 & 5.216e-06 & \textbf{5.130e+02} $-$ & 1.426e+01 & 7.516e-07 & 5.236e+02 & 2.592e+01 & 1.890e-06 \\
			$f_5$ & 7.061e+02 $+$ & 2.271e+01 & 3.865e-05 & \textbf{6.396e+02} $\approx$ & 2.501e+01 & 4.899e-05 & 6.641e+02 & 1.390e+01 & 5.024e-05 \\
			$f_6$ & 6.222e+02 $+$ & 6.503e+00 & 1.265e-05 & \textbf{6.000e+02} $-$ & 6.600e-02 & 7.581e-09 & 6.001e+02 & 1.176e-01 & 2.669e-07 \\
			$f_7$ & 1.020e+03 $+$ & 3.137e+01 & 1.376e-05 & \textbf{8.889e+02} $-$ & 1.402e+01 & 1.728e-05 & 9.084e+02 & 1.520e+01 & 1.783e-05 \\
			$f_8$ & 9.948e+02 $+$ & 2.097e+01 & 4.495e-05 & \textbf{9.406e+02} $-$ & 1.512e+01 & 5.170e-05 & 9.625e+02 & 1.466e+01 & 5.297e-05 \\
			$f_9$ & 3.124e+03 $+$ & 8.847e+02 & 4.361e-05 & \textbf{9.098e+02} $\approx$ & 7.915e+00 & 4.181e-07 & 9.187e+02 & 2.133e+01 & 2.933e-06 \\
			$f_{10}$ & 8.190e+03 $+$ & 3.571e+02 & 1.757e-04 & \textbf{7.645e+03} $-$ & 3.475e+02 & 1.961e-04 & 8.055e+03 & 2.678e+02 & 1.934e-04 \\
			$f_{11}$ & 2.533e+03 $+$ & 6.505e+02 & 5.947e-05 & \textbf{1.218e+03} $-$ & 2.974e+01 & 4.200e-06 & 1.315e+03 & 4.247e+01 & 1.392e-05 \\
			$f_{12}$ & 3.344e+07 $+$ & 2.713e+07 & 4.224e-06 & \textbf{8.197e+05} $-$ & 4.619e+05 & 1.377e-05 & 1.995e+06 & 1.371e+06 & 7.736e-06 \\
			$f_{13}$ & 1.190e+05 $+$ & 5.213e+04 & 6.330e-06 & 1.259e+04 $\approx$ & 6.274e+03 & 6.889e-06 & \textbf{1.008e+04} & 8.801e+03 & 1.335e-05 \\
			$f_{14}$ & 2.424e+05 $+$ & 2.860e+05 & 6.302e-05 & \textbf{2.035e+03} $-$ & 3.436e+02 & 9.493e-05 & 4.411e+03 & 1.749e+03 & 1.048e-04 \\
			$f_{15}$ & 2.193e+05 $+$ & 6.165e+05 & 2.208e-05 & \textbf{6.145e+03} $-$ & 2.830e+03 & 3.272e-05 & 1.827e+04 & 1.258e+04 & 2.843e-05 \\
			$f_{16}$ & 3.247e+03 $+$ & 2.368e+02 & 5.840e-05 & \textbf{2.779e+03} $-$ & 2.152e+02 & 1.134e-04 & 3.056e+03 & 1.724e+02 & 9.672e-05 \\
			$f_{17}$ & 2.189e+03 $+$ & 1.650e+02 & 3.916e-05 & \textbf{1.982e+03} $\approx$ & 1.050e+02 & 1.112e-04 & 2.012e+03 & 1.175e+02 & 8.191e-05 \\
			$f_{18}$ & 1.628e+06 $+$ & 1.619e+06 & 7.981e-05 & \textbf{5.570e+04} $-$ & 1.968e+04 & 7.674e-05 & 1.472e+05 & 1.025e+05 & 7.412e-05 \\
			$f_{19}$ & 6.148e+05 $+$ & 8.825e+05 & 1.724e-05 & \textbf{5.940e+03} $-$ & 2.695e+03 & 1.884e-05 & 1.812e+04 & 1.415e+04 & 1.945e-05 \\
			$f_{20}$ & 2.617e+03 $+$ & 1.206e+02 & 1.365e-04 & \textbf{2.447e+03} $-$ & 9.323e+01 & 1.653e-04 & 2.529e+03 & 1.062e+02 & 1.552e-04 \\
			$f_{21}$ & 2.502e+03 $+$ & 3.564e+01 & 4.676e-05 & \textbf{2.434e+03} $-$ & 2.149e+01 & 5.296e-05 & 2.457e+03 & 1.535e+01 & 5.275e-05 \\
			$f_{22}$ & 5.835e+03 $+$ & 6.606e+02 & 1.551e-04 & \textbf{2.301e+03} $-$ & 1.329e+00 & 2.353e-08 & 2.307e+03 & 2.504e+00 & 2.334e-07 \\
			$f_{23}$ & 2.933e+03 $+$ & 5.326e+01 & 2.512e-05 & \textbf{2.762e+03} $-$ & 3.315e+01 & 3.697e-05 & 2.816e+03 & 2.290e+01 & 3.958e-05 \\
			$f_{24}$ & 3.143e+03 $+$ & 6.500e+01 & 3.852e-05 & \textbf{2.918e+03} $-$ & 3.907e+01 & 3.891e-05 & 3.000e+03 & 2.020e+01 & 5.036e-05 \\
			$f_{25}$ & 3.012e+03 $+$ & 4.311e+01 & 5.221e-06 & \textbf{2.913e+03} $\approx$ & 1.866e+01 & 1.726e-06 & 2.921e+03 & 2.447e+01 & 3.244e-06 \\
			$f_{26}$ & 6.281e+03 $+$ & 7.724e+02 & 3.110e-05 & \textbf{4.384e+03} $-$ & 5.965e+02 & 2.521e-05 & 4.932e+03 & 8.009e+02 & 2.679e-05 \\
			$f_{27}$ & 3.334e+03 $+$ & 5.584e+01 & 3.924e-06 & \textbf{3.222e+03} $-$ & 8.999e+00 & 1.566e-06 & 3.231e+03 & 1.292e+01 & 3.219e-06 \\
			$f_{28}$ & 3.565e+03 $+$ & 2.107e+02 & 5.584e-06 & \textbf{3.290e+03} $\approx$ & 2.093e+01 & 1.862e-06 & 3.302e+03 & 2.243e+01 & 3.857e-06 \\
			$f_{29}$ & 4.275e+03 $+$ & 2.202e+02 & 3.641e-05 & \textbf{3.715e+03} $-$ & 1.171e+02 & 6.289e-05 & 3.929e+03 & 1.641e+02 & 6.484e-05 \\
			$f_{30}$ & 5.578e+06 $+$ & 6.638e+06 & 6.684e-06 & \textbf{1.632e+04} $-$ & 1.142e+04 & 6.808e-06 & 8.865e+04 & 1.091e+05 & 1.169e-05 \\
			\midrule
			$+$/$\approx$/$-$: & \multicolumn{3}{c}{29/0/0} & \multicolumn{3}{c}{0/6/23} & \multicolumn{3}{c}{-} \\
			\midrule
			avg. rank & \multicolumn{3}{c}{3.0} & \multicolumn{3}{c}{1.03} & \multicolumn{3}{c}{1.97} \\
			\bottomrule
		\end{tabular}
\end{table}

Based on the pre-experimental results, the effectiveness and superiority of search operators in the battle phase are evident. Conversely, shortcomings in the search operators of the movement phase are apparent, highlighting the necessity for improvement. This observation is further supported by the convergence curves of population diversity in Figure \ref{fig:3.2}. In simpler problems like the unimodal function $f_1$, MBGO1 consistently maintains high population diversity throughout the optimization process, contrary to expectations of rapid convergence and high-quality exploitative search. However, for more complex problems such as hybrid functions $f_{10}$, $f_{12}$, $f_{14}$, and $f_{16}$, as well as composite functions $f_{20}$, $f_{21}$, $f_{23}$, $f_{29}$, and $f_{30}$, the population diversity of MBGO1 experiences swift degeneration as optimization progresses. This is contrary to the expectations that the optimizer, in these cases, exhibits powerful explorative search capabilities and the ability to escape local optima. Consequently, there is an urgent need to develop more efficient and high-performing search operators for the movement phase.

\section{Our proposal: EMBGO} \label{sec:4}
\begin{figure}[htbp]
    \centering
    \includegraphics[width=12cm]{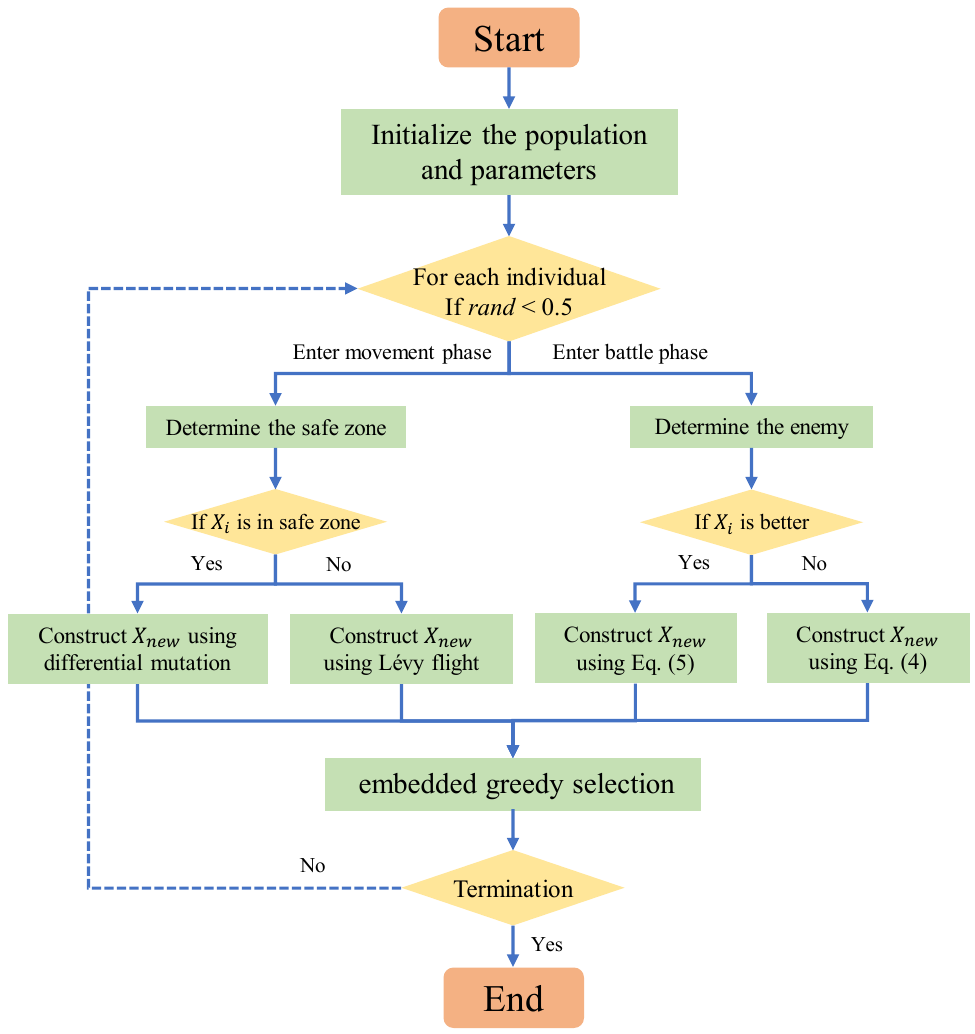}
    \caption{The flowchart of our proposed EMBGO.}
    \label{fig:4.1}
\end{figure}

This section provides a comprehensive introduction to our proposed EMBGO in detail, with the corresponding flowchart is first presented in Figure \ref{fig:4.1}.
The initial modification in EMBGO entails the removal of the strict separation between the movement and battle phases. According to the flowchart, each individual now has an equal probability of entering either the movement or the battle phase. This adjustment is designed to offer individuals in a single iteration a higher probability of adopting different search operators. It is anticipated that this correction will contribute to sustaining population diversity during optimization and alleviating premature convergence.

Leveraging the superior performance observed in the battle phase of the original MBGO, EMBGO integrates high-quality search operators from this phase. The key emphasis of this modification is on improving the search operators in the movement phase. Therefore, two operators are employed: the differential mutation and the Lévy flight operator.

\textbf{Differential mutation}: The differential mutation, derived from differential evolution (DE), is a technique widely applied to simulate animal foraging \cite{Min:20, Zhao:24, Niu:24} and human cooperation behaviors \cite{Faridmehr:23, Lian:24}. In this context, we introduce a novel current-to-best\&mean differential mutation strategy, as formulated in Eq. (\ref{eq:4.1}).
\begin{equation}
    \label{eq:4.1}
    \begin{aligned}
        X_{new} = X_i + (X_{best} - X_i) \cdot \sin(2\pi r) + (X_{mean} - X_i) \cdot \sin(2\pi r) 
    \end{aligned}
\end{equation}

\begin{figure}[htbp]
    \centering
    \includegraphics[width=11cm]{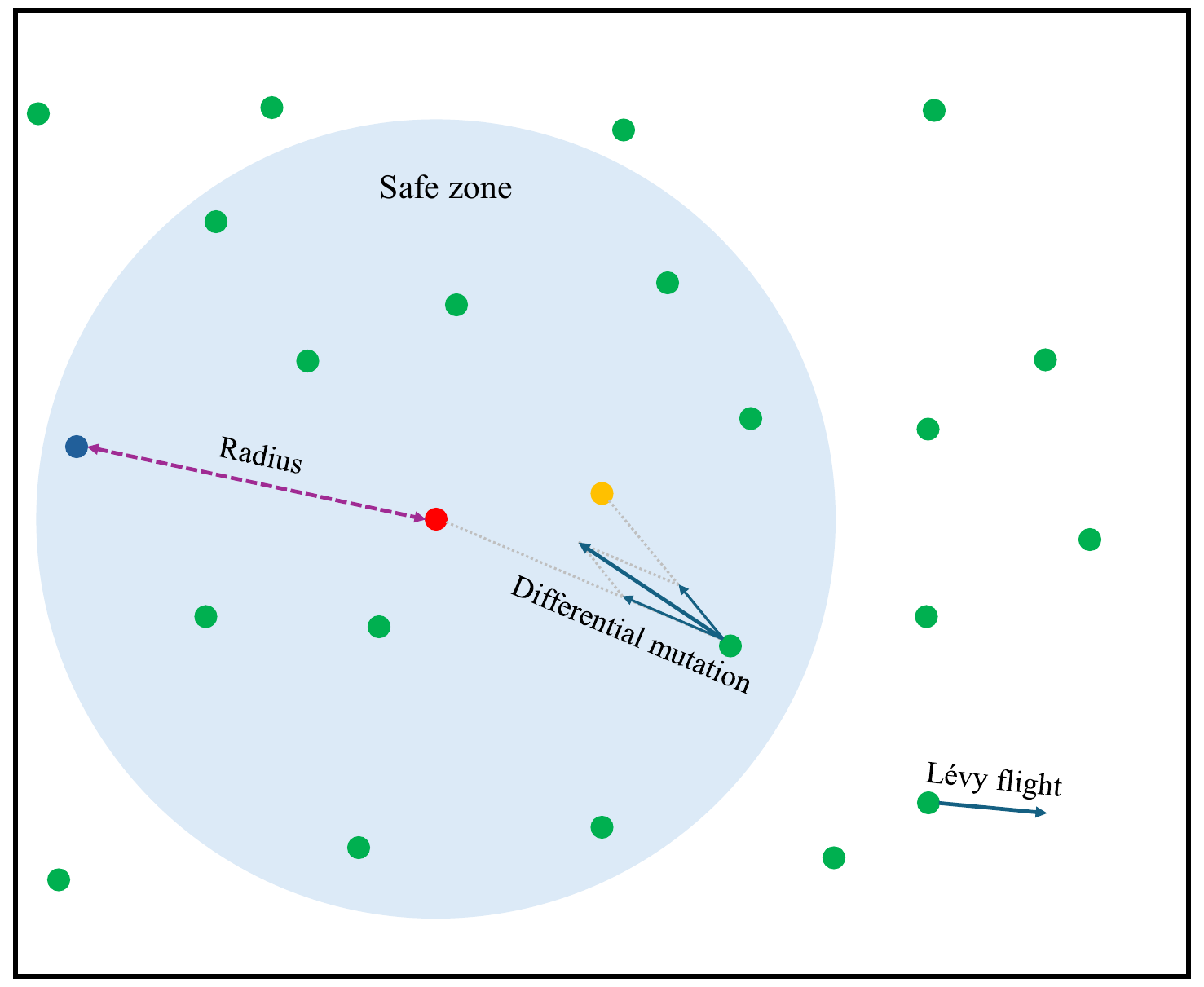}
    \caption{A visual demonstration of the differential mutation and Lévy flight in a 2-D search space is provided. The worst, best, and centroid individuals of the population are represented by blue, red, and orange dots, respectively. The light blue circle depicts the safe zone. Individuals located within the safe zone employ the differential mutation, while those outside the safe zone utilize the Lévy flight to construct offspring individuals.}
    \label{fig:4.2}
\end{figure}

The proposed mutation strategy involves two differential vectors: current to best and current to mean. A visual representation is provided in Figure \ref{fig:4.2}. Individuals positioned within the safe zone are in close proximity to the current optimum. According to the proximate optimality principle \cite{Yaguchi:11}, optimal solutions in a given optimization problem remain approximately optimal when subjected to small perturbations or changes in the problem formulation. In other words, the principle suggests that high-quality solutions share similar structures, making the exploitative search within the safe zone promising. The movement of the current individual $X_i$ toward both the current optimum $X_{best}$ and the centroid of the population $X_{mean}$ is driven by two distinct differential vectors, expediting optimization convergence. The inclusion of a random coefficient $\sin(2\pi r)$ contributes to maintaining population diversity.

\textbf{Lévy flight}: The Lévy flight is a form of random walk characterized by occasional long jumps, inspired by the Lévy distribution. Subsequently, the Lévy flight, serving as an effective explorative search operator, has been integrated into numerous MA approaches to enhance their performance \cite{Wu:23, Saravanan:23, Chang:23, Syama:23}. Eq. (\ref{eq:4.2}) presents the incorporation with the Lévy flight operator.
\begin{equation}
    \label{eq:4.2}
    \begin{aligned}
        X_{new} & = X_i + L\acute{e}vy()\\
        L\acute{e}vy() & \sim \frac{u}{\vert v\vert ^{\frac{1}{\beta}}} 
    \end{aligned}
\end{equation}
where $\beta$ is the Lévy distribution index bounded as $0<\beta \leq 2$, while $u$ and $v$ are defined in Eq. (\ref{eq:4.3}).
\begin{equation}
    \label{eq:4.3}
    \begin{aligned}
		u \sim N(0, \sigma^2), u \sim N(0, 1) 
    \end{aligned}
\end{equation}
and the standard deviation $\sigma$ is defined using Eq. (\ref{eq:4.4}).
\begin{equation}
    \label{eq:4.4}
    \begin{aligned}
        \sigma = \{\frac{\Gamma(1+\beta) \sin(\frac{\pi\beta}{2})}{\beta\Gamma(\frac{1+\beta}{2})2^{\frac{\beta-1}{2}}}\}^{\frac{1}{\beta}}
    \end{aligned}
\end{equation}
the gamma function $\Gamma$ for an integer $z$ can be formulated as Eq. (\ref{eq:4.5}).
\begin{equation}
    \label{eq:4.5}
    \begin{aligned}
        \Gamma(z)=\int^{+\infty}_0t^{z-1}e^{-t}dt
    \end{aligned}
\end{equation}

The extensive exploration of a larger search domain facilitated by the proposed EMBGO algorithm, through its incorporation of long-range jumps, proves particularly effective for global optimization problems that demand broad exploration to find the global optimum. In contrast to traditional random walks, which may become entrenched in local solutions, the occasional long jumps in Lévy flights enhance the algorithm's ability to escape from local optima and explore new regions. Moreover, Lévy flights exhibit a memoryless property, indicating that the length of the next jump is not influenced by the history of previous steps. This property can be advantageous in certain optimization scenarios where adaptive exploration is required without bias from past search history.

In summary, the pseudocode of EMBGO is presented in Algorithm \ref{alg:4.1}.
\begin{algorithm}
	\label{alg:4.1}
	\DontPrintSemicolon
	\SetAlgoLined
	\KwIn {Population size: $N$, Dimension: $D$, Maximum iteration: $T_{max}$}
	\KwOut {Optimum: $X_{best}$}
	\SetKwFunction{FEMBGO}{\textbf{EMBGO}}
	\SetKwProg{Fn}{Function}{:}{}
	\Fn{\FEMBGO{$N, D, T_{max}$}}{
		Initialize the population randomly \;
        $X_{best} \gets \textbf{best}(R)$ \;
        $t \gets 0$ \;
        \While {$t < T_{max}$} {
            \For {$i=0 \ to \ N$} {
                \If {rand() < 0.5} {
                    Determine the safe zone using Eq. (\ref{eq:2.1.1}) \;
                    \If {$X_i$ is within the safe zone} {
                        Construct $X_{new}$ with differential mutation using Eq. (\ref{eq:4.1}) \;
                    } \Else {
                        Construct $X_{new}$ with Lévy flight using Eq. (\ref{eq:4.2}) \;
                    }
                } \Else{
                    Select a random enemy $X_{enemy}$ for $X_{i}$ \;
                    \If {$X_{enemy}$ has a better fitness value} {
                        Construct $X_{new}$ using Eq. (\ref{eq:2.1.4}) \;
                    } \Else {
                        Construct $X_{new}$ using Eq. (\ref{eq:2.1.5}) \;
                    }
                }
                Embedded greedy selection \;
            }
            $X_{best} \gets \textbf{best}(R)$ \;
            $t \gets t+1$ \;
        }
        \textbf{return} $X_{best}$ \;
        }
\caption{EMBGO}
\end{algorithm}

\section{Numerical experiments} \label{sec:5}
This section offers a detailed overview of the experiment settings and the corresponding results. Section \ref{sec:5.1} provides detailed insights into the experimental environments. Subsequently, Sections \ref{sec:5.2}, \ref{sec:5.3}, and \ref{sec:5.4} present the experimental results and conduct statistical analyses on the benchmark functions.

\subsection{Experimental settings} \label{sec:5.1}
The competitor MA approaches are implemented using Python 3.11 and executed on the Lenovo Legion R9000P, running on Windows 11. The system's hardware configuration comprises an AMD Ryzen 7 5800H processor with Radeon Graphics clocked at 3.20 GHz and 16GB RAM. With the exception of MBGO and EMBGO, all MA approaches are made available through the MEALPY library \cite{Nguyen:23}. The CEC2017, CEC2020, and CEC2022 benchmark functions used for evaluation are sourced from the OpFuNu library \cite{Thieu:20}. eight engineering optimization problems \cite{Zhong:24} are obtained from the ENOPPY library \cite{Thieu:23}, and the ARNAS benchmark suite, developed based on the NAS-Bench-201, is provided in \cite{Jung:23}.

\subsection{Experiments on CEC benchmark functions} \label{sec:5.2}
This section introduces the numerical experiments conducted on CEC benchmark functions. In the following context, we sequentially present the information on CEC benchmarks, competitor algorithms and parameters, and experimental results obtained on CEC benchmarks.

\textbf{Information on CEC benchmark functions}: The details of of CEC2017, CEC2020, and CEC2022 benchmark functions are summarized in Tables \ref{tbl:5.2.1}, \ref{tbl:5.2.2}, and \ref{tbl:5.2.3}, respectively.
\begin{table}[htbp]
	\scriptsize
	\centering
	\renewcommand\arraystretch{1.3}
	\caption{Summary of the CEC2017 benchmark functions: Uni.=Unimodal function, Multi.=Simple multimodal function, Hybrid.=Hybrid function, Comp.=Composition function}
	\label{tbl:5.2.1}
	\begin{tabular}{cccc}
		\toprule
		No. & Func. & Feature & Optimum  \\
		\midrule
            $f_1$ & Shifted and Rotated Bent Cigar function  & Uni. & 100 \\
            \midrule
            $f_3$ & Shifted and Rotated Rosenbrock’s function & \multirow{7}{*}{Multi.} & 300 \\
            $f_4$ & Shifted and Rotated Rastrigin’s function & ~ & 400 \\
            $f_5$ & Shifted and Rotated Expanded Scaffer’s F6 function & ~ & 500 \\
            $f_6$ & Shifted and Rotated Lunacek Bi\_Rastrigin function & ~ & 600 \\
            $f_7$ & Shifted and Rotated Non-Continuous Rastrigin’s function & ~ & 700 \\
            $f_8$ & Shifted and Rotated Levy function & ~ & 800 \\
            $f_9$ & Shifted and Rotated Schwefel’s function & ~ & 900 \\
            \midrule
            $f_{10}$ & Hybrid function 1 (N = 3) & \multirow{10}{*}{Hybrid.} & 1000 \\
            $f_{11}$ & Hybrid function 2 (N = 3) & ~ & 1100 \\
            $f_{12}$ & Hybrid function 3 (N = 3) & ~ & 1200 \\
            $f_{13}$ & Hybrid function 4 (N = 4) & ~ & 1300 \\
            $f_{14}$ & Hybrid function 5 (N = 4) & ~ & 1400 \\
            $f_{15}$ & Hybrid function 6 (N = 4) & ~ & 1500 \\
            $f_{16}$ & Hybrid function 6 (N = 5) & ~ & 1600 \\
            $f_{17}$ & Hybrid function 6 (N = 5) & ~ & 1700 \\
            $f_{18}$ & Hybrid function 6 (N = 5) & ~ & 1800 \\
            $f_{19}$ & Hybrid function (N = 6) & ~ & 1900 \\
            \midrule
            $f_{20}$ & Composition function 1 (N = 3) & \multirow{11}{*}{Comp.} & 2000 \\
            $f_{21}$ & Composition function 2 (N = 3) & ~ & 2100 \\
            $f_{22}$ & Composition function 3 (N = 4) & ~ & 2200 \\
            $f_{23}$ & Composition function 4 (N = 4) & ~ & 2300 \\
            $f_{24}$ & Composition function 5 (N = 5) & ~ & 2400 \\
            $f_{25}$ & Composition function 6 (N = 5) & ~ & 2500 \\
            $f_{26}$ & Composition function 7 (N = 6) & ~ & 2600 \\
            $f_{27}$ & Composition function 8 (N = 6) & ~ & 2700 \\
            $f_{28}$ & Composition function 9 (N = 3) & ~ & 2800 \\
            $f_{29}$ & Composition function 10 (N = 3) & ~ & 2900 \\
            $f_{30}$ & Composition function 11 (N = 3) & ~ & 3000 \\
        \midrule
            \multicolumn{4}{c}{Search range: [-100, 100]$^D$} \\
		\bottomrule
	\end{tabular}
\end{table}
\begin{table}[htbp]
	\scriptsize
	\centering
	\renewcommand\arraystretch{1.3}
	\caption{Summary of the CEC2020 benchmark functions: Uni.=Unimodal function, Multi.=Multimodal function, Hybrid.=Hybrid function, Comp.=Composition function}
	\label{tbl:5.2.2}
	\begin{tabular}{cccc}
		\toprule
		No. & Func. & Feature & Optimum  \\
		\midrule
            $f_1$ & Shifted and Rotated Bent Cigar Function & Uni. & 100 \\
            \midrule
            $f_2$ & Shifted and Rotated Schwefel’s function & \multirow{3}{*}{Multi.} & 1100 \\
            $f_3$ & Shifted and Rotated Lunacek bi-Rastrigin function & ~ & 700 \\
            $f_4$ & Expanded Rosenbrock’s plus Griewangk’s function & ~ & 1900 \\
            \midrule
            $f_5$ & Hybrid function 1 (N = 3) & \multirow{3}{*}{Hybrid.} & 1700 \\
            $f_6$ & Hybrid function 2 (N = 4) & ~ & 1600 \\
            $f_7$ & Hybrid function 3 (N = 5) & ~ & 2100 \\
            \midrule
            $f_8$ &  Composition function 1 (N = 3) & \multirow{3}{*}{Comp.}  & 2200 \\
            $f_9$ &  Composition function 2 (N = 4) & ~ & 2400 \\
            $f_{10}$ &  Composition function 3 (N = 5) & ~ & 2500 \\
        \midrule
            \multicolumn{4}{c}{Search range: [-100, 100]$^D$} \\
		\bottomrule
	\end{tabular}
\end{table}
\begin{table}[htbp]
	\scriptsize
	\centering
	\renewcommand\arraystretch{1.3}
	\caption{Summary of the CEC2022 benchmark functions: Uni.=Unimodal function, Basic.=Basic function, Hybrid.=Hybrid function, Comp.=Composition function}
	\label{tbl:5.2.3}
	\begin{tabular}{cccc}
		\toprule
		Func. & Description & Feature & Optimum  \\
		\midrule
            $f_1$ & Shifted and full Rotated Zakharov & Uni. & 300 \\
            \midrule
            $f_2$ & Shifted and full Rotated Rosenbrock & \multirow{4}{*}{Basic.} & 400 \\
            $f_3$ & Shifted and full Rotated Expanded Schaffer $f_6$ & ~ & 600 \\
            $f_4$ & Shifted and full Rotated Non-Continuous Rastrigin & ~ & 800 \\
            $f_5$ & Shifted and full Rotated Levy & ~ & 900 \\
            \midrule
            $f_6$ & Hybrid function 1 (N = 3) & \multirow{3}{*}{Hybrid.} & 1800 \\
            $f_7$ & Hybrid function 2 (N = 6) & ~ & 2000 \\
            $f_8$ & Hybrid function 3 (N = 5) & ~ & 2200 \\
            \midrule
            $f_9$ & Composition function 1 (N = 5) & \multirow{4}{*}{Comp.}  & 2300 \\
            $f_{10}$ & Composition function 2 (N = 4) & ~ & 2400 \\
            $f_{11}$ & Composition function 3 (N = 5) & ~ & 2600 \\
            $f_{12}$ & Composition function 3 (N = 6) & ~ & 2700 \\
        \midrule
            \multicolumn{4}{c}{Search range: [-100, 100]$^D$} \\
		\bottomrule
	\end{tabular}
\end{table}

\textbf{Competitor algorithms and parameters}: To comprehensively and fairly evaluate the performance of our proposed EMBGO, twelve carefully selected and well-known MA approaches are employed as competitor algorithms. These are listed as follows:

\begin{itemize}
  \item Classic MAs: genetic algorithm (GA) \cite{Srinivas:94}, particle swarm optimization (PSO) \cite{Kennedy:95}, differential evolution (DE) \cite{Storn:97}, and covariance matrix adaptation evolution strategy (CMA-ES) \cite{Hansen:01}. 
  \item Highly-cited MAs: grey wolf optimizer (GWO) \cite{Seyedali:14}, sine cosine algorithm (SCA) \cite{Seyedali:16}, whale optimization algorithm (WOA) \cite{Andrew:16}, and Harris hawks optimization (HHO) \cite{Ali:19}. 
  \item Latest MAs: circle search algorithm (CSA) \cite{Qais:22}, honey badger algorithm (HBA) \cite{Fatma:22}, RIME algorithm \cite{Su:23}, and original MBGO.
\end{itemize}

The population size of all algorithms is 100. The maximum fitness evaluations (FEs) for CEC2017, CEC2020, and CEC2022 benchmark functions are fixed at 1000 $\cdot D$ ($D$=dimension size). To mitigate the impact of randomness in the optimization, each MA is executed with 30 trial runs. The detailed parameter settings of the competitor algorithms are presented in Table \ref{tbl:5.2.4}, and these settings are in accordance with the recommendations provided in the corresponding papers.
\begin{table}[htbp]
	\scriptsize
	\centering
	\renewcommand\arraystretch{1.3}
	\caption{The parameter setting of competitor algorithms.}
	\label{tbl:5.2.4}
		\begin{tabular}{ccc}
			\toprule
			MAs & Parameters & Value \\
			\midrule
			\multirow{3}{*}{GA} & Crossover probability $pc$ & 0.95 \\
			~ & Mutation probability $pm$ & 0.025 \\
			~ & Selection & tournament \\
			\midrule
			\multirow{3}{*}{PSO} & Inertia factor $w$ & 1 \\
			~ & Acceleration coefficients $c_1$ and $c_2$ & 2.05 \\
			~ & Max. and min. speed & 2, -2 \\
			\midrule 
			\multirow{3}{*}{DE} & Mutation scheme & DE/cur-to-rand/1 \\
                ~ & Scaling factor $F$ & 0.8 \\
			~ & Crossover rate $Cr$ & 0.9 \\
			\midrule 
                CMA-ES & \multicolumn{2}{c}{parameter-free} \\
                \midrule
			GWO & \multicolumn{2}{c}{parameter-free} \\
		      \midrule
                SCA & Constant $A$ & 2 \\
		      \midrule
                WOA & Constant $b$ & 1 \\
		      \midrule
                HHO & \multicolumn{2}{c}{parameter-free} \\
		      \midrule
                \multirow{2}{*}{HBA} & Constant $C$ & 2 \\
                ~ & Ability parameter $\beta$ & 6 \\
		      \midrule
                CSA & Constant $C$ & 0.75 \\
			\midrule
			RIME & parameter $w$ & 5 \\
		      \midrule
                MBGO & Ratio of radius & 0.8 and 1.2 \\
		      \midrule
                EMBGO & Ratio of radius & 0.8 and 1.2 \\
			\bottomrule
		\end{tabular}
\end{table}

\begin{figure}[htbp]
    \centering
    \includegraphics[width=16cm]{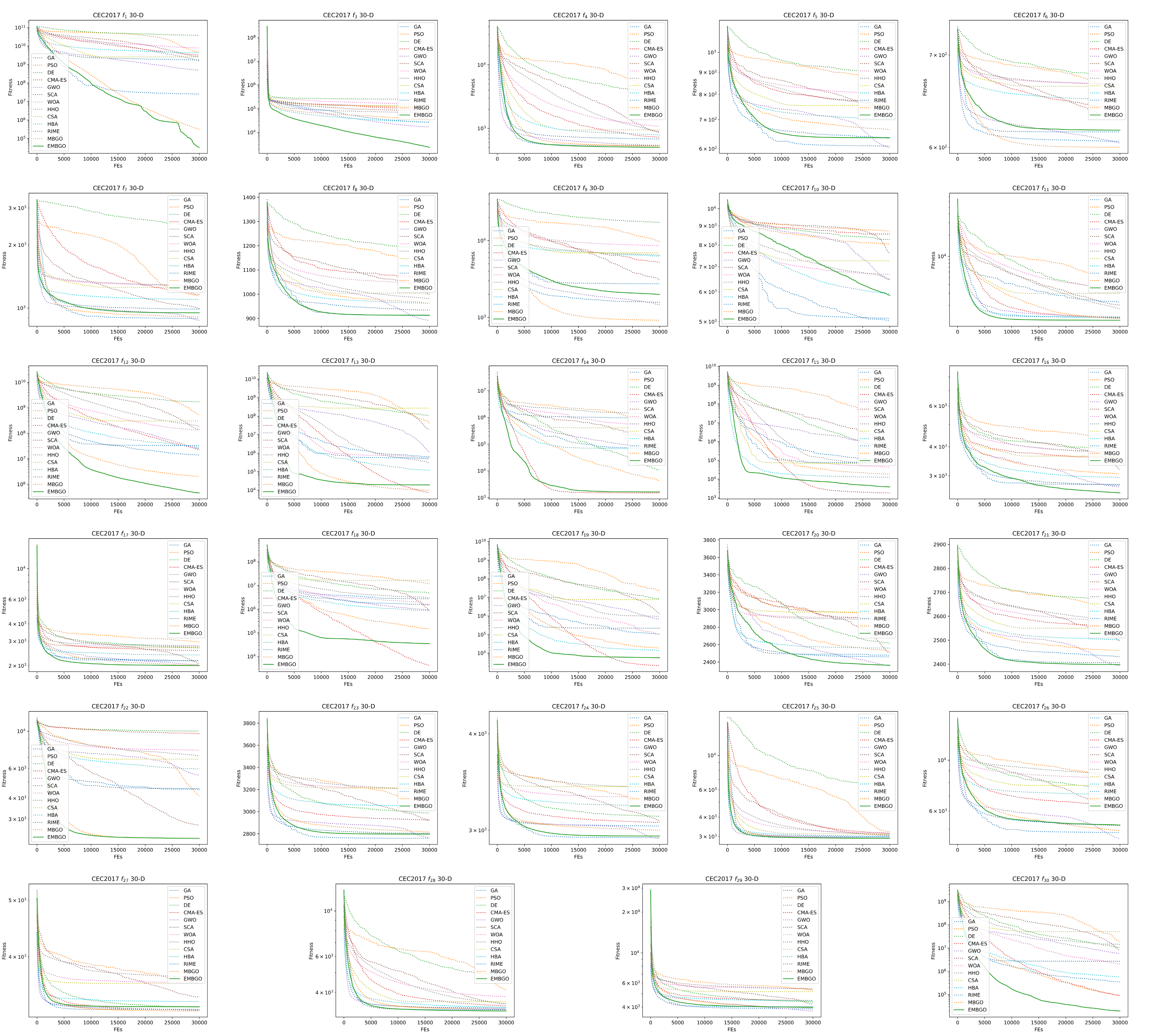}
    \caption{Convergence curves of thirteen MA approaches on 30-D CEC2017 benchmark functions.}
    \label{fig:5.2.1}
\end{figure}

\begin{figure}[htbp]
    \centering
    \includegraphics[width=16cm]{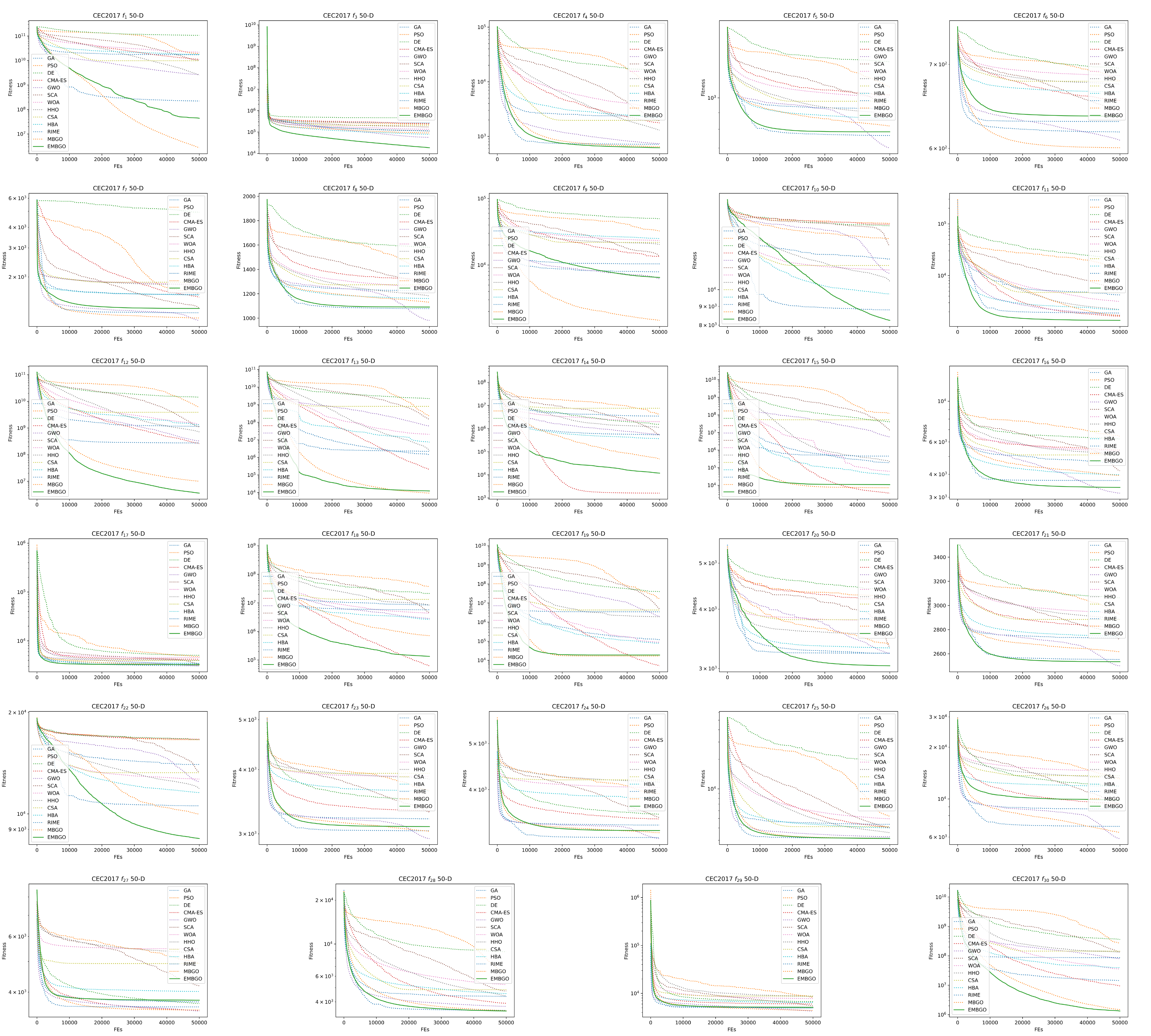}
    \caption{Convergence curves of thirteen MA approaches on 50-D CEC2017 benchmark functions.}
    \label{fig:5.2.2}
\end{figure}

\begin{figure}[htbp]
    \centering
    \includegraphics[width=16cm]{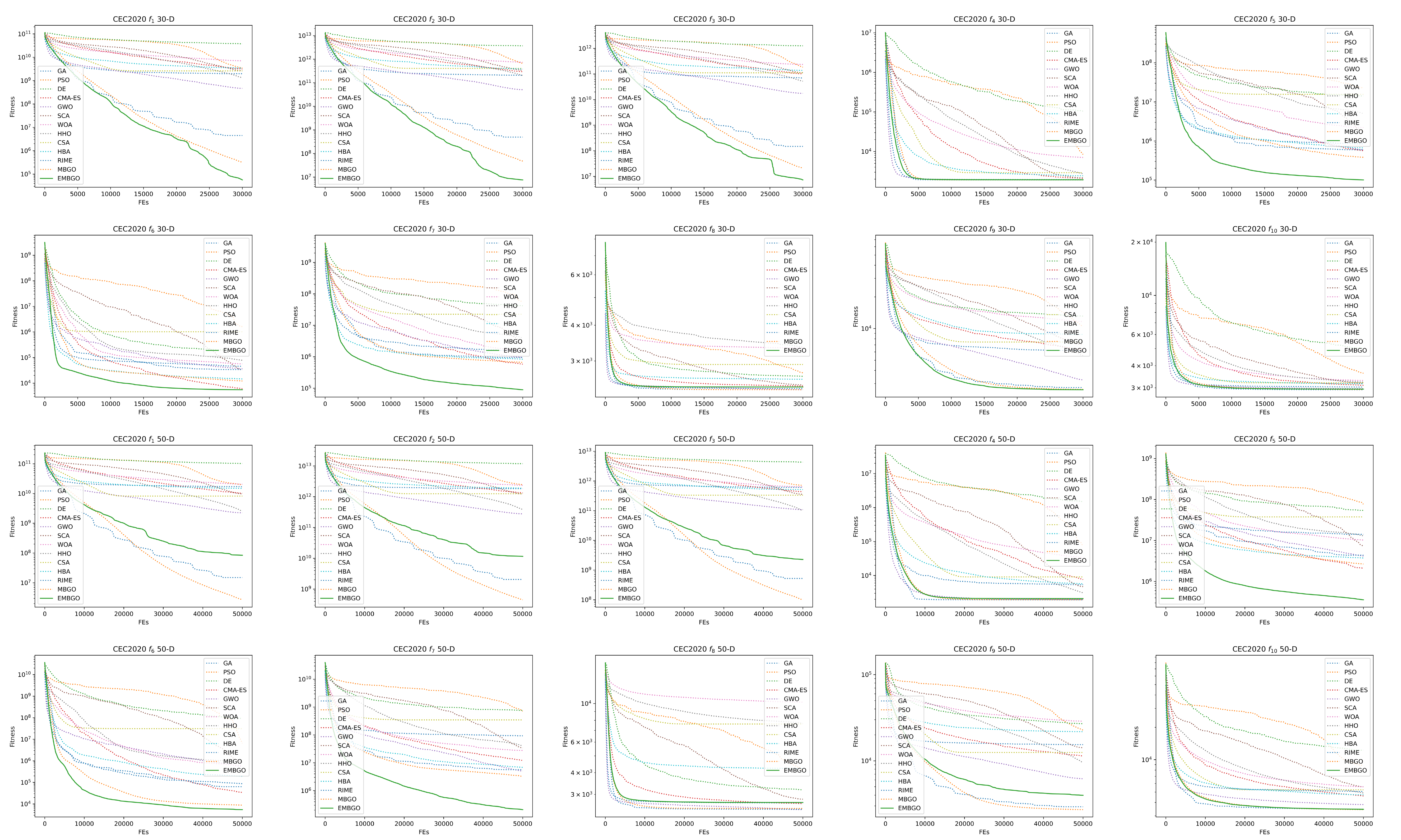}
    \caption{Convergence curves of thirteen MA approaches on CEC2020 benchmark functions.}
    \label{fig:5.2.3}
\end{figure}

\begin{figure}[htbp]
    \centering
    \includegraphics[width=16cm]{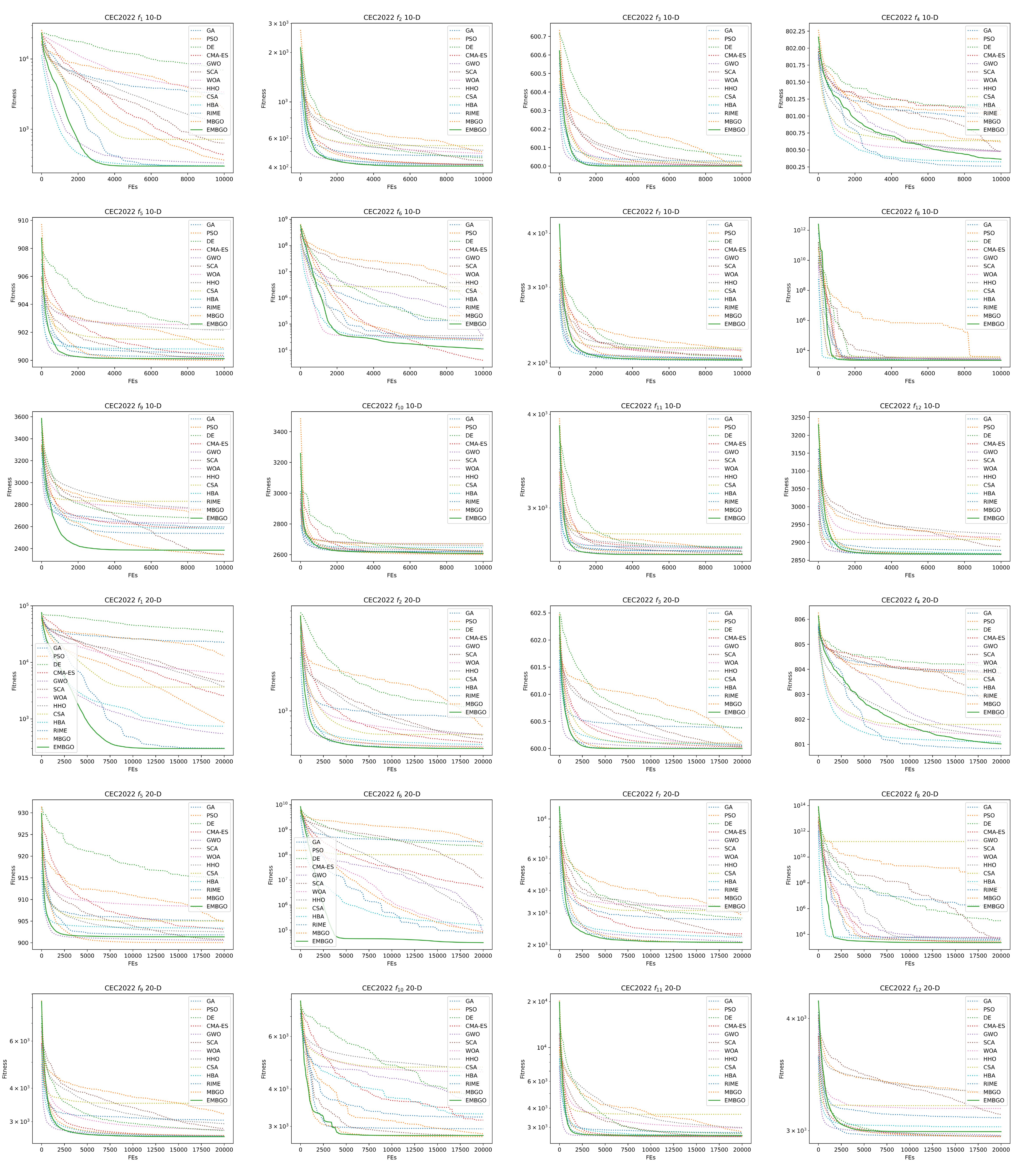}
    \caption{Convergence curves of thirteen MA approaches on CEC2022 benchmark functions.}
    \label{fig:5.2.4}
\end{figure}

\textbf{Experimental results on CEC benchmark functions}: The experimental results and statistical analyses conducted on CEC2017, CEC2020, and CEC2022 benchmark functions are demonstrated. To determine the significance between EMBGO and other competitor algorithms, the Mann–Whitney U test is applied to every pair of algorithms. Subsequently, the Holm multiple comparison test is employed to correct the p-values obtained from the Mann–Whitney U test. Symbols $+$, $\approx$, and $-$ indicate that EMBGO is significantly better, has no significance, or is significantly worse compared to the method being considered. The metric of average rank is computed, and the fitness value of the best-performing algorithm is denoted in bold. Tables \ref{tbl:5.2.5} and \ref{tbl:5.2.6} summarize the results on CEC2017 benchmark functions, Tables \ref{tbl:5.2.7} and \ref{tbl:5.2.8} summarize the results on CEC2020 benchmark functions, and Tables \ref{tbl:5.2.9} and \ref{tbl:5.2.10} summarize the results on CEC2022 benchmark functions. The corresponding convergence curves are presented in Figures \ref{fig:5.2.1}, \ref{fig:5.2.2}, \ref{fig:5.2.3}, and \ref{fig:5.2.4}.
\begin{sidewaystable}[htbp]
	\scriptsize
	\centering
	\renewcommand\arraystretch{1.5}
	\caption{Experimental results and statistical analyses on 30-D CEC2017 benchmark functions. $f_1$: Unimodal function; $f_3-f_9$: Simple multimodal functions; $f_{10}-f_{19}$: Hybrid functions; $f_{20}-f_{30}$: Composition functions. Due to the limitation of space, only the mean of optima among 30 trial runs is provided}
	\label{tbl:5.2.5}
	\resizebox{\columnwidth}{!}{
		\begin{tabular}{ccccccccccccccccc}
			\toprule
			Func. & GA & PSO & DE & CMA-ES & GWO & SCA & WOA & HHO & CSA & HBA & RIME & MBGO & EMBGO \\
			\midrule
			$f_1$ & 1.787e+09 $+$ & 4.350e+09 $+$ & 3.873e+10 $+$ & 3.202e+09 $+$ & 5.099e+08 $+$ & 2.753e+09 $+$ & 7.948e+09 $+$ & 1.612e+09 $+$ & 2.597e+09 $+$ & 4.422e+09 $+$ & 2.536e+07 $+$ & 3.143e+05 $+$ & \textbf{3.350e+04} \\
			\midrule
			$f_3$ & 1.296e+05 $+$ & 1.091e+05 $+$ & 2.507e+05 $+$ & 9.489e+04 $+$ & 1.716e+04 $+$ & 7.853e+04 $+$ & 1.521e+05 $+$ & 6.281e+04 $+$ & 1.371e+05 $+$ & 2.642e+04 $+$ & 2.688e+04 $+$ & 3.557e+04 $+$ & \textbf{2.349e+03} \\
			$f_4$ & 6.765e+02 $+$ & 2.386e+03 $+$ & 3.566e+03 $+$ & 7.201e+02 $+$ & 5.408e+02 $+$ & 8.855e+02 $+$ & 1.075e+03 $+$ & 8.577e+02 $+$ & 9.443e+02 $+$ & 7.855e+02 $+$ & 5.359e+02 $+$ & 5.236e+02 $+$ & \textbf{5.024e+02} \\
			$f_5$ & 6.345e+02 $\approx$ & 8.444e+02 $+$ & 8.989e+02 $+$ & 7.665e+02 $+$ & \textbf{6.048e+02} $-$ & 7.141e+02 $+$ & 8.027e+02 $+$ & 7.543e+02 $+$ & 7.550e+02 $+$ & 7.023e+02 $+$ & 6.088e+02 $-$ & 6.641e+02 $+$ & 6.364e+02 \\
			$f_6$ & 6.062e+02 $-$ & 6.566e+02 $+$ & 6.742e+02 $+$ & 6.404e+02 $+$ & 6.045e+02 $-$ & 6.300e+02 $+$ & 6.666e+02 $+$ & 6.664e+02 $+$ & 6.645e+02 $+$ & 6.500e+02 $+$ & 6.158e+02 $\approx$ & \textbf{6.001e+02} $-$ & 6.175e+02 \\
			$f_7$ & 9.887e+02 $+$ & 1.147e+03 $+$ & 2.430e+03 $+$ & 1.161e+03 $+$ & \textbf{8.760e+02} $-$ & 9.962e+02 $+$ & 1.277e+03 $+$ & 1.281e+03 $+$ & 1.256e+03 $+$ & 1.100e+03 $+$ & 8.935e+02 $-$ & 9.084e+02 $\approx$ & 9.503e+02 \\
			$f_8$ & 9.343e+02 $+$ & 1.101e+03 $+$ & 1.190e+03 $+$ & 1.067e+03 $+$ & \textbf{8.926e+02} $-$ & 9.989e+02 $+$ & 1.045e+03 $+$ & 9.810e+02 $+$ & 1.004e+03 $+$ & 9.624e+02 $+$ & 9.122e+02 $\approx$ & 9.625e+02 $+$ & 9.133e+02 \\
			$f_9$ & 1.576e+03 $-$ & 9.754e+03 $+$ & 1.733e+04 $+$ & 5.216e+03 $+$ & 1.462e+03 $-$ & 3.105e+03 $+$ & 8.646e+03 $+$ & 6.458e+03 $+$ & 6.943e+03 $+$ & 6.301e+03 $+$ & 2.751e+03 $+$ & \textbf{9.187e+02} $-$ & 1.999e+03 \\
			\midrule

			$f_{10}$ & \textbf{5.045e+03} $-$ & 8.728e+03 $+$ & 8.270e+03 $+$ & 8.558e+03 $+$ & 6.476e+03 $+$ & 7.634e+03 $+$ & 6.679e+03 $+$ & 6.465e+03 $+$ & 7.265e+03 $+$ & 5.930e+03 $+$ & 5.104e+03 $\approx$ & 8.055e+03 $+$ & 5.873e+03 \\
			$f_{11}$ & 2.226e+03 $+$ & 3.970e+03 $+$ & 3.132e+03 $+$ & 1.343e+03 $+$ & 1.331e+03 $+$ & 2.040e+03 $+$ & 3.513e+03 $+$ & 1.803e+03 $+$ & 2.978e+03 $+$ & 1.346e+03 $+$ & 1.381e+03 $+$ & 1.315e+03 $+$ & \textbf{1.241e+03} \\
			$f_{12}$ & 1.406e+07 $+$ & 5.129e+08 $+$ & 1.721e+09 $+$ & 2.240e+07 $+$ & 2.507e+07 $+$ & 1.420e+08 $+$ & 1.343e+08 $+$ & 1.931e+08 $+$ & 3.067e+08 $+$ & 2.713e+07 $+$ & 3.290e+07 $+$ & 1.995e+06 $+$ & \textbf{4.604e+05} \\
			$f_{13}$ & 6.388e+05 $+$ & 4.287e+07 $+$ & 1.103e+08 $+$ & \textbf{7.161e+03} $\approx$ & 1.209e+06 $+$ & 1.889e+07 $+$ & 5.584e+05 $+$ & 2.866e+05 $+$ & 2.727e+08 $+$ & 1.212e+05 $+$ & 5.132e+05 $+$ & 1.008e+04 $\approx$ & 1.931e+04 \\
			$f_{14}$ & 9.630e+05 $+$ & 5.616e+05 $+$ & 1.053e+04 $+$ & \textbf{1.508e+03} $-$ & 7.418e+04 $+$ & 7.593e+04 $+$ & 4.249e+05 $+$ & 1.224e+06 $+$ & 4.132e+05 $+$ & 6.460e+04 $+$ & 6.939e+04 $+$ & 4.411e+03 $+$ & 1.635e+03 \\
			$f_{15}$ & 8.551e+04 $+$ & 5.590e+06 $+$ & 4.137e+05 $+$ & \textbf{1.839e+03} $-$ & 3.057e+05 $+$ & 3.888e+05 $+$ & 4.275e+04 $+$ & 7.128e+04 $+$ & 6.925e+04 $+$ & 1.252e+04 $+$ & 7.331e+04 $+$ & 1.827e+04 $+$ & 3.869e+03 \\
			$f_{16}$ & 2.729e+03 $+$ & 4.288e+03 $+$ & 3.914e+03 $+$ & 3.585e+03 $+$ & 2.688e+03 $\approx$ & 3.203e+03 $+$ & 3.798e+03 $+$ & 3.718e+03 $+$ & 3.647e+03 $+$ & 2.954e+03 $+$ & 2.755e+03 $+$ & 3.056e+03 $+$ & \textbf{2.532e+03} \\
			$f_{17}$ & 2.178e+03 $+$ & 2.966e+03 $+$ & 2.672e+03 $+$ & 2.572e+03 $+$ & 2.032e+03 $+$ & 2.085e+03 $+$ & 2.753e+03 $+$ & 2.775e+03 $+$ & 2.702e+03 $+$ & 2.389e+03 $+$ & 2.173e+03 $+$ & 2.012e+03 $\approx$ & \textbf{2.005e+03} \\
			$f_{18}$ & 2.745e+06 $+$ & 1.188e+07 $+$ & 4.787e+06 $+$ & \textbf{4.140e+03} $-$ & 9.544e+05 $+$ & 9.112e+05 $+$ & 1.551e+06 $+$ & 3.093e+06 $+$ & 1.654e+07 $+$ & 8.655e+05 $+$ & 1.523e+06 $+$ & 1.472e+05 $+$ & 3.517e+04 \\
			$f_{19}$ & 1.062e+05 $+$ & 2.023e+07 $+$ & 8.323e+06 $+$ & \textbf{2.140e+03} $-$ & 6.653e+05 $+$ & 1.320e+06 $+$ & 1.010e+05 $+$ & 9.421e+05 $+$ & 7.599e+06 $+$ & 1.370e+04 $+$ & 2.190e+05 $+$ & 1.812e+04 $+$ & 5.677e+03 \\
			\midrule
			$f_{20}$ & 2.460e+03 $+$ & 2.952e+03 $+$ & 2.617e+03 $+$ & 2.880e+03 $+$ & \textbf{2.359e+03} $\approx$ & 2.507e+03 $+$ & 2.904e+03 $+$ & 2.882e+03 $+$ & 2.973e+03 $+$ & 2.560e+03 $+$ & 2.478e+03 $+$ & 2.529e+03 $+$ & 2.363e+03 \\
			$f_{21}$ & 2.431e+03 $+$ & 2.603e+03 $+$ & 2.665e+03 $+$ & 2.558e+03 $+$ & \textbf{2.394e+03} $-$ & 2.498e+03 $+$ & 2.580e+03 $+$ & 2.581e+03 $+$ & 2.550e+03 $+$ & 2.502e+03 $+$ & 2.406e+03 $\approx$ & 2.457e+03 $+$ & 2.398e+03 \\
			$f_{22}$ & 4.485e+03 $+$ & 4.129e+03 $+$ & 1.003e+04 $+$ & 9.677e+03 $+$ & 5.484e+03 $+$ & 2.769e+03 $+$ & 7.721e+03 $+$ & 7.134e+03 $+$ & 6.823e+03 $+$ & 5.971e+03 $+$ & 4.565e+03 $+$ & \textbf{2.307e+03} $\approx$ & 2.308e+03 \\
			$f_{23}$ & 2.804e+03 $\approx$ & 3.084e+03 $+$ & 2.985e+03 $+$ & 2.925e+03 $+$ & 2.767e+03 $\approx$ & 2.922e+03 $+$ & 3.207e+03 $+$ & 3.194e+03 $+$ & 3.213e+03 $+$ & 3.050e+03 $+$ & \textbf{2.758e+03} $-$ & 2.816e+03 $+$ & 2.797e+03 \\
			$f_{24}$ & 3.039e+03 $+$ & 3.238e+03 $+$ & 3.128e+03 $+$ & 3.073e+03 $+$ & \textbf{2.929e+03} $-$ & 3.089e+03 $+$ & 3.309e+03 $+$ & 3.390e+03 $+$ & 3.418e+03 $+$ & 3.229e+03 $+$ & 2.936e+03 $\approx$ & 3.000e+03 $+$ & 2.949e+03 \\
			$f_{25}$ & 2.995e+03 $+$ & 3.218e+03 $+$ & 6.450e+03 $+$ & 3.144e+03 $+$ & 2.953e+03 $+$ & 3.051e+03 $+$ & 3.139e+03 $+$ & 3.081e+03 $+$ & 3.123e+03 $+$ & 3.014e+03 $+$ & 2.941e+03 $+$ & \textbf{2.921e+03} $\approx$ & 2.923e+03 \\
			$f_{26}$ & 5.195e+03 $\approx$ & 8.205e+03 $+$ & 7.597e+03 $+$ & 6.366e+03 $+$ & \textbf{4.563e+03} $-$ & 6.016e+03 $+$ & 8.347e+03 $+$ & 8.677e+03 $+$ & 7.770e+03 $+$ & 7.087e+03 $+$ & 4.840e+03 $-$ & 4.932e+03 $-$ & 5.212e+03 \\
			$f_{27}$ & 3.253e+03 $-$ & 3.616e+03 $+$ & 3.281e+03 $\approx$ & 3.241e+03 $-$ & 3.237e+03 $-$ & 3.413e+03 $+$ & 3.615e+03 $+$ & 3.692e+03 $+$ & 3.602e+03 $+$ & 3.354e+03 $+$ & 3.249e+03 $-$ & \textbf{3.231e+03} $-$ & 3.286e+03 \\
			$f_{28}$ & 3.367e+03 $+$ & 4.129e+03 $+$ & 4.790e+03 $+$ & 3.448e+03 $+$ & 3.325e+03 $+$ & 3.527e+03 $+$ & 3.820e+03 $+$ & 3.534e+03 $+$ & 3.612e+03 $+$ & 3.396e+03 $+$ & 3.300e+03 $+$ & 3.302e+03 $+$ & \textbf{3.253e+03} \\
			$f_{29}$ & 3.873e+03 $\approx$ & 5.333e+03 $+$ & 4.353e+03 $+$ & 4.464e+03 $+$ & \textbf{3.740e+03} $-$ & 4.205e+03 $+$ & 5.487e+03 $+$ & 5.400e+03 $+$ & 5.192e+03 $+$ & 4.445e+03 $+$ & 4.023e+03 $\approx$ & 3.929e+03 $\approx$ & 3.987e+03 \\
			$f_{30}$ & 3.500e+05 $+$ & 1.910e+07 $+$ & 1.060e+07 $+$ & 9.112e+04 $+$ & 5.912e+06 $+$ & 8.187e+06 $+$ & 2.056e+06 $+$ & 1.452e+07 $+$ & 5.126e+07 $+$ & 5.729e+05 $+$ & 2.719e+06 $+$ & 8.865e+04 $+$ & \textbf{1.970e+04} \\
			\midrule
			$+$/$\approx$/$-$ & 21/4/4 & 29/0/0 & 28/1/0 & 23/1/5 & 16/3/10 & 29/0/0 & 29/0/0 & 29/0/0 & 29/0/0 & 29/0/0 & 18/6/5 & 19/6/4 & - \\
                \midrule
                Avg. rank & 5.2 & 11.2 & 11.0 & 6.6 & 3.7 & 10.0 & 7.2 & 9.4 & 10.1 & 6.2 & 4.2 & 3.3 & \textbf{2.5} \\
			\bottomrule
		\end{tabular}
	}
\end{sidewaystable}
\begin{sidewaystable}[htbp]
	\scriptsize
	\centering
	\renewcommand\arraystretch{1.5}
	\caption{Experimental results and statistical analyses on 50-D CEC2017 benchmark functions.}
	\label{tbl:5.2.6}
	\resizebox{\columnwidth}{!}{
		\begin{tabular}{ccccccccccccccccc}
			\toprule
			Func. & GA & PSO & DE & CMA-ES & GWO & SCA & WOA & HHO & CSA & HBA & RIME & MBGO & EMBGO \\
			\midrule
			$f_1$ & 1.726e+10 $+$ & 1.907e+10 $+$ & 1.074e+11 $+$ & 1.106e+10 $+$ & 2.699e+09 $+$ & 1.025e+10 $+$ & 2.188e+10 $+$ & 2.542e+09 $+$ & 9.802e+09 $+$ & 1.779e+10 $+$ & 2.223e+08 $+$ & \textbf{2.749e+06} $-$ & 4.398e+07 \\
			$f_3$ & 2.656e+05 $+$ & 2.338e+05 $+$ & 4.624e+05 $+$ & 2.449e+05 $+$ & 5.636e+04 $+$ & 1.732e+05 $+$ & 1.340e+05 $+$ & 1.091e+05 $+$ & 2.077e+05 $+$ & 7.803e+04 $+$ & 1.162e+05 $+$ & 9.073e+04 $+$ & \textbf{1.857e+04} \\
			$f_4$ & 2.321e+03 $+$ & 5.625e+03 $+$ & 1.593e+04 $+$ & 1.771e+03 $+$ & 7.396e+02 $+$ & 1.956e+03 $+$ & 3.348e+03 $+$ & 1.301e+03 $+$ & 1.964e+03 $+$ & 2.175e+03 $+$ & 7.161e+02 $+$ & 6.489e+02 $+$ & \textbf{6.206e+02} \\
			$f_5$ & 9.263e+02 $+$ & 1.080e+03 $+$ & 1.293e+03 $+$ & 1.021e+03 $+$ & \textbf{7.018e+02} $-$ & 9.193e+02 $+$ & 1.004e+03 $+$ & 8.922e+02 $+$ & 9.760e+02 $+$ & 8.630e+02 $+$ & 7.652e+02 $\approx$ & 8.194e+02 $+$ & 7.855e+02 \\
			$f_6$ & 6.183e+02 $-$ & 6.795e+02 $+$ & 6.949e+02 $+$ & 6.550e+02 $+$ & 6.092e+02 $-$ & 6.435e+02 $+$ & 6.857e+02 $+$ & 6.749e+02 $+$ & 6.787e+02 $+$ & 6.655e+02 $+$ & 6.303e+02 $\approx$ & \textbf{6.007e+02} $-$ & 6.364e+02 \\
			$f_7$ & 1.566e+03 $+$ & 1.593e+03 $+$ & 4.928e+03 $+$ & 1.510e+03 $+$ & \textbf{1.092e+03} $-$ & 1.306e+03 $\approx$ & 1.809e+03 $+$ & 1.781e+03 $+$ & 1.859e+03 $+$ & 1.554e+03 $+$ & 1.214e+03 $\approx$ & 1.129e+03 $\approx$ & 1.290e+03 \\
			$f_8$ & 1.224e+03 $+$ & 1.352e+03 $+$ & 1.581e+03 $+$ & 1.312e+03 $+$ & \textbf{9.815e+02} $-$ & 1.221e+03 $+$ & 1.255e+03 $+$ & 1.182e+03 $+$ & 1.274e+03 $+$ & 1.161e+03 $+$ & 1.078e+03 $\approx$ & 1.131e+03 $+$ & 1.091e+03 \\
			$f_9$ & 7.915e+03 $+$ & 3.250e+04 $+$ & 4.987e+04 $+$ & 1.338e+04 $+$ & 6.227e+03 $\approx$ & 1.353e+04 $+$ & 2.340e+04 $+$ & 2.042e+04 $+$ & 2.193e+04 $+$ & 2.504e+04 $+$ & 1.041e+04 $+$ & \textbf{1.469e+03} $-$ & 6.500e+03 \\
			\midrule
			$f_{10}$ & 1.212e+04 $+$ & 1.518e+04 $+$ & 1.495e+04 $+$ & 1.508e+04 $+$ & 1.108e+04 $+$ & 1.319e+04 $+$ & 1.132e+04 $+$ & 1.057e+04 $+$ & 1.164e+04 $+$ & 9.733e+03 $+$ & 8.805e+03 $+$ & 1.379e+04 $+$ & \textbf{8.241e+03} \\
			$f_{11}$ & 4.243e+03 $+$ & 1.317e+04 $+$ & 2.093e+04 $+$ & 1.642e+03 $+$ & 1.720e+03 $+$ & 5.699e+03 $+$ & 3.173e+03 $+$ & 2.205e+03 $+$ & 4.728e+03 $+$ & 2.191e+03 $+$ & 1.895e+03 $+$ & 1.680e+03 $+$ & \textbf{1.363e+03} \\
			$f_{12}$ & 1.144e+09 $+$ & 6.037e+09 $+$ & 1.441e+10 $+$ & 2.571e+08 $+$ & 3.275e+08 $+$ & 1.121e+09 $+$ & 1.333e+09 $+$ & 7.205e+08 $+$ & 3.889e+09 $+$ & 1.316e+09 $+$ & 2.746e+08 $+$ & 1.005e+07 $+$ & \textbf{3.606e+06} \\
			$f_{13}$ & 1.446e+06 $+$ & 2.179e+08 $+$ & 2.222e+09 $+$ & 2.115e+05 $+$ & 6.219e+07 $+$ & 1.526e+08 $+$ & 2.054e+07 $+$ & 2.438e+06 $+$ & 8.009e+08 $+$ & 7.404e+06 $+$ & 2.222e+06 $+$ & \textbf{9.303e+03} $\approx$ & 1.231e+04 \\
			$f_{14}$ & 3.481e+06 $+$ & 4.154e+06 $+$ & 1.537e+06 $+$ & \textbf{1.622e+03} $-$ & 5.598e+05 $+$ & 5.315e+05 $+$ & 1.991e+06 $+$ & 1.108e+06 $+$ & 7.575e+06 $+$ & 3.663e+05 $+$ & 5.392e+05 $+$ & 4.853e+04 $+$ & 1.178e+04 \\
			$f_{15}$ & 1.921e+05 $+$ & 1.223e+08 $+$ & 4.088e+07 $+$ & \textbf{3.582e+03} $-$ & 5.667e+06 $+$ & 1.402e+07 $+$ & 6.214e+04 $+$ & 2.221e+05 $+$ & 5.281e+07 $+$ & 3.936e+04 $+$ & 4.550e+05 $+$ & 7.263e+03 $\approx$ & 1.089e+04 \\
			$f_{16}$ & 4.608e+03 $+$ & 6.448e+03 $+$ & 6.230e+03 $+$ & 5.425e+03 $+$ & \textbf{3.160e+03} $-$ & 4.187e+03 $+$ & 5.345e+03 $+$ & 5.184e+03 $+$ & 5.098e+03 $+$ & 3.980e+03 $+$ & 3.700e+03 $+$ & 3.943e+03 $+$ & 3.393e+03 \\
			$f_{17}$ & 3.092e+03 $\approx$ & 4.755e+03 $+$ & 5.068e+03 $+$ & 4.226e+03 $+$ & \textbf{3.091e+03} $\approx$ & 3.378e+03 $\approx$ & 4.561e+03 $+$ & 3.975e+03 $+$ & 3.953e+03 $+$ & 3.489e+03 $\approx$ & 3.361e+03 $\approx$ & 3.203e+03 $\approx$ & 3.270e+03 \\
			$f_{18}$ & 8.380e+06 $+$ & 3.663e+07 $+$ & 2.135e+07 $+$ & \textbf{6.284e+04} $-$ & 2.857e+06 $+$ & 4.662e+06 $+$ & 4.380e+06 $+$ & 8.408e+06 $+$ & 1.324e+07 $+$ & 2.624e+06 $+$ & 5.731e+06 $+$ & 7.103e+05 $+$ & 1.342e+05 \\
			$f_{19}$ & 1.211e+05 $+$ & 1.750e+07 $+$ & 3.896e+07 $+$ & \textbf{5.168e+03} $-$ & 2.021e+06 $+$ & 5.190e+06 $+$ & 8.860e+04 $+$ & 1.895e+06 $+$ & 4.336e+06 $+$ & 7.797e+04 $+$ & 3.493e+06 $+$ & 1.689e+04 $\approx$ & 1.902e+04 \\
			\midrule
			$f_{20}$ & 3.231e+03 $+$ & 4.234e+03 $+$ & 4.391e+03 $+$ & 4.202e+03 $+$ & 3.230e+03 $\approx$ & 3.332e+03 $+$ & 3.788e+03 $+$ & 3.529e+03 $+$ & 3.801e+03 $+$ & 3.316e+03 $+$ & 3.231e+03 $+$ & 3.385e+03 $+$ & \textbf{3.040e+03} \\
			$f_{21}$ & 2.729e+03 $+$ & 2.924e+03 $+$ & 3.064e+03 $+$ & 2.817e+03 $+$ & \textbf{2.500e+03} $-$ & 2.721e+03 $+$ & 2.937e+03 $+$ & 2.884e+03 $+$ & 2.884e+03 $+$ & 2.745e+03 $+$ & 2.553e+03 $\approx$ & 2.617e+03 $+$ & 2.534e+03 \\
			$f_{22}$ & 1.402e+04 $+$ & 1.659e+04 $+$ & 1.663e+04 $+$ & 1.667e+04 $+$ & 1.248e+04 $+$ & 1.332e+04 $+$ & 1.269e+04 $+$ & 1.190e+04 $+$ & 1.325e+04 $+$ & 1.157e+04 $+$ & 1.057e+04 $+$ & 9.986e+03 $\approx$ & \textbf{8.453e+03} \\
			$f_{23}$ & 3.208e+03 $+$ & 3.637e+03 $+$ & 3.495e+03 $+$ & 3.321e+03 $+$ & \textbf{2.938e+03} $-$ & 3.309e+03 $+$ & 3.880e+03 $+$ & 3.833e+03 $+$ & 3.933e+03 $+$ & 3.635e+03 $+$ & 3.042e+03 $\approx$ & 3.035e+03 $\approx$ & 3.099e+03 \\
			$f_{24}$ & 3.367e+03 $+$ & 3.917e+03 $+$ & 3.550e+03 $+$ & 3.466e+03 $+$ & \textbf{3.155e+03} $-$ & 3.496e+03 $+$ & 4.042e+03 $+$ & 4.147e+03 $+$ & 4.193e+03 $+$ & 3.842e+03 $+$ & 3.170e+03 $-$ & 3.246e+03 $\approx$ & 3.280e+03 \\
			$f_{25}$ & 4.358e+03 $+$ & 5.270e+03 $+$ & 1.873e+04 $+$ & 4.023e+03 $+$ & 3.261e+03 $+$ & 4.101e+03 $+$ & 4.965e+03 $+$ & 3.594e+03 $+$ & 4.131e+03 $+$ & 4.007e+03 $+$ & 3.164e+03 $+$ & 3.140e+03 $\approx$ & \textbf{3.128e+03} \\
			$f_{26}$ & 8.731e+03 $-$ & 1.309e+04 $+$ & 1.186e+04 $+$ & 9.398e+03 $-$ & \textbf{5.885e+03} $-$ & 9.333e+03 $-$ & 1.446e+04 $+$ & 1.330e+04 $+$ & 1.367e+04 $+$ & 1.186e+04 $+$ & 6.920e+03 $-$ & 6.386e+03 $-$ & 9.866e+03 \\
			$f_{27}$ & 3.726e+03 $\approx$ & 4.937e+03 $+$ & 3.683e+03 $\approx$ & 3.495e+03 $-$ & \textbf{3.487e+03} $-$ & 4.192e+03 $+$ & 5.485e+03 $+$ & 5.286e+03 $+$ & 4.925e+03 $+$ & 4.020e+03 $+$ & 3.594e+03 $-$ & 3.514e+03 $-$ & 3.772e+03 \\
			$f_{28}$ & 4.355e+03 $+$ & 6.329e+03 $+$ & 8.856e+03 $+$ & 3.905e+03 $+$ & 3.727e+03 $+$ & 4.751e+03 $+$ & 5.258e+03 $+$ & 4.388e+03 $+$ & 4.850e+03 $+$ & 4.557e+03 $+$ & 3.495e+03 $\approx$ & 3.480e+03 $\approx$ & \textbf{3.460e+03} \\
			$f_{29}$ & 4.908e+03 $-$ & 8.236e+03 $+$ & 6.730e+03 $+$ & 5.997e+03 $+$ & \textbf{4.284e+03} $-$ & 5.491e+03 $+$ & 7.779e+03 $+$ & 8.555e+03 $+$ & 8.790e+03 $+$ & 6.443e+03 $+$ & 5.071e+03 $\approx$ & 4.401e+03 $-$ & 5.114e+03 \\
			$f_{30}$ & 1.440e+07 $+$ & 2.662e+08 $+$ & 3.577e+08 $+$ & 9.529e+06 $+$ & 8.078e+07 $+$ & 1.379e+08 $+$ & 3.562e+07 $+$ & 1.342e+08 $+$ & 1.450e+08 $+$ & 3.971e+07 $+$ & 8.678e+07 $+$ & 1.492e+06 $\approx$ & \textbf{1.332e+06} \\
			\midrule
			$+$/$\approx$/$-$ & 24/2/3 & 29/0/0 & 28/1/0 & 23/0/6 & 15/3/11 & 26/2/1 & 29/0/0 & 29/0/0 & 29/0/0 & 28/1/0 & 17/9/3 & 12/11/6 & - \\
                \midrule
                Avg. rank & 6.6 & 11.5 & 11.7 & 6.4 & 3.4 & 7.4 & 9.5 & 7.8 & 10.1 & 6.5 & 4.2 & 2.9 & \textbf{2.6} \\
			\bottomrule
		\end{tabular}
	}
\end{sidewaystable}
\begin{sidewaystable}[htbp]
	\scriptsize
	\centering
	\renewcommand\arraystretch{1.5}
	\caption{Experimental results and statistical analyses on 30-D CEC2020 benchmark functions. $f_1$: Unimodal function; $f_2-f_4$: Multimodal functions; $f_5-f_7$: Hybrid functions; $f_8-f_{10}$: Composition functions; mean and std: the mean and the standard deviation of 30 trial runs.}
	\label{tbl:5.2.7}
	\resizebox{\columnwidth}{!}{
		\begin{tabular}{ccccccccccccccccc}
			\toprule
			\multicolumn{2}{c}{Func.} & GA & PSO & DE & CMA-ES & GWO & SCA & WOA & HHO & CSA & HBA & RIME & MBGO & EMBGO \\
			\midrule
			\multirow{2}{*}{$f_1$} & mean & 1.983e+09 $+$ & 3.479e+09 $+$ & 3.796e+10 $+$ & 3.154e+09 $+$ & 4.704e+08 $+$ & 2.678e+09 $+$ & 7.037e+09 $+$ & 1.352e+09 $+$ & 2.612e+09 $+$ & 3.445e+09 $+$ & 4.507e+06 $+$ & 3.196e+05 $+$ & \textbf{5.651e+04} \\
                ~ & std & 3.017e+08 & 1.505e+09 & 4.492e+09 & 5.271e+08 & 3.956e+08 & 4.755e+08 & 4.018e+09 & 5.796e+08 & 1.515e+09 & 3.683e+09 & 1.777e+06 & 2.001e+05 & 1.171e+05 \\
                \midrule
			\multirow{2}{*}{$f_2$} & mean & 2.087e+11 $+$ & 6.703e+11 $+$ & 3.706e+12 $+$ & 2.971e+11 $+$ & 5.126e+10 $+$ & 3.438e+11 $+$ & 7.427e+11 $+$ & 2.081e+11 $+$ & 4.065e+11 $+$ & 3.815e+11 $+$ & 4.976e+08 $+$ & 4.790e+07 $+$ & \textbf{7.621e+06} \\
                ~ & std & 3.854e+10 & 4.442e+11 & 4.166e+11 & 4.725e+10 & 6.168e+10 & 7.599e+10 & 4.105e+11 & 8.444e+10 & 3.589e+11 & 2.296e+11 & 3.341e+08 & 2.438e+07 & 2.319e+07 \\
			\multirow{2}{*}{$f_3$} & mean & 6.981e+10 $+$ & 1.855e+11 $+$ & 1.269e+12 $+$ & 1.009e+11 $+$ & 1.765e+10 $+$ & 1.131e+11 $+$ & 2.369e+11 $+$ & 5.251e+10 $+$ & 1.112e+11 $+$ & 1.383e+11 $+$ & 1.485e+08 $+$ & 2.067e+07 $+$ & \textbf{7.312e+06} \\
                ~ & std & 1.205e+10 & 1.106e+11 & 1.409e+11 & 1.864e+10 & 1.446e+10 & 2.130e+10 & 1.289e+11 & 2.744e+10 & 8.701e+10 & 1.045e+11 & 7.087e+07 & 1.310e+07 & 2.398e+07 \\
			\multirow{2}{*}{$f_4$} & mean & 1.941e+03 $+$ & 8.403e+03 $+$ & 1.046e+05 $+$ & 2.104e+03 $+$ & 1.919e+03 $\approx$ & 2.159e+03 $+$ & 7.069e+03 $+$ & 2.767e+03 $+$ & 2.917e+03 $+$ & 2.428e+03 $+$ & \textbf{1.913e+03} $-$ & 1.918e+03 $\approx$ & 1.929e+03 \\
                ~ & std & 6.685e+00 & 1.238e+04 & 4.793e+04 & 1.120e+02 & 4.666e+00 & 1.728e+02 & 6.069e+03 & 1.324e+03 & 1.595e+03 & 7.134e+02 & 2.231e+00 & 2.509e+00 & 1.221e+01 \\
                \midrule
			\multirow{2}{*}{$f_5$} & mean & 8.487e+05 $+$ & 2.238e+07 $+$ & 1.426e+07 $+$ & 5.498e+05 $+$ & 5.930e+05 $+$ & 1.734e+06 $+$ & 2.210e+06 $+$ & 5.071e+06 $+$ & 1.565e+07 $+$ & 6.657e+05 $+$ & 5.991e+05 $+$ & 3.826e+05 $+$ & \textbf{1.012e+05} \\
                ~ & std & 2.604e+05 & 2.113e+07 & 4.766e+06 & 1.614e+05 & 3.503e+05 & 8.428e+05 & 4.339e+06 & 3.986e+06 & 3.571e+07 & 4.428e+05 & 3.293e+05 & 1.643e+05 & 3.969e+04 \\
			\multirow{2}{*}{$f_6$} & mean & 4.540e+04 $+$ & 1.424e+06 $+$ & 1.907e+05 $+$ & 6.081e+03 $\approx$ & 3.680e+04 $+$ & 3.318e+04 $+$ & 5.669e+04 $+$ & 7.908e+04 $+$ & 1.016e+06 $+$ & 1.463e+04 $+$ & 3.405e+04 $+$ & 1.221e+04 $+$ & \textbf{5.550e+03} \\
                ~ & std & 4.785e+04 & 4.678e+06 & 3.004e+05 & 1.186e+03 & 1.017e+04 & 1.072e+04 & 7.296e+04 & 1.025e+05 & 2.312e+06 & 9.023e+03 & 1.544e+04 & 6.507e+03 & 3.753e+03 \\
			\multirow{2}{*}{$f_7$} & mean & 1.509e+06 $+$ & 5.479e+07 $+$ & 4.159e+07 $+$ & 5.817e+05 $+$ & 9.624e+05 $+$ & 3.042e+06 $+$ & 1.586e+06 $+$ & 4.039e+06 $+$ & 2.252e+07 $+$ & 8.323e+05 $+$ & 9.515e+05 $+$ & 6.595e+05 $+$ & \textbf{8.949e+04} \\
                ~ & std & 7.894e+05 & 5.888e+07 & 1.171e+07 & 1.319e+05 & 5.147e+05 & 1.489e+06 & 1.186e+06 & 3.539e+06 & 6.035e+07 & 3.666e+05 & 5.809e+05 & 3.979e+05 & 5.118e+04 \\
                \midrule
			\multirow{2}{*}{$f_8$} & mean & 2.403e+03 $-$ & 2.726e+03 $+$ & 2.652e+03 $+$ & 2.455e+03 $+$ & \textbf{2.381e+03} $-$ & 2.476e+03 $+$ & 3.332e+03 $+$ & 3.355e+03 $+$ & 2.917e+03 $+$ & 2.590e+03 $+$ & 2.384e+03 $-$ & 2.383e+03 $-$ & 2.434e+03 \\
                ~ & std & 7.781e+00 & 1.867e+02 & 2.917e+01 & 9.179e+00 & 9.729e+00 & 1.069e+01 & 5.418e+02 & 4.176e+02 & 3.068e+02 & 9.884e+01 & 8.445e+00 & 7.392e+00 & 2.483e+01 \\
			\multirow{2}{*}{$f_9$} & mean & 6.172e+03 $+$ & 1.041e+04 $+$ & 1.316e+04 $+$ & 6.313e+03 $+$ & 3.262e+03 $+$ & 7.041e+03 $+$ & 1.166e+04 $+$ & 5.552e+03 $+$ & 7.449e+03 $+$ & 8.794e+03 $+$ & 2.752e+03 $+$ & 2.642e+03 $+$ & \textbf{2.624e+03} \\
                ~ & std & 2.824e+02 & 3.823e+03 & 4.679e+02 & 3.331e+02 & 5.494e+02 & 3.491e+02 & 5.863e+03 & 1.118e+03 & 1.959e+03 & 2.512e+03 & 3.932e+01 & 5.286e+01 & 1.028e+02 \\
			\multirow{2}{*}{$f_{10}$} & mean & 3.029e+03 $+$ & 3.627e+03 $+$ & 5.041e+03 $+$ & 3.100e+03 $+$ & 2.969e+03 $+$ & 3.189e+03 $+$ & 3.303e+03 $+$ & 3.232e+03 $+$ & 3.203e+03 $+$ & 3.163e+03 $+$ & \textbf{2.927e+03} $-$ & 2.937e+03 $\approx$ & 2.950e+03 \\
                ~ & std & 1.885e+01 & 5.366e+02 & 2.991e+02 & 3.047e+01 & 3.408e+01 & 5.356e+01 & 1.554e+02 & 1.181e+02 & 1.197e+02 & 1.800e+02 & 5.012e+00 & 1.119e+01 & 2.240e+01 \\
                \midrule
			\multicolumn{2}{c}{$+$/$\approx$/$-$} & 9/0/1 & 10/0/0 & 10/0/0 & 9/1/0 & 8/1/1 & 10/0/0 & 10/0/0 & 10/0/0 & 10/0/0 & 10/0/0 & 7/0/3 & 7/2/1 & - \\
                \midrule
                \multicolumn{2}{c}{Avg. rank} & 6.0 & 11.7 & 12.1 & 5.5 & 4.1 & 7.7 & 10.8 & 8.2 & 9.9 & 7.6 & 3.3 & 2.2 & \textbf{1.9} \\
			\bottomrule
		\end{tabular}
	}
\end{sidewaystable}
\begin{sidewaystable}[htbp]
	\scriptsize
	\centering
	\renewcommand\arraystretch{1.5}
	\caption{Experimental results and statistical analyses on 50-D CEC2020 benchmark functions.}
	\label{tbl:5.2.8}
	\resizebox{\columnwidth}{!}{
		\begin{tabular}{ccccccccccccccccc}
			\toprule
			\multicolumn{2}{c}{Func.} & GA & PSO & DE & CMA-ES & GWO & SCA & WOA & HHO & CSA & HBA & RIME & MBGO & EMBGO \\
			\midrule
			\multirow{2}{*}{$f_1$} & mean & 1.690e+10 $+$ & 1.979e+10 $+$ & 1.017e+11 $+$ & 1.018e+10 $+$ & 2.250e+09 $+$ & 9.744e+09 $+$ & 2.048e+10 $+$ & 2.605e+09 $+$ & 8.152e+09 $+$ & 1.481e+10 $+$ & 1.501e+07 $\approx$ & \textbf{2.697e+06} $-$ & 8.450e+07 \\
                ~ & std & 1.782e+09 & 4.876e+09 & 9.010e+09 & 2.459e+09 & 1.263e+09 & 1.369e+09 & 8.090e+09 & 1.146e+09 & 2.686e+09 & 6.327e+09 & 5.375e+06 & 9.567e+05 & 1.869e+08 \\
                \midrule
			\multirow{2}{*}{$f_2$} & mean & 1.832e+12 $+$ & 2.482e+12 $+$ & 1.192e+13 $+$ & 1.339e+12 $+$ & 2.794e+11 $+$ & 1.242e+12 $+$ & 2.364e+12 $+$ & 3.905e+11 $+$ & 1.251e+12 $+$ & 1.899e+12 $+$ & 2.056e+09 $\approx$ & \textbf{4.492e+08} $-$ & 1.154e+10 \\
                ~ & std & 1.748e+11 & 6.330e+11 & 1.004e+12 & 3.446e+11 & 1.469e+11 & 1.565e+11 & 1.074e+12 & 1.337e+11 & 3.593e+11 & 7.235e+11 & 1.130e+09 & 2.740e+08 & 2.312e+10 \\
			\multirow{2}{*}{$f_3$} & mean & 6.120e+11 $+$ & 7.088e+11 $+$ & 4.321e+12 $+$ & 4.346e+11 $+$ & 1.092e+11 $+$ & 3.584e+11 $+$ & 6.332e+11 $+$ & 1.028e+11 $+$ & 3.332e+11 $+$ & 5.147e+11 $+$ & 5.210e+08 $\approx$ & \textbf{1.001e+08} $-$ & 2.282e+09 \\
                ~ & std & 5.464e+10 & 2.018e+11 & 3.090e+11 & 1.293e+11 & 5.545e+10 & 5.674e+10 & 2.142e+11 & 3.790e+10 & 1.274e+11 & 1.944e+11 & 1.708e+08 & 4.421e+07 & 4.519e+09 \\
			\multirow{2}{*}{$f_4$} & mean & 5.343e+03 $+$ & 7.472e+04 $+$ & 1.488e+06 $+$ & 7.536e+03 $+$ & 1.959e+03 $-$ & 4.729e+03 $+$ & 3.380e+04 $+$ & 3.121e+03 $+$ & 9.057e+03 $+$ & 5.694e+03 $+$ & \textbf{1.931e+03} $-$ & 1.946e+03 $-$ & 2.093e+03 \\
                ~ & std & 1.222e+03 & 1.384e+05 & 6.471e+05 & 3.163e+03 & 4.349e+01 & 1.702e+03 & 4.676e+04 & 1.101e+03 & 7.256e+03 & 3.870e+03 & 5.593e+00 & 5.781e+00 & 1.772e+02 \\
                \midrule
			\multirow{2}{*}{$f_5$} & mean & 1.416e+07 $+$ & 7.795e+07 $+$ & 5.450e+07 $+$ & 2.120e+06 $+$ & 4.214e+06 $+$ & 7.419e+06 $+$ & 1.026e+07 $+$ & 1.302e+07 $+$ & 3.768e+07 $+$ & 3.744e+06 $+$ & 4.395e+06 $+$ & 2.673e+06 $+$ & \textbf{3.581e+05} \\
                ~ & std & 3.986e+06 & 5.727e+07 & 1.709e+07 & 3.948e+05 & 3.037e+06 & 2.884e+06 & 6.716e+06 & 6.501e+06 & 9.288e+07 & 2.366e+06 & 1.880e+06 & 1.798e+06 & 2.494e+05 \\
			\multirow{2}{*}{$f_6$} & mean & 8.729e+04 $+$ & 6.528e+06 $+$ & 9.271e+07 $+$ & 3.477e+04 $+$ & 5.629e+05 $+$ & 8.942e+05 $+$ & 6.003e+05 $+$ & 8.792e+05 $+$ & 3.112e+07 $+$ & 1.933e+05 $+$ & 6.292e+04 $+$ & 8.837e+03 $+$ & \textbf{5.455e+03} \\
                ~ & std & 5.375e+04 & 9.114e+06 & 2.877e+07 & 9.391e+03 & 8.440e+05 & 4.065e+05 & 1.075e+06 & 6.637e+05 & 8.364e+07 & 2.613e+05 & 2.799e+04 & 5.245e+03 & 3.949e+03 \\
			\multirow{2}{*}{$f_7$} & mean & 9.355e+07 $+$ & 7.203e+08 $+$ & 7.549e+08 $+$ & 1.211e+07 $+$ & 5.076e+06 $+$ & 3.439e+07 $+$ & 2.767e+07 $+$ & 4.058e+07 $+$ & 3.468e+08 $+$ & 6.826e+06 $+$ & 5.645e+06 $+$ & 3.242e+06 $+$ & \textbf{2.074e+05} \\
                ~ & std & 2.224e+07 & 6.680e+08 & 2.112e+08 & 3.095e+06 & 2.914e+06 & 1.118e+07 & 2.628e+07 & 2.716e+07 & 7.887e+08 & 4.685e+06 & 3.074e+06 & 2.169e+06 & 1.335e+05 \\
                \midrule
			\multirow{2}{*}{$f_8$} & mean & 2.682e+03 $\approx$ & 4.142e+03 $+$ & 3.192e+03 $+$ & 2.659e+03 $\approx$ & \textbf{2.453e+03} $-$ & 2.817e+03 $+$ & 1.019e+04 $+$ & 7.687e+03 $+$ & 7.620e+03 $+$ & 4.227e+03 $+$ & 2.466e+03 $-$ & 2.491e+03 $-$ & 2.698e+03 \\
                ~ & std & 2.058e+01 & 5.913e+02 & 8.934e+01 & 4.332e+01 & 2.033e+01 & 5.880e+01 & 2.367e+03 & 1.482e+03 & 2.649e+03 & 1.435e+03 & 1.702e+01 & 2.030e+01 & 1.269e+02 \\
			\multirow{2}{*}{$f_9$} & mean & 1.544e+04 $+$ & 2.273e+04 $+$ & 2.699e+04 $+$ & 1.141e+04 $+$ & 6.200e+03 $+$ & 1.423e+04 $+$ & 2.866e+04 $+$ & 9.570e+03 $+$ & 1.252e+04 $+$ & 2.176e+04 $+$ & 2.934e+03 $-$ & \textbf{2.715e+03} $-$ & 3.989e+03 \\
                ~ & std & 5.469e+02 & 3.618e+03 & 2.014e+03 & 1.403e+03 & 1.225e+03 & 1.011e+03 & 9.777e+03 & 3.714e+03 & 3.697e+03 & 7.779e+03 & 8.038e+01 & 1.336e+02 & 1.463e+03 \\
			\multirow{2}{*}{$f_{10}$} & mean & 5.064e+03 $+$ & 8.750e+03 $+$ & 1.225e+04 $+$ & 4.601e+03 $+$ & 3.828e+03 $+$ & 5.594e+03 $+$ & 5.592e+03 $+$ & 4.908e+03 $+$ & 5.247e+03 $+$ & 4.716e+03 $+$ & 3.462e+03 $\approx$ & \textbf{3.431e+03} $\approx$ & 3.455e+03 \\
                ~ & std & 2.193e+02 & 2.612e+03 & 1.641e+03 & 3.072e+02 & 2.097e+02 & 2.599e+02 & 9.995e+02 & 4.791e+02 & 7.436e+02 & 8.225e+02 & 1.579e+02 & 8.643e+01 & 1.348e+02 \\
                \midrule
			\multicolumn{2}{c}{$+$/$\approx$/$-$} & 9/1/0 & 10/0/0 & 10/0/0 & 9/1/0 & 8/0/2 & 10/0/0 & 10/0/0 & 10/0/0 & 10/0/0 & 10/0/0 & 3/4/3 & 3/1/6 & - \\
                \midrule
                \multicolumn{2}{c}{Avg. rank} & 8.3 & 11.5 & 12.3 & 5.9 & 4.0 & 7.7 & 10.4 & 7.0 & 9.0 & 7.7 & 2.8 & \textbf{1.7} & 2.7 \\
			\bottomrule
		\end{tabular}
	}
\end{sidewaystable}
\begin{sidewaystable}[htbp]
	\scriptsize
	\centering
	\renewcommand\arraystretch{1.5}
	\caption{Experimental results and statistical analyses on 10-D CEC2022 benchmark functions. $f_1$: Unimodal function; $f_2-f_5$: Basic functions; $f_6-f_8$: Hybrid functions; $f_9-f_{12}$: Composition functions.}
	\label{tbl:5.2.9}
	\resizebox{\columnwidth}{!}{
		\begin{tabular}{ccccccccccccccccc}
			\toprule
			\multicolumn{2}{c}{Func.} & GA & PSO & DE & CMA-ES & GWO & SCA & WOA & HHO & CSA & HBA & RIME & MBGO & EMBGO \\
			\midrule
			\multirow{2}{*}{$f_1$} & mean & 3.196e+03 $+$ & 1.906e+03 $+$ & 6.936e+03 $+$ & 4.382e+02 $+$ & 3.324e+02 $+$ & 6.376e+02 $+$ & 3.390e+03 $+$ & 1.071e+03 $+$ & 7.207e+02 $+$ & 3.025e+02 $+$ & 3.013e+02 $+$ & 3.661e+02 $+$ & \textbf{3.000e+02} \\
                ~ & std & 1.201e+03 & 1.437e+03 & 1.749e+03 & 5.239e+01 & 2.505e+01 & 1.472e+02 & 2.096e+03 & 4.834e+02 & 8.063e+02 & 4.576e+00 & 1.296e+00 & 5.994e+01 & 1.476e-05 \\
                \midrule
			\multirow{2}{*}{$f_2$} & mean & 4.686e+02 $+$ & 4.965e+02 $+$ & 4.576e+02 $+$ & 4.154e+02 $+$ & 4.177e+02 $+$ & 4.393e+02 $+$ & 4.859e+02 $+$ & 5.088e+02 $+$ & 5.422e+02 $+$ & 4.186e+02 $+$ & 4.127e+02 $+$ & 4.139e+02 $+$ & \textbf{4.070e+02} \\
                ~ & std & 1.478e+01 & 3.685e+01 & 1.232e+01 & 2.620e+00 & 2.033e+01 & 1.783e+01 & 8.800e+01 & 1.098e+02 & 1.575e+02 & 2.709e+01 & 1.904e+01 & 1.194e+01 & 1.665e+01 \\
			\multirow{2}{*}{$f_3$} & mean & 6.000e+02 $+$ & 6.000e+02 $+$ & 6.001e+02 $+$ & 6.000e+02 $+$ & 6.000e+02 $+$ & 6.000e+02 $+$ & 6.000e+02 $+$ & 6.000e+02 $+$ & 6.000e+02 $+$ & 6.000e+02 $+$ & 6.000e+02 $+$ & 6.000e+02 $+$ & \textbf{6.000e+02} \\
                ~ & std & 7.968e-03 & 5.018e-03 & 1.233e-02 & 6.888e-04 & 1.700e-03 & 2.521e-03 & 7.701e-03 & 3.751e-03 & 5.069e-02 & 2.403e-03 & 2.005e-05 & 2.096e-06 & 5.267e-13 \\
			\multirow{2}{*}{$f_4$} & mean & 8.010e+02 $+$ & 8.010e+02 $+$ & 8.011e+02 $+$ & 8.011e+02 $+$ & 8.005e+02 $\approx$ & 8.005e+02 $\approx$ & 8.005e+02 $\approx$ & 8.005e+02 $\approx$ & 8.006e+02 $+$ & 8.003e+02 $\approx$ & \textbf{8.003e+02} $\approx$ & 8.006e+02 $+$ & 8.004e+02 \\
                ~ & std & 1.527e-01 & 2.001e-01 & 2.034e-01 & 1.757e-01 & 3.635e-01 & 1.312e-01 & 3.282e-01 & 2.725e-01 & 4.954e-01 & 2.015e-01 & 1.343e-01 & 1.731e-01 & 1.445e-01 \\
			\multirow{2}{*}{$f_5$} & mean & 9.005e+02 $+$ & 9.009e+02 $+$ & 9.022e+02 $+$ & 9.004e+02 $+$ & 9.001e+02 $\approx$ & 9.001e+02 $+$ & 9.025e+02 $+$ & 9.022e+02 $+$ & 9.015e+02 $+$ & 9.008e+02 $+$ & 9.003e+02 $+$ & \textbf{9.000e+02} $\approx$ & 9.001e+02 \\
                ~ & std & 1.260e-01 & 6.853e-01 & 5.628e-01 & 7.714e-02 & 1.450e-01 & 6.395e-02 & 1.594e+00 & 1.413e+00 & 1.085e+00 & 9.263e-01 & 4.076e-01 & 1.288e-03 & 1.931e-01 \\
                \midrule
			\multirow{2}{*}{$f_6$} & mean & 1.264e+05 $+$ & 1.261e+06 $+$ & 1.045e+05 $+$ & \textbf{4.123e+03} $-$ & 3.848e+04 $+$ & 2.436e+05 $+$ & 2.685e+04 $+$ & 3.592e+04 $+$ & 2.686e+06 $+$ & 2.390e+04 $+$ & 2.881e+04 $+$ & 2.398e+04 $+$ & 1.113e+04 \\
                ~ & std & 2.025e+05 & 2.146e+06 & 3.778e+04 & 6.533e+02 & 1.925e+04 & 1.510e+05 & 1.062e+04 & 1.770e+04 & 1.425e+07 & 1.329e+04 & 1.547e+04 & 8.323e+03 & 6.294e+03 \\
			\multirow{2}{*}{$f_7$} & mean & 2.054e+03 $+$ & 2.132e+03 $+$ & 2.072e+03 $+$ & 2.077e+03 $+$ & 2.044e+03 $+$ & 2.069e+03 $+$ & 2.144e+03 $+$ & 2.151e+03 $+$ & 2.169e+03 $+$ & 2.044e+03 $+$ & \textbf{2.030e+03} $-$ & 2.038e+03 $+$ & 2.033e+03 \\
                ~ & std & 8.550e+00 & 6.052e+01 & 1.172e+01 & 1.623e+01 & 2.666e+01 & 1.585e+01 & 9.680e+01 & 8.695e+01 & 1.378e+02 & 1.869e+01 & 1.800e+01 & 5.734e+00 & 5.448e+00 \\
			\multirow{2}{*}{$f_8$} & mean & 2.398e+03 $+$ & 3.743e+03 $+$ & 2.255e+03 $+$ & 2.228e+03 $+$ & 2.498e+03 $+$ & 2.314e+03 $+$ & 3.041e+03 $+$ & 2.959e+03 $+$ & 3.617e+03 $+$ & 2.249e+03 $+$ & 2.226e+03 $\approx$ & 2.231e+03 $+$ & \textbf{2.225e+03} \\
                ~ & std & 2.488e+02 & 4.228e+03 & 7.304e+00 & 3.227e+00 & 5.435e+02 & 4.600e+01 & 7.416e+02 & 7.006e+02 & 1.319e+03 & 4.444e+01 & 4.927e+00 & 2.637e+00 & 8.642e-01 \\
                \midrule
			\multirow{2}{*}{$f_9$} & mean & 2.596e+03 $+$ & 2.657e+03 $+$ & 2.665e+03 $+$ & 2.582e+03 $+$ & 2.629e+03 $+$ & \textbf{2.346e+03} $-$ & 2.761e+03 $+$ & 2.755e+03 $+$ & 2.830e+03 $+$ & 2.583e+03 $+$ & 2.537e+03 $+$ & 2.348e+03 $-$ & 2.385e+03 \\
                ~ & std & 8.587e+01 & 1.633e+02 & 4.502e+01 & 1.306e+02 & 1.280e+02 & 3.395e+01 & 1.217e+02 & 1.135e+02 & 1.898e+02 & 1.550e+02 & 1.802e+02 & 1.080e+02 & 1.539e+02 \\
			\multirow{2}{*}{$f_{10}$} & mean & 2.618e+03 $+$ & 2.618e+03 $+$ & 2.623e+03 $+$ & 2.606e+03 $\approx$ & 2.627e+03 $+$ & 2.605e+03 $\approx$ & 2.671e+03 $+$ & 2.658e+03 $+$ & 2.671e+03 $+$ & 2.647e+03 $+$ & 2.619e+03 $\approx$ & \textbf{2.601e+03} $\approx$ & 2.608e+03 \\
                ~ & std & 3.251e+01 & 9.079e+00 & 4.233e+00 & 8.057e-01 & 5.505e+01 & 1.092e+00 & 1.191e+02 & 7.482e+01 & 7.931e+01 & 6.642e+01 & 4.477e+01 & 1.647e+00 & 2.616e+01 \\
			\multirow{2}{*}{$f_{11}$} & mean & 2.626e+03 $+$ & 2.623e+03 $+$ & 2.661e+03 $+$ & 2.652e+03 $+$ & 2.602e+03 $+$ & 2.608e+03 $+$ & 2.646e+03 $+$ & 2.654e+03 $+$ & 2.768e+03 $+$ & 2.660e+03 $+$ & 2.630e+03 $+$ & 2.601e+03 $+$ & \textbf{2.600e+03} \\
                ~ & std & 4.712e+00 & 1.356e+01 & 1.513e+02 & 1.747e+02 & 1.930e+00 & 2.696e+00 & 1.594e+02 & 1.693e+02 & 3.177e+02 & 2.183e+02 & 1.573e+02 & 4.183e-01 & 1.171e-04 \\
			\multirow{2}{*}{$f_{12}$} & mean & 2.878e+03 $+$ & 2.905e+03 $+$ & 2.868e+03 $+$ & 2.867e+03 $\approx$ & \textbf{2.866e+03} $\approx$ & 2.889e+03 $+$ & 2.915e+03 $+$ & 2.924e+03 $+$ & 2.908e+03 $+$ & 2.869e+03 $\approx$ & 2.866e+03 $\approx$ & 2.867e+03 $\approx$ & 2.867e+03 \\
                ~ & std & 2.012e+00 & 1.978e+01 & 7.310e-01 & 5.121e-01 & 9.189e-01 & 5.036e+00 & 6.120e+01 & 4.391e+01 & 3.815e+01 & 5.173e+00 & 1.171e+00 & 7.432e-01 & 1.512e+00 \\
			\midrule
			\multicolumn{2}{c}{$+$/$\approx$/$-$} & 12/0/0 & 12/0/0 & 12/0/0 & 9/2/1 & 9/3/0 & 9/2/1 & 11/1/0 & 11/1/0 & 12/0/0 & 10/2/0 & 7/4/1 & 8/3/1 & - \\
                \midrule
                \multicolumn{2}{c}{Avg. rank} & 8.3 & 9.5 & 9.8 & 5.7 & 5.3 & 6.1 & 9.9 & 10.0 & 11.5 & 5.8 & 3.5 & 3.2 & \textbf{2.1} \\
			\bottomrule
		\end{tabular}
	}
\end{sidewaystable}
\begin{sidewaystable}[htbp]
	\scriptsize
	\centering
	\renewcommand\arraystretch{1.5}
	\caption{Experimental results and statistical analyses on 20-D CEC2022 benchmark functions.}
	\label{tbl:5.2.10}
	\resizebox{\columnwidth}{!}{
		\begin{tabular}{ccccccccccccccccc}
			\toprule
			\multicolumn{2}{c}{Func.} & GA & PSO & DE & CMA-ES & GWO & SCA & WOA & HHO & CSA & HBA & RIME & MBGO & EMBGO \\
			\midrule
			\multirow{2}{*}{$f_1$} & mean & 2.271e+04 $+$ & 1.298e+04 $+$ & 3.428e+04 $+$ & 2.593e+03 $+$ & 5.538e+02 $+$ & 4.619e+03 $+$ & 6.206e+03 $+$ & 3.766e+03 $+$ & 3.665e+03 $+$ & 7.420e+02 $+$ & 3.040e+02 $+$ & 8.572e+02 $+$ & \textbf{3.000e+02} \\
                ~ & std & 3.653e+03 & 5.865e+03 & 4.837e+03 & 4.391e+02 & 1.777e+02 & 1.128e+03 & 3.686e+03 & 1.849e+03 & 2.452e+03 & 8.288e+02 & 1.996e+00 & 2.629e+02 & 1.278e-03 \\
                \midrule
			\multirow{2}{*}{$f_2$} & mean & 8.726e+02 $+$ & 7.208e+02 $+$ & 1.047e+03 $+$ & 5.167e+02 $+$ & 4.771e+02 $+$ & 5.583e+02 $+$ & 6.172e+02 $+$ & 6.044e+02 $+$ & 6.122e+02 $+$ & 4.977e+02 $+$ & 4.577e+02 $\approx$ & 4.677e+02 $+$ & \textbf{4.569e+02} \\
                ~ & std & 8.414e+01 & 1.628e+02 & 1.361e+02 & 1.523e+01 & 2.281e+01 & 2.533e+01 & 8.449e+01 & 5.545e+01 & 1.313e+02 & 4.216e+01 & 1.308e+01 & 1.834e+01 & 1.678e+01 \\
			\multirow{2}{*}{$f_3$} & mean & 6.004e+02 $+$ & 6.001e+02 $+$ & 6.004e+02 $+$ & 6.000e+02 $+$ & 6.000e+02 $+$ & 6.001e+02 $+$ & 6.001e+02 $+$ & 6.000e+02 $+$ & 6.001e+02 $+$ & 6.001e+02 $+$ & 6.000e+02 $+$ & 6.000e+02 $+$ & \textbf{6.000e+02} \\
                ~ & std & 4.878e-02 & 1.075e-01 & 6.400e-02 & 5.566e-03 & 4.784e-03 & 1.222e-02 & 7.576e-02 & 2.086e-02 & 1.169e-01 & 1.044e-01 & 2.819e-05 & 1.130e-06 & 7.106e-10 \\

			\multirow{2}{*}{$f_4$} & mean & 8.038e+02 $+$ & 8.037e+02 $+$ & 8.041e+02 $+$ & 8.039e+02 $+$ & 8.015e+02 $\approx$ & 8.021e+02 $+$ & 8.014e+02 $\approx$ & 8.013e+02 $\approx$ & 8.018e+02 $+$ & 8.011e+02 $\approx$ & \textbf{8.008e+02} $\approx$ & 8.028e+02 $+$ & 8.010e+02 \\
                ~ & std & 2.737e-01 & 4.074e-01 & 3.267e-01 & 4.267e-01 & 1.209e+00 & 2.875e-01 & 6.313e-01 & 5.388e-01 & 1.102e+00 & 3.880e-01 & 3.458e-01 & 3.045e-01 & 4.190e-01 \\
			\multirow{2}{*}{$f_5$} & mean & 9.051e+02 $+$ & 9.049e+02 $+$ & 9.150e+02 $+$ & 9.032e+02 $+$ & 9.006e+02 $\approx$ & 9.008e+02 $\approx$ & 9.083e+02 $+$ & 9.025e+02 $\approx$ & 9.050e+02 $+$ & 9.034e+02 $+$ & 9.018e+02 $\approx$ & \textbf{9.000e+02} $-$ & 9.014e+02 \\
                ~ & std & 7.761e-01 & 3.275e+00 & 2.456e+00 & 5.982e-01 & 6.392e-01 & 3.187e-01 & 3.529e+00 & 1.658e+00 & 2.779e+00 & 2.419e+00 & 1.686e+00 & 8.438e-02 & 9.215e-01 \\
                \midrule
			\multirow{2}{*}{$f_6$} & mean & 3.330e+08 $+$ & 2.626e+08 $+$ & 2.185e+08 $+$ & 4.948e+06 $+$ & 9.516e+04 $+$ & 1.187e+07 $+$ & 8.644e+04 $+$ & 2.649e+05 $+$ & 1.012e+08 $+$ & 1.569e+05 $+$ & 7.779e+04 $+$ & 8.551e+04 $+$ & \textbf{3.171e+04} \\
                ~ & std & 1.113e+08 & 3.635e+08 & 7.114e+07 & 1.895e+06 & 9.054e+04 & 6.503e+06 & 4.205e+04 & 2.763e+05 & 4.187e+08 & 8.517e+04 & 2.872e+04 & 3.343e+04 & 1.162e+04 \\
			\multirow{2}{*}{$f_7$} & mean & 2.758e+03 $+$ & 2.938e+03 $+$ & 2.819e+03 $+$ & 2.306e+03 $+$ & 2.079e+03 $\approx$ & 2.237e+03 $+$ & 3.285e+03 $+$ & 3.023e+03 $+$ & 3.105e+03 $+$ & 2.205e+03 $+$ & 2.062e+03 $\approx$ & \textbf{2.058e+03} $\approx$ & 2.063e+03 \\
                ~ & std & 1.646e+02 & 5.545e+02 & 2.297e+02 & 6.084e+01 & 3.673e+01 & 9.081e+01 & 6.353e+02 & 5.086e+02 & 9.258e+02 & 1.134e+02 & 1.641e+01 & 1.603e+01 & 2.760e+01 \\
			\multirow{2}{*}{$f_8$} & mean & 1.690e+06 $+$ & 4.229e+07 $+$ & 1.113e+05 $+$ & 2.415e+03 $+$ & 4.751e+03 $+$ & 4.080e+03 $+$ & 5.794e+03 $+$ & 5.359e+03 $+$ & 1.616e+11 $+$ & 3.307e+03 $+$ & 4.102e+03 $+$ & 2.934e+03 $+$ & \textbf{2.266e+03} \\
                ~ & std & 4.982e+06 & 1.097e+08 & 2.665e+05 & 7.471e+01 & 1.244e+03 & 1.028e+03 & 2.452e+03 & 1.756e+03 & 8.704e+11 & 8.471e+02 & 1.101e+03 & 4.198e+02 & 5.021e+01 \\
                \midrule
			\multirow{2}{*}{$f_9$} & mean & 3.039e+03 $+$ & 3.207e+03 $+$ & 2.782e+03 $+$ & 2.653e+03 $+$ & 2.665e+03 $+$ & 2.832e+03 $+$ & 2.793e+03 $+$ & 2.947e+03 $+$ & 3.515e+03 $+$ & 2.652e+03 $+$ & 2.649e+03 $+$ & 2.662e+03 $+$ & \textbf{2.637e+03} \\
                ~ & std & 5.120e+01 & 2.208e+02 & 5.008e+01 & 6.486e+00 & 2.084e+01 & 3.968e+01 & 1.537e+02 & 1.675e+02 & 8.150e+02 & 2.245e+01 & 6.778e+00 & 1.548e+01 & 1.212e+00 \\
			\multirow{2}{*}{$f_{10}$} & mean & 2.942e+03 $+$ & 2.832e+03 $\approx$ & 3.594e+03 $+$ & 3.148e+03 $+$ & 3.562e+03 $+$ & 2.792e+03 $\approx$ & 4.564e+03 $+$ & 4.634e+03 $+$ & 4.737e+03 $+$ & 3.301e+03 $+$ & 3.223e+03 $+$ & \textbf{2.770e+03} $\approx$ & 2.796e+03 \\
                ~ & std & 1.305e+02 & 3.482e+01 & 1.362e+03 & 8.562e+02 & 1.136e+03 & 3.555e+01 & 1.210e+03 & 1.286e+03 & 1.558e+03 & 9.018e+02 & 7.964e+02 & 3.200e+01 & 5.978e+01 \\
			\multirow{2}{*}{$f_{11}$} & mean & 2.771e+03 $+$ & 2.801e+03 $+$ & 2.732e+03 $+$ & 2.621e+03 $\approx$ & 2.611e+03 $\approx$ & 2.629e+03 $\approx$ & 2.969e+03 $+$ & 2.993e+03 $+$ & 3.648e+03 $+$ & 2.659e+03 $\approx$ & \textbf{2.603e+03} $-$ & 2.628e+03 $\approx$ & 2.648e+03 \\
                ~ & std & 4.293e+01 & 3.397e+02 & 2.868e+01 & 2.187e+00 & 1.004e+01 & 4.521e+00 & 5.820e+02 & 6.397e+02 & 1.556e+03 & 6.037e+01 & 5.887e+00 & 7.805e+00 & 2.482e+02 \\
			\multirow{2}{*}{$f_{12}$} & mean & 3.101e+03 $+$ & 3.238e+03 $+$ & 2.956e+03 $-$ & \textbf{2.954e+03} $-$ & 2.964e+03 $-$ & 3.128e+03 $+$ & 3.175e+03 $+$ & 3.304e+03 $+$ & 3.199e+03 $+$ & 3.030e+03 $+$ & 2.960e+03 $-$ & 2.957e+03 $-$ & 2.992e+03 \\
                ~ & std & 2.046e+01 & 9.552e+01 & 5.356e+00 & 6.520e+00 & 2.026e+01 & 2.747e+01 & 1.454e+02 & 1.427e+02 & 1.251e+02 & 5.742e+01 & 1.586e+01 & 1.131e+01 & 3.492e+01 \\
			\midrule
			\multicolumn{2}{c}{$+$/$\approx$/$-$} & 12/0/0 & 11/1/0 & 11/0/1 & 10/1/1 & 7/4/1 & 9/3/0 & 11/1/0 & 10/2/0 & 12/0/0 & 10/2/0 & 6/4/2 & 7/3/2 & - \\
                \midrule
                \multicolumn{2}{c}{Avg. rank} & 10.3 & 10.3 & 10.0 & 5.6 & 4.8 & 6.5 & 9.3 & 8.7 & 10.6 & 5.6 & 3.1 & 3.2 & \textbf{2.5} \\
			\bottomrule
		\end{tabular}
	}
\end{sidewaystable}


\subsection{Experiments on engineering problems} \label{sec:5.3}
This section introduces the numerical experiments conducted on engineering optimization problems. In the following context, we sequentially present the information on engineering problems, competitor algorithms and parameters, and experimental results obtained on engineering problems.

\textbf{Information on engineering problems}: We present eight engineering optimization problems in Table \ref{tbl:5.3.1}. Due to the space limitation, the detailed mathematical model and visualized demonstration can be found in \cite{Hadi:21}.
\begin{table}[htbp]
	\scriptsize
	\centering
	\renewcommand\arraystretch{1.3}
	\caption{Summary of eight engineering optimization problems.}
	\label{tbl:5.3.1}
	\begin{tabular}{cccc}
		\toprule
		Name & Abbr. & Dim. & \# of constraints \\
		\midrule
            Cantilever Beam Problem & CBD & 5 & 1 \\
            Corrugated Bulkhead Problem & CBHD & 4 & 6 \\
            I Beam Problem & IBD & 4 & 2 \\
            Piston Lever Problem & PLD & 4 & 4 \\
            Speed Reducer Problem & SRD & 7 & 11 \\
            Three Bar Truss Design Problem & TBTD & 2 & 3 \\
            Tubular Column Problem & TCD & 2 & 6 \\
            Welded Beam Problem & WBP & 4 & 7 \\
		\bottomrule
	\end{tabular}
\end{table}

\textbf{Competitor algorithms and parameters}: Numerical experiments on engineering problems inherit experimental settings from experiments on CEC benchmark functions, with the only difference is the maximum FEs are fixed at 10,000. Moreover, recognizing that engineering problems often involve constraints and primary MAs cannot handle constrained optimization problems. Therefore, all MA approaches are equipped with the static penalty function in this case, as defined by Eq. (\ref{eq:5.3.1}).
\begin{equation}
    \label{eq:5.3.1}
    \begin{aligned}
        F(X_{i}) = f(X_{i}) + w\cdot \sum^m_{i=1}(\max(0, g_i(X_{i})))
    \end{aligned}
\end{equation}
where $F(\cdot)$ is the fitness function, $f(\cdot)$ is the objective function, and $g_i(\cdot)$ is the constraint function. $w$ is a constant set to $10e7$ by default in the ENOPPY library.

\begin{figure}[htbp]
    \centering
    \includegraphics[width=16cm]{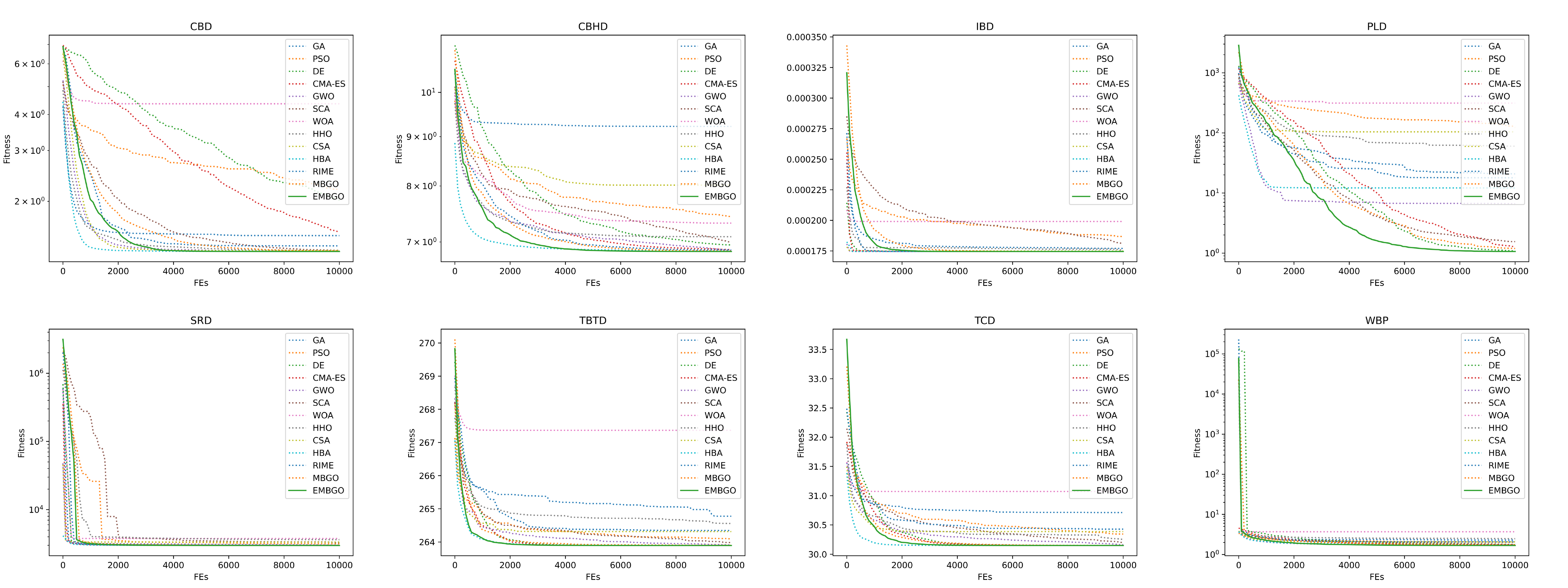}
    \caption{Convergence curves of thirteen MA approaches on engineering problems.}
    \label{fig:5.3.1}
\end{figure}

\textbf{Experimental results on engineering problems}: The experimental results and statistical analyses on engineering problems are presented in Table \ref{tbl:5.3.1}, and the Holm multiple comparison test is also employed to determine the significance. Convergence curves are provided in Figure \ref{fig:5.3.1} to visualize the convergence tendency during optimization.
\begin{sidewaystable}[htbp]
	\scriptsize
	\centering
	\renewcommand\arraystretch{1.5}
	\caption{Experimental results and statistical analyses on engineering problems. best and worst: the best and worst fitness value among 30 trial runs.}
	\label{tbl:5.3.2}
	\resizebox{\columnwidth}{!}{
		\begin{tabular}{ccccccccccccccccc}
			\toprule
			\multicolumn{2}{c}{Prob.} & GA & PSO & DE & CMA-ES & GWO & SCA & WOA & HHO & CSA & HBA & RIME & MBGO & EMBGO \\
			\midrule
			\multirow{4}{*}{CBD} & mean & 1.520e+00 $+$ & 2.204e+00 $+$ & 2.015e+00 $+$ & 1.559e+00 $+$ & 1.340e+00 $+$ & 1.353e+00 $+$ & 4.361e+00 $+$ & 1.346e+00 $+$ & 1.350e+00 $+$ & 1.341e+00 $+$ & 1.401e+00 $+$ & 1.346e+00 $+$ & \textbf{1.340e+00} \\
                ~ & std & 1.299e-01 & 2.871e-01 & 2.139e-01 & 8.311e-02 & 1.593e-04 & 5.027e-03 & 1.323e+00 & 4.421e-03 & 1.868e-02 & 5.765e-04 & 4.339e-02 & 3.213e-03 & 2.196e-05 \\
                ~ & best & 1.471e+00 & 2.329e+00 & 1.705e+00 & 1.531e+00 & 1.340e+00 & 1.360e+00 & 5.462e+00 & 1.345e+00 & 1.349e+00 & 1.342e+00 & 1.363e+00 & 1.341e+00 & 1.340e+00 \\
                ~ & worst & 4.504e+00 & 6.729e+00 & 6.729e+00 & 6.729e+00 & 5.370e+00 & 7.067e+00 & 6.318e+00 & 5.258e+00 & 6.128e+00 & 5.164e+00 & 6.729e+00 & 5.421e+00 & 4.850e+00 \\
                \midrule
			\multirow{4}{*}{CBHD} & mean & 9.222e+00 $+$ & 7.439e+00 $+$ & 6.949e+00 $+$ & 6.874e+00 $+$ & 6.862e+00 $+$ & 7.003e+00 $+$ & 7.322e+00 $+$ & 7.088e+00 $+$ & 8.018e+00 $+$ & 6.852e+00 $+$ & 6.868e+00 $+$ & 6.849e+00 $+$ & \textbf{6.843e+00} \\
                ~ & std & 1.082e+00 & 2.430e-01 & 3.671e-02 & 9.187e-03 & 8.570e-03 & 4.093e-02 & 4.652e-01 & 2.573e-01 & 1.326e+00 & 1.331e-02 & 1.467e-02 & 3.435e-03 & 1.963e-04 \\
                ~ & best & 1.021e+01 & 7.667e+00 & 6.950e+00 & 6.860e+00 & 6.882e+00 & 7.017e+00 & 7.984e+00 & 6.907e+00 & 6.899e+00 & 6.847e+00 & 6.860e+00 & 6.849e+00 & 6.843e+00 \\
                ~ & worst & 1.169e+01 & 1.078e+01 & 1.323e+01 & 1.359e+01 & 9.596e+00 & 9.167e+00 & 9.434e+00 & 8.720e+00 & 9.650e+00 & 8.120e+00 & 1.137e+01 & 9.878e+00 & 1.097e+01 \\
                \midrule
			\multirow{4}{*}{IBD} & mean & 1.771e-04 $+$ & 1.868e-04 $+$ & 1.746e-04 $+$ & 1.746e-04 $-$ & 1.746e-04 $+$ & 1.814e-04 $+$ & 1.990e-04 $+$ & 1.758e-04 $+$ & 1.749e-04 $+$ & \textbf{1.746e-04} $-$ & 1.746e-04 $+$ & 1.746e-04 $+$ & 1.746e-04 \\
                ~ & std & 3.638e-06 & 4.466e-06 & 2.661e-11 & 1.680e-14 & 4.757e-10 & 1.852e-06 & 5.129e-05 & 6.047e-06 & 6.428e-07 & 1.426e-14 & 4.345e-09 & 2.005e-09 & 1.388e-12 \\
                ~ & best & 1.754e-04 & 1.798e-04 & 1.746e-04 & 1.746e-04 & 1.746e-04 & 1.809e-04 & 1.746e-04 & 1.746e-04 & 1.746e-04 & 1.746e-04 & 1.746e-04 & 1.746e-04 & 1.746e-04 \\
                ~ & worst & 2.533e-04 & 3.412e-04 & 1.843e-04 & 1.804e-04 & 1.907e-04 & 2.561e-04 & 1.933e-04 & 2.420e-04 & 1.750e-04 & 1.748e-04 & 2.057e-04 & 2.784e-04 & 5.108e-04 \\
                \midrule
			\multirow{4}{*}{PLD} & mean & 2.079e+01 $+$ & 1.234e+02 $+$ & 1.087e+00 $+$ & 1.281e+00 $+$ & 6.626e+00 $+$ & 1.555e+00 $+$ & 3.134e+02 $+$ & 6.022e+01 $+$ & 1.042e+02 $+$ & 1.215e+01 $+$ & 1.794e+01 $+$ & 1.197e+00 $+$ & \textbf{1.066e+00} \\
                ~ & std & 4.439e+01 & 6.073e+01 & 2.028e-02 & 1.509e-01 & 2.996e+01 & 2.230e-01 & 1.796e+02 & 1.049e+02 & 1.807e+02 & 4.151e+01 & 6.269e+01 & 3.245e-01 & 1.661e-02 \\
                ~ & best & 5.276e+00 & 1.680e+02 & 1.125e+00 & 1.271e+00 & 1.063e+00 & 1.967e+00 & 1.559e+02 & 6.193e+00 & 1.073e+00 & 1.057e+00 & 1.063e+00 & 1.134e+00 & 1.058e+00 \\
                ~ & worst & 2.695e+02 & 3.975e+02 & 1.154e+03 & 1.154e+03 & 1.154e+03 & 4.541e+02 & 1.154e+03 & 1.154e+03 & 2.995e+02 & 9.895e+01 & 1.154e+03 & 1.107e+03 & 8.606e+02 \\
                \midrule
			\multirow{4}{*}{SRD} & mean & 2.995e+03 $+$ & 3.156e+03 $+$ & 2.991e+03 $+$ & 2.990e+03 $+$ & 3.011e+03 $+$ & 3.269e+03 $+$ & 3.785e+03 $+$ & 3.625e+03 $+$ & 3.350e+03 $+$ & 2.991e+03 $+$ & 3.003e+03 $+$ & 2.993e+03 $+$ & \textbf{2.987e+03} \\
                ~ & std & 4.492e+00 & 5.313e+01 & 1.774e+00 & 9.769e-01 & 4.713e+00 & 1.115e+02 & 8.245e+02 & 5.926e+02 & 6.739e+02 & 2.263e+00 & 8.867e+00 & 1.703e+00 & 2.018e-01 \\
                ~ & best & 2.991e+03 & 3.096e+03 & 2.992e+03 & 2.990e+03 & 3.012e+03 & 3.301e+03 & 4.199e+03 & 3.683e+03 & 3.181e+03 & 2.995e+03 & 2.989e+03 & 2.993e+03 & 2.987e+03 \\
                ~ & worst & 1.062e+06 & 8.241e+05 & 3.209e+03 & 4.387e+03 & 3.364e+03 & 6.013e+06 & 4.810e+03 & 5.067e+03 & 3.239e+03 & 4.082e+03 & 1.938e+06 & 6.081e+03 & 2.800e+06 \\
                \midrule
			\multirow{4}{*}{TBTD} & mean & 2.648e+02 $+$ & 2.641e+02 $+$ & 2.639e+02 $-$ & \textbf{2.639e+02} $-$ & 2.639e+02 $+$ & 2.640e+02 $+$ & 2.674e+02 $+$ & 2.646e+02 $+$ & 2.643e+02 $+$ & 2.639e+02 $+$ & 2.643e+02 $+$ & 2.639e+02 $+$ & 2.639e+02 \\
                ~ & std & 8.837e-01 & 1.324e-01 & 7.558e-06 & 1.576e-06 & 1.122e-02 & 5.808e-02 & 3.615e+00 & 1.032e+00 & 7.523e-01 & 3.807e-03 & 5.532e-01 & 4.715e-03 & 4.443e-05 \\
                ~ & best & 2.657e+02 & 2.640e+02 & 2.639e+02 & 2.639e+02 & 2.639e+02 & 2.640e+02 & 2.686e+02 & 2.680e+02 & 2.678e+02 & 2.639e+02 & 2.640e+02 & 2.639e+02 & 2.639e+02 \\
                ~ & worst & 2.691e+02 & 2.668e+02 & 2.718e+02 & 2.718e+02 & 2.709e+02 & 2.703e+02 & 2.692e+02 & 2.691e+02 & 2.711e+02 & 2.670e+02 & 2.651e+02 & 2.642e+02 & 2.718e+02 \\
                \midrule
			\multirow{4}{*}{TCD} & mean & 3.071e+01 $+$ & 3.035e+01 $+$ & 3.015e+01 $+$ & 3.015e+01 $+$ & 3.016e+01 $+$ & 3.020e+01 $+$ & 3.107e+01 $+$ & 3.026e+01 $+$ & 3.039e+01 $+$ & \textbf{3.015e+01} $-$ & 3.043e+01 $+$ & 3.015e+01 $+$ & 3.015e+01 \\
                ~ & std & 7.337e-01 & 9.032e-02 & 8.503e-05 & 3.461e-05 & 5.315e-03 & 3.227e-02 & 1.262e+00 & 1.132e-01 & 6.155e-01 & 5.423e-07 & 3.068e-01 & 7.819e-04 & 1.627e-05 \\
                ~ & best & 3.026e+01 & 3.034e+01 & 3.015e+01 & 3.015e+01 & 3.016e+01 & 3.017e+01 & 3.029e+01 & 3.017e+01 & 3.017e+01 & 3.015e+01 & 3.037e+01 & 3.015e+01 & 3.015e+01 \\
                ~ & worst & 3.089e+01 & 3.199e+01 & 3.156e+01 & 3.156e+01 & 3.156e+01 & 3.064e+01 & 3.065e+01 & 3.113e+01 & 3.143e+01 & 3.147e+01 & 3.156e+01 & 3.139e+01 & 3.294e+01 \\
                \midrule
			\multirow{4}{*}{WBP} & mean & 2.295e+00 $+$ & 2.031e+00 $+$ & 1.761e+00 $+$ & 1.713e+00 $+$ & 1.692e+00 $+$ & 1.796e+00 $+$ & 3.695e+00 $+$ & 2.486e+00 $+$ & 2.069e+00 $+$ & 1.726e+00 $+$ & 2.070e+00 $+$ & 1.753e+00 $+$ & \textbf{1.685e+00} \\
                ~ & std & 5.252e-01 & 1.454e-01 & 2.811e-02 & 1.094e-02 & 4.982e-03 & 3.893e-02 & 1.739e+00 & 4.820e-01 & 5.345e-01 & 1.293e-01 & 3.723e-01 & 4.712e-02 & 1.445e-03 \\
                ~ & best & 2.288e+00 & 2.189e+00 & 1.764e+00 & 1.706e+00 & 1.699e+00 & 1.832e+00 & 3.652e+00 & 2.990e+00 & 2.566e+00 & 1.711e+00 & 1.876e+00 & 1.752e+00 & 1.684e+00 \\
                ~ & worst & 3.653e+00 & 3.239e+00 & 3.653e+00 & 3.653e+00 & 3.288e+00 & 4.715e+00 & 3.653e+00 & 3.653e+00 & 3.653e+00 & 3.653e+00 & 3.653e+00 & 6.741e+05 & 2.613e+00 \\
			\midrule
			\multicolumn{2}{c}{$+$/$\approx$/$-$} & 8/0/0 & 8/0/0 & 7/0/1 & 6/0/2 & 8/0/0 & 8/0/0 & 8/0/0 & 8/0/0 & 8/0/0 & 6/0/2 & 8/0/0 & 8/0/0 & - \\
                \midrule
                \multicolumn{2}{c}{Avg. rank} & 10.2 & 10.1 & 5.0 & 3.8 & 4.8 & 7.7 & 12.6 & 9.5 & 9.5 & 3.2 & 8.1 & 4.5 & \textbf{1.6} \\
			\bottomrule
		\end{tabular}
	}
\end{sidewaystable}

\subsection{Experiments on ARNAS tasks} \label{sec:5.4}
This section introduces the numerical experiments conducted on ARNAS tasks. In the following context, we sequentially present the information on the ARNAS benchmark, competitor algorithms and parameters, and experimental results obtained on ARNAS tasks.

\begin{figure}[htbp]
    \centering
    \includegraphics[width=13cm]{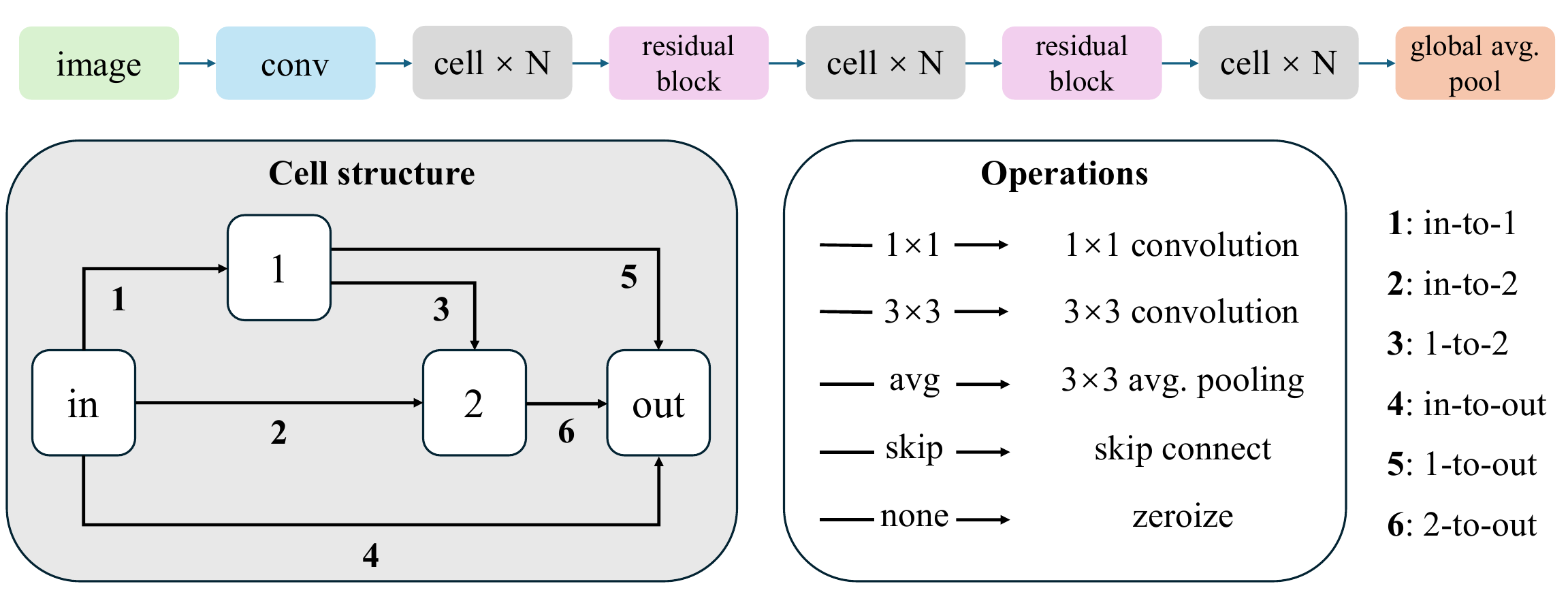}
    \caption{The ARNAS search space. A cell in this context comprises four nodes (representing feature maps) and six edges (representing possible operations) connecting them. The set of permissible operations encompasses 1x1/3x3 convolutions, 3x3 average pooling, skip connections, and zeroize (dropping the edge). As a result, the search space comprises 5$^6$=15,625 potential architectures, among which 6,466 are non-isomorphic \cite{Dong:22, Jung:23}.}
    \label{fig:5.4.1}
\end{figure}

\textbf{Information on the ARNAS benchmark}: NAS-Bench-201 based ARNAS benchmark suite is demonstrated in Figure \ref{fig:5.4.1}, and optima are collected in Table \ref{tbl:5.4.1}.
\begin{table}[htbp]
	\scriptsize
	\centering
	\renewcommand\arraystretch{1.3}
	\caption{Summary of the optimal accuracy in NAS-Bench-201 based ARNAS benchmark suite.}
	\label{tbl:5.4.1}
	\begin{tabular}{cccc}
		\toprule
		Dataset & Attack method & Optimum  \\
		\midrule
            \multirow{5}{*}{CIFAR-10} & Clean & 94.6 \\
            ~ & FGSM & 69.2 \\
            ~ & PGD & 58.8 \\
            ~ & APGD & 54.0 \\
            ~ & Square & 73.6 \\
		\midrule
            \multirow{5}{*}{CIFAR-100} & Clean & 73.6 \\
            ~ & FGSM & 29.4 \\
            ~ & PGD & 29.8 \\
            ~ & APGD & 26.3 \\
            ~ & Square & 40.4 \\
		\bottomrule
	\end{tabular}
\end{table}

\begin{figure}[htbp]
    \centering
    \includegraphics[width=13cm]{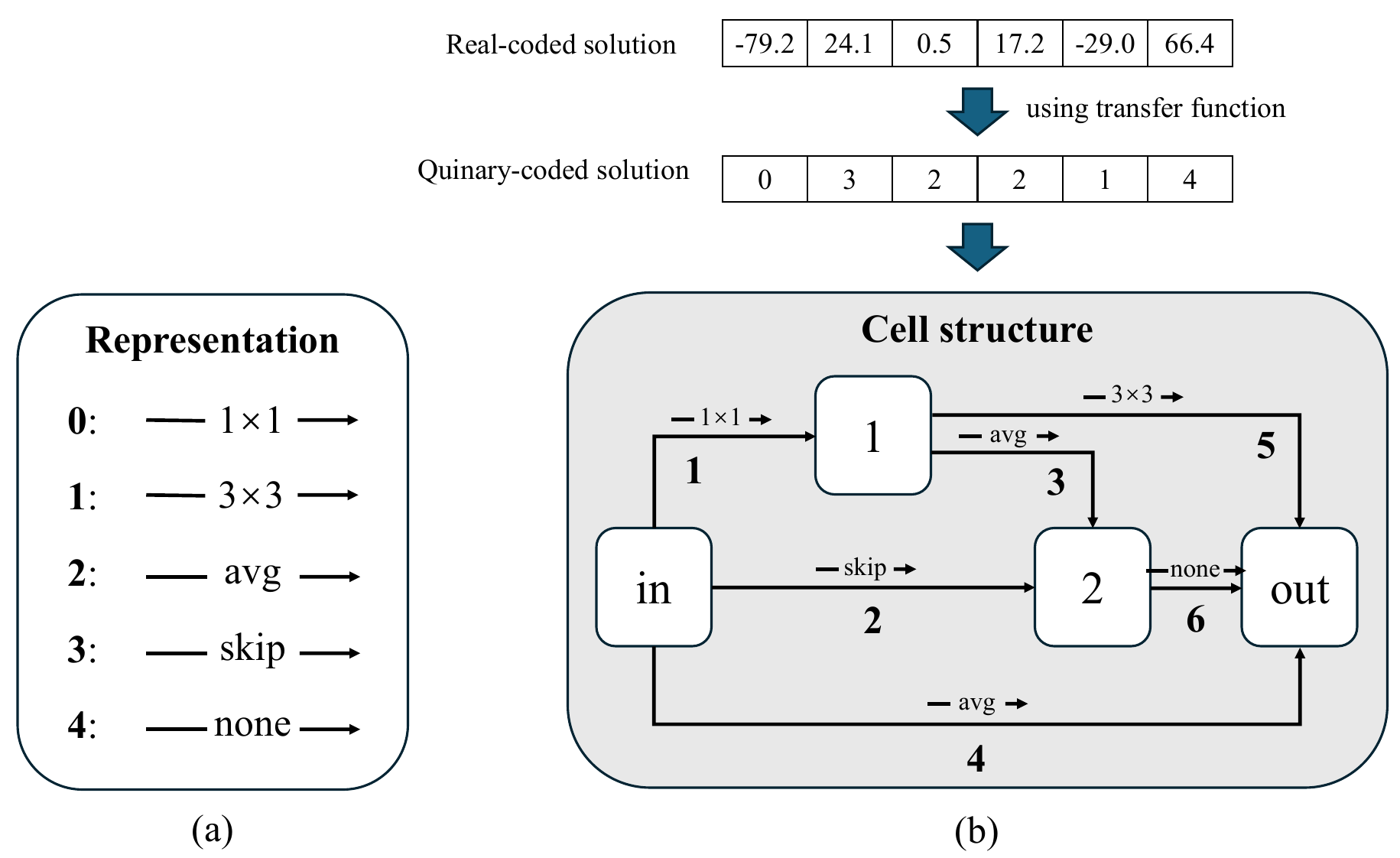}
    \caption{A demonstration of encoding and decoding in the ARNAS task.}
    \label{fig:5.4.2}
\end{figure}

\textbf{Competitor algorithms and parameters}: The internal parameters of the MA are set consistently as detailed in Table \ref{tbl:5.2.4}, where the distinct parameters of the population size and maximum FEs are fixed at 50 and 5000, respectively. Additionally, the nature of the ARNAS task is a combinatorial optimization problem, and it is necessary to transfer the continuous search space to a discrete one. Here, we provide a demonstration of encoding and decoding in Figure \ref{fig:5.4.2}. 
The transfer function is incorporated into MA approaches to convert the real-coded solution to quinary-coded, and then the cells are equipped with specific connections for evaluation. Since parameters of metaheuristic algorithms are often designed for search spaces ranging from -100 to 100, for the sake of simplicity, the truncation function is employed as the transfer function, which is defined in Eq. (\ref{eq:5.4.1}).
\begin{equation}
    \label{eq:5.4.1}
    \begin{aligned}
        g(X_{ij}) = \begin{cases}
		  0, \ if \  X_{ij} < -60 \\
		  1, \ elif \  X_{ij} < -20 \\
		  2, \ elif \  X_{ij} < 20 \\
		  3, \ elif \  X_{ij} < 60 \\
		  4, \ else
	    \end{cases}
    \end{aligned}
\end{equation}

Moreover, four representative adversarial attacks are employed in this numerical experiment: FGSM, PGD, APGD, and Square Attack. These methods are made available in \cite{Jung:23}.

\begin{itemize}
  \item \textbf{FGSM}: Given an input sample $x$ and a neural network model $F$ with parameters $\theta$, the fast gradient sign method (FGSM) \cite{Ian:15} generates adversarial samples using Eq. (\ref{eq:5.4.2}).
  \begin{equation}
    \label{eq:5.4.2}
    \begin{aligned}
        x_{adv} = x + \epsilon \cdot \text{sign}((\nabla_xJ(F(x;\theta)), y_{true}))
    \end{aligned}
    \end{equation}
    where $\epsilon$ is a adjustable parameter, $J(\cdot)$ is the loss function, $y_{true}$ denotes the true label, and $\nabla_x$ represents the gradient with respect to the input.
  \item \textbf{PGD}: Different from the one-step FGSM method, the projected gradient descent (PGD) \cite{Alexey:17} is an iterative method that iteratively refines the perturbation to maximize the model's loss using Eq. (\ref{eq:2.2.2}).
    \begin{equation}
    \label{eq:2.2.2}
    \begin{aligned}
        x_{adv} = \text{clip}_x (x_{adv} + \alpha \cdot \text{sign}((\nabla_xJ(F(x_{adv};\theta)), y_{true})))
    \end{aligned}
    \end{equation}
    where $\alpha$ is a tiny step parameter and $\text{clip}_x$ function  clips to range $[x-\epsilon, x+\epsilon]$.
  \item \textbf{APGD}: As the name suggests, APGD \cite{Croce:20} is an adaptive version of PGD in step size during the iterative optimization process. The core iterative equation is defined in Eq. (\ref{eq:2.2.3}).
    \begin{equation}
    \label{eq:2.2.3}
    \begin{aligned}
        x_{adv} = \text{clip}_x (x_{adv} + \alpha_t \cdot \text{sign}((\nabla_xJ(F(x_{adv};\theta)), y_{true})))
    \end{aligned}
    \end{equation}
    where $\alpha_t$ is the adaptive step size determined based on the gradient magnitude and the optimization progress.
    \item \textbf{Square}: The square method \cite{Maksym:20} is a kind of black-box attack technique. When the gradient information is inaccessible, the square method utilizes the stochastic search to solve the optimization problem presented in Eq. (\ref{eq:2.2.4}).
    \begin{equation}
    \label{eq:2.2.4}
    \begin{aligned}
        \min_{x_{adv}} \{f_{y_{true}, \theta}(x_{adv}) - \max_{k\neq y}f_{k,\theta}(x_{adv})\}, s.t. \Vert x_{adv} - x\Vert_p \leq \epsilon 
    \end{aligned}
    \end{equation}
    where $f_{k,\theta}(\cdot)$ are the network predictions for class $k$ given an image.
\end{itemize}

\begin{figure}[htbp]
    \centering
    \includegraphics[width=16cm]{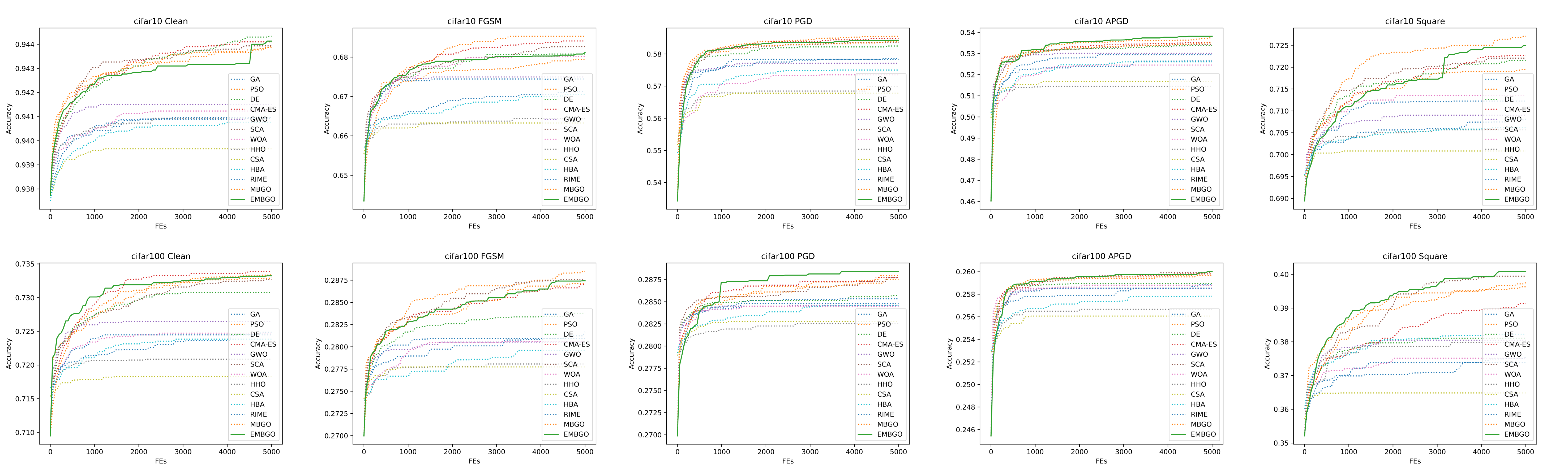}
    \caption{Convergence curves of thirteen MA approaches on the ARNAS benchmark suite.}
    \label{fig:5.4.3}
\end{figure}

\textbf{Experimental results on ARNAS tasks}: The experimental results of thirteen MA approaches on the ARNAS tasks are summarized in Table \ref{tbl:5.4.2}, and the convergence curves are presented in Figure \ref{fig:5.4.3}.
\begin{sidewaystable}[htbp]
	\scriptsize
	\centering
	\renewcommand\arraystretch{1.5}
	\caption{Experimental results of classification accuracy in the ARNAS benchmark suite.}
	\label{tbl:5.4.2}
	\resizebox{0.8\columnwidth}{!}{
		\begin{tabular}{ccccccccccccccccc}
			\toprule
			\multicolumn{2}{c}{Prob.} & GA & PSO & DE & CMA-ES & GWO & SCA & WOA & HHO & CSA & HBA & RIME & MBGO & EMBGO \\
			\midrule
			\multirow{5}{*}{CIFAR-10} & Clean & 94.10 & 94.39 & \textbf{94.43} & 94.41 & 94.15 & 94.39 & 94.12 & 94.09 & 93.97 & 94.08 & 94.09 & 94.39 & 94.41 \\
                ~ & FGSM & 67.12 & \textbf{68.53} & 68.08 & 68.41 & 67.49 & 68.26 & 68.01 & 66.43 & 66.33 & 67.05 & 67.44 & 67.94 & 68.12 \\
                ~ & PGD & 57.86 & \textbf{58.55} & 58.25 & 58.48 & 57.71 & 58.35 & 57.35 & 56.99 & 56.78 & 57.50 & 57.84 & 58.38 & 58.42 \\
                ~ & APGD & 52.65 & 53.74 & 53.38 & 53.54 & 53.01 & 53.42 & 52.45 & 51.45 & 51.68 & 52.60 & 52.94 & 53.50 & \textbf{53.81} \\
                ~ & Squares & 70.81 & \textbf{72.71} & 72.15 & 72.27 & 70.90 & 72.21 & 71.35 & 70.57 & 70.08 & 70.61 & 71.23 & 71.94 & 72.49 \\
			\midrule
                \multicolumn{2}{c}{Avg. rank} & 9.0 & \textbf{2.0} & 4.6 & 2.6 & 8.0 & 4.2 & 8.6 & 12.0 & 12.8 & 10.8 & 8.6 & 5.4 & 2.4 \\
			\midrule
			\multirow{5}{*}{CIFAR-100} & Clean & 72.49 & 73.34 & 73.07 & \textbf{73.39} & 72.65 & 73.27 & 72.47 & 72.09 & 71.83 & 72.38 & 72.45 & 73.30 & 73.32 \\
                ~ & FGSM & 28.17 & \textbf{28.85} & 28.38 & 28.70 & 28.05 & 28.76 & 28.06 & 27.81 & 27.77 & 27.96 & 28.09 & 28.71 & 28.74 \\
                ~ & PGD & 28.53 & 28.79 & 28.57 & 28.77 & 28.46 & 28.77 & 28.45 & 28.25 & 28.28 & 28.46 & 28.48 & 28.75 & \textbf{28.84} \\
                ~ & APGD & 25.88 & 25.98 & 25.90 & 25.98 & 25.85 & 25.99 & 25.88 & 25.67 & 25.61 & 25.78 & 25.85 & 25.97 & \textbf{26.00} \\
                ~ & Squares & 37.55 & 39.75 & 38.11 & 39.14 & 38.05 & 39.95 & 37.51 & 37.98 & 36.48 & 38.22 & 37.38 & 39.63 & \textbf{40.09} \\
			\midrule
                \multicolumn{2}{c}{Avg. rank} & 7.8 & 2.4 & 6.2 & 3.4 & 8.8 & 3.0 & 9.6 & 11.6 & 12.8 & 9.8 & 9.4 & 4.4 & \textbf{1.8} \\
			\bottomrule
		\end{tabular}
	}
\end{sidewaystable}

\begin{figure}[htbp]
    \centering
    \includegraphics[width=16cm]{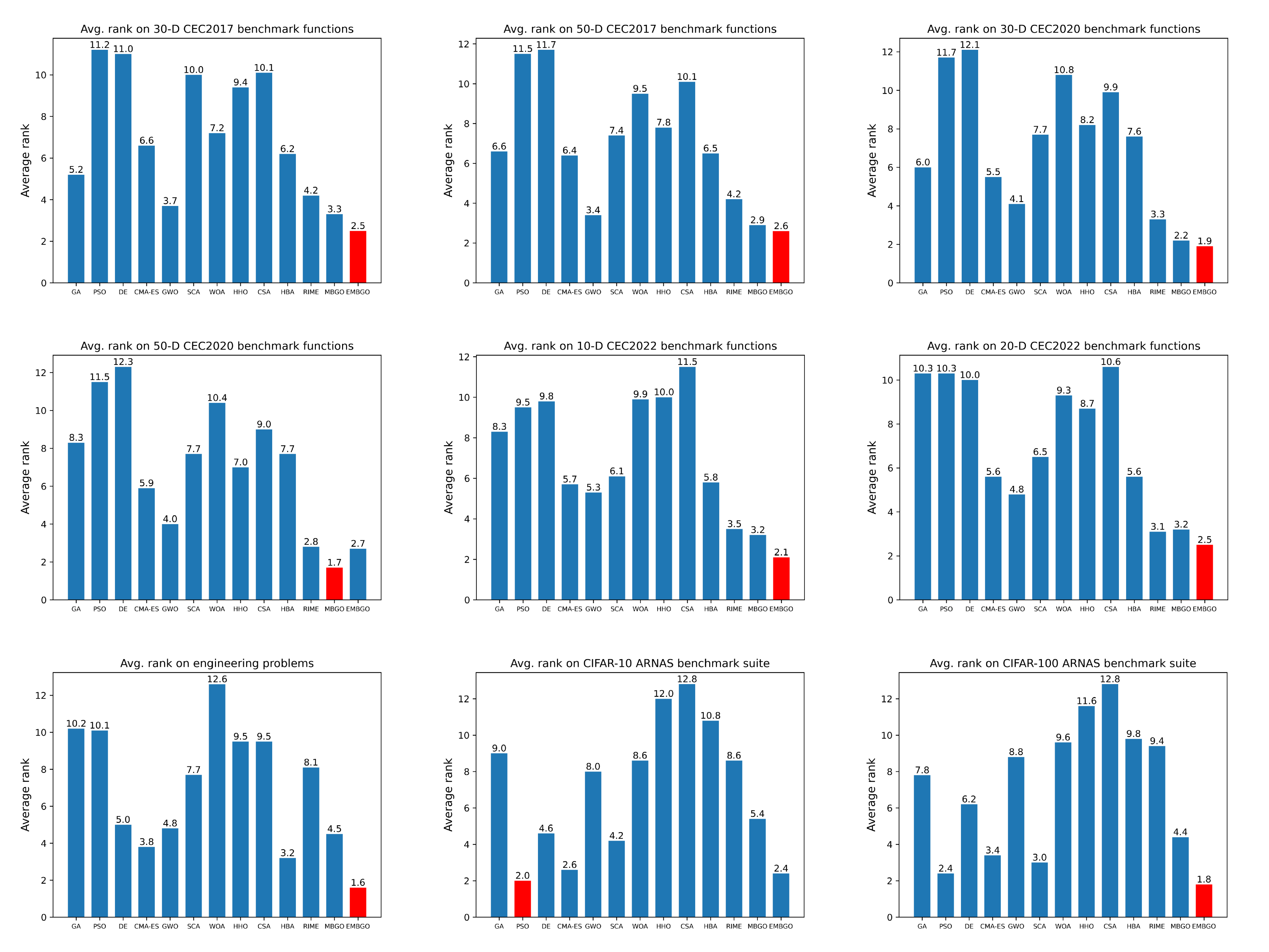}
    \caption{Average ranks of thirteen MA approaches on CEC2017, CEC2020, and CEC2022 benchmark functions, engineering problems, and ARNAS tasks.}
    \label{fig:5.2}
\end{figure}

In summary, the average ranks of thirteen MA approaches on CEC2017, CEC2020, and CEC2022 benchmark functions, engineering problems, and ARNAS tasks are presented in Figure \ref{fig:5.2}.

\section{Discussion} \label{sec:6}
Through the aforementioned numerical experiments, the efficiency and effectiveness of our proposed EMBGO are evident. In the following discussion, we will analyze and delve into the performance of EMBGO. 

\subsection{Computational complexity analysis} \label{sec:6.1}
We begin by offering a theoretical computational complexity analysis for EMBGO. Supposing the population size is $N$, the dimension size is $D$, and the maximum iteration is $T$. In accordance with the pseudocode presented in Algorithm \ref{alg:4.1}, the essential procedures are demonstrated as follows:
\begin{itemize}
  \item Population initialization: $O(N\cdot D)$.
  \item Safe zone determination for a single individual: $O(D)$.
  \item Differential mutation for a single individual: $O(D)$.
  \item Lévy flight for a single individual: $O(D)$.
  \item Search operator in Eq. for a single individual (\ref{eq:2.1.4}): $O(D)$.
  \item Search operator in Eq. for a single individual (\ref{eq:2.1.5}): $O(D)$.
  \item Embedded greedy selection for a single individual: $O(1)$.
\end{itemize}

In summary, the computational complexity of EMBGO is $O(N\cdot D + T\cdot (N\cdot (4D + 1))):= O(T\cdot N\cdot D)$, which is identical to the original MBGO. 

\subsection{Performance analysis on CEC benchmark functions} \label{sec:6.2}
Extensive numerical experiments were conducted on CEC2017, CEC2020, and CEC2022 benchmark functions to comprehensively evaluate our proposed EMBGO. Twelve well-known MAs are employed as competitor algorithms. The experimental results and statistical analyses demonstrate the superior and outstanding performance of EMBGO in most instances. These significant improvements can be attributed to the incorporation of the integrated mixed movement and battle phase, alongside the innovative utilization of differential mutation and Lévy flight operator. 

EMBGO derives advantages from its amalgamated architecture, comprising two integral components: the movement and battle phase. Within a given iteration, individuals exhibit an elevated likelihood of adopting various search operators, thereby maintaining population diversity throughout the optimization process. A well-maintained population diversity allows the optimization to avoid premature convergence, explore unknown regions, and escape from local optima. Convergence curves, exemplified by $f_1$ and $f_3$ in CEC2017 benchmark functions, $f_1$ to $f_3$ and $f_7$ in CEC2020 benchmark functions, and $f_1$ and $f_4$ in CEC2022 benchmark functions, indicate that many MA approaches tend to prematurely converge, with their convergence curves plateauing in the early and middle stages of optimization. In contrast, EMBGO persists in the optimization process until its scheduled termination, partly owing to its well-designed structure.

Furthermore, the combination of the novel differential mutation and Lévy flight operator plays a crucial role in augmenting the performance of EMBGO. The proposed differential mutation operator effectively leverages valuable information from the both current best individual $X_{best}$ and the centroid location $X_{mean}$. Adhering to the proximate optimality principle, this operator significantly accelerates the movement of individuals within the safe zone towards promising regions, introducing a controlled degree of randomness through carefully designed coefficients. Moreover, the incorporation of Lévy flight into EMBGO effectively emulates the movement of players outside the safe zone, promisingly enhancing exploration and global search capacities. In conclusion, the demonstrated superior performance of our proposed EMBGO is evident across multiple dimensions, including experimental results, statistical analyses, convergence curves, and average ranks on diverse benchmark functions.

While EMBGO exhibits remarkable performance across various scenarios, certain noteworthy shortcomings become apparent in specific instances: (1). Sub-optimal performance in certain cases. EMBGO shows a significant performance gap compared to CMA-ES and GWO on certain hybrid and composite functions within the CEC2017 benchmark. Similar observations arise in CEC2020 and CEC2022 benchmark functions. Despite these limitations, it is crucial to acknowledge the substantial improvements over MBGO. The performance gap is attributed to inherent search preferences inherited from the original MBGO, highlighting the need for further refinement in future research. (2). Performance deterioration with increased dimensionality. EMBGO experiences notable performance deterioration as the dimensionality of the problem increases. This suggests that EMBGO may lack scalability and robustness in handling various high-dimensional optimization problems. A potential avenue for addressing this challenge involves integrating the cooperative coevolution (CC) framework \cite{Zhong:23_1} into EMBGO. CC has the capacity to decompose large-scale problems into multiple sub-components, optimizing them iteratively. This integration could enhance the algorithm's capability to handle high-dimensional scenarios more effectively. Future research efforts may explore and integrate such strategies for improved performance.

\subsection{Performance analysis on engineering problems} \label{sec:6.3}
The engineering simulation problems employed in our study represent well-established benchmarks, serving as standard metrics for evaluating the efficacy of optimization algorithms in real-world scenarios. Given the significance of robustness and stability in practical applications, we present both the best and worst performance metrics observed among the 30 independent trial runs for each problem. The exceptional performance of EMBGO across eight engineering problems is summarized in Table \ref{tbl:5.3.2} and visually illustrated in Figure \ref{fig:5.3.1}. These comprehensive results indicate that underscore the potential of EMBGO as a viable alternative optimization technique for addressing real-world engineering optimization problems. Further exploration and application of EMBGO in diverse real-world scenarios are promising avenues for future research.

\subsection{Performance analysis on ARNAS tasks} \label{sec:6.4}
In this study, we extended the application of our proposed EMBGO to ARNAS tasks, providing a further evaluation of its performance in addressing complex discrete optimization problems. The experimental results and average ranks, as presented in Table \ref{tbl:5.4.2}, indicate that our EMBGO performs competitively when compared to other competitor algorithms in ARNAS tasks. Interestingly, the practically less successful PSO outperforms other MA approaches in ARNAS tasks involving the CIFAR-10 dataset and achieves the second-best performance in ARNAS tasks with the CIFAR-100 dataset. This observation aligns with the No Free Lunch Theorem, reinforcing the idea that the performance of an algorithm is problem-specific.

We hypothesize that the unique selection mechanism of PSO, where all constructed offspring individuals replace parent individuals, plays a significant role in its effectiveness in handling complex ARNAS tasks. This is in contrast to other MA approaches that follow the "survival of the fittest" principle, allowing only improved offspring individuals to survive. The less stringent selection process of PSO may mitigate the risk of getting trapped in local optima. Given the limited search space in ARNAS tasks, this unique selection mechanism of PSO appears to offer increased efficiency. Therefore, considering the advantages of this all-acceptance selection mechanism from PSO, incorporating it into EMBGO could be a promising direction for addressing discrete optimization problems. Further exploration and experimentation in this direction have the potential to reveal new insights and enhance the algorithm's applicability to a broader range of problems.

\section{Conclusion} \label{sec:7}
This paper introduces the efficient multiplayer battle game optimizer (EMBGO) as a novel MA designed for addressing complex numerical optimization problems. The motivation behind EMBGO arises from identified shortcomings in the original MBGO, particularly concerning the design of search operators in the movement phase, as revealed through ablation experiments. To address these issues, EMBGO integrates the movement and battle phases of MBGO, incorporating two efficient search operators in place of the original movement phase operators: differential mutation and Lévy flight.

The performance of EMBGO undergoes thorough evaluation through comprehensive numerical experiments on CEC2017, CEC2020, and CEC2022 benchmark functions, alongside eight engineering problems. Twelve well-established MA approaches are employed as competitor algorithms for comparative analysis. The experimental results and statistical analyses affirm the efficiency and effectiveness of EMBGO across diverse optimization scenarios. However, the paper acknowledges specific limitations of EMBGO, particularly in addressing certain hybrid and composite problems as well as grappling with challenges in high-dimensional spaces. We duly emphasize the potential for enhancement in future research endeavors.

Furthermore, the paper extends the application of EMBGO to address the challenges of complex adversarial robust neural architecture search (ARNAS), demonstrating its competitive performance in this specialized domain. The conclusion suggests that EMBGO, as a potential technique, holds promise for widespread application in real-world problems and deep learning applications.

\section{Statement and Declarations}

\subsection{Competing interest}
The authors declare that they have no known competing financial interests or personal relationships that could have appeared to influence the work reported in this paper.

\subsection{Data availability}
The source code of this research can be downloaded from \url{https://github.com/RuiZhong961230/EMBGO}.

\subsection{Acknowledgement}
This work was supported by JST SPRING Grant Number JPMJSP2119.

\bibliographystyle{elsarticle-num}
\bibliography{paper}

\begin{thebibliography}{10}
\expandafter\ifx\csname url\endcsname\relax
  \def\url#1{\texttt{#1}}\fi
\expandafter\ifx\csname urlprefix\endcsname\relax\def\urlprefix{URL }\fi
\expandafter\ifx\csname href\endcsname\relax
  \def\href#1#2{#2} \def\path#1{#1}\fi

\bibitem{Zhong:23_2}
R.~Zhong, E.~Zhang, M.~Munetomo, Cooperative coevolutionary differential
  evolution with linkage measurement minimization for large-scale optimization
  problems in noisy environments, Complex \& Intelligent System 9 (2023)
  4439–4456.
\newblock \href {https://doi.org/https://doi.org/10.1007/s40747-022-00957-6}
  {\path{doi:https://doi.org/10.1007/s40747-022-00957-6}}.

\bibitem{Sayed:24}
E.-S.~M. El-kenawy, N.~Khodadadi, S.~Mirjalili, A.~A. Abdelhamid, M.~M. Eid,
  A.~Ibrahim, Greylag goose optimization: Nature-inspired optimization
  algorithm, Expert Systems with Applications 238 (2024) 122147.
\newblock \href {https://doi.org/https://doi.org/10.1016/j.eswa.2023.122147}
  {\path{doi:https://doi.org/10.1016/j.eswa.2023.122147}}.

\bibitem{Guo:24}
W.~Zhao, L.~Wang, Z.~Zhang, H.~Fan, J.~Zhang, S.~Mirjalili, N.~Khodadadi,
  Q.~Cao, Electric eel foraging optimization: A new bio-inspired optimizer for
  engineering applications, Expert Systems with Applications 238 (2024) 122200.
\newblock \href {https://doi.org/https://doi.org/10.1016/j.eswa.2023.122200}
  {\path{doi:https://doi.org/10.1016/j.eswa.2023.122200}}.

\bibitem{Zhang:23}
Q.~Zhang, H.~Gao, Z.-H. Zhan, J.~Li, H.~Zhang, Growth optimizer: A powerful
  metaheuristic algorithm for solving continuous and discrete global
  optimization problems, Knowledge-Based Systems 261 (2023) 110206.
\newblock \href {https://doi.org/https://doi.org/10.1016/j.knosys.2022.110206}
  {\path{doi:https://doi.org/10.1016/j.knosys.2022.110206}}.

\bibitem{Rui:24}
R.~Zhong, C.~Zhang, J.~Yu, Chaotic vegetation evolution: leveraging multiple
  seeding strategies and a mutation module for global optimization problems,
  Evolutionary Intelligence (2024) 1--25\href
  {https://doi.org/https://doi.org/10.1007/s12065-023-00892-6}
  {\path{doi:https://doi.org/10.1007/s12065-023-00892-6}}.

\bibitem{Siamak:21}
S.~Talatahari, M.~Azizi, A.~H. Gandomi, Material generation algorithm: A novel
  metaheuristic algorithm for optimization of engineering problems, Processes
  9~(5) (2021).
\newblock \href {https://doi.org/10.3390/pr9050859}
  {\path{doi:10.3390/pr9050859}}.

\bibitem{Shehadeh:23}
H.~Shehadeh, Chernobyl disaster optimizer (cdo): a novel meta-heuristic method
  for global optimization, Neural Computing and Applications (02 2023).
\newblock \href {https://doi.org/10.1007/s00521-023-08261-1}
  {\path{doi:10.1007/s00521-023-08261-1}}.

\bibitem{Basset:22}
M.~Abdel-Basset, R.~Mohamed, K.~M. Sallam, R.~K. Chakrabortty, Light spectrum
  optimizer: A novel physics-inspired metaheuristic optimization algorithm,
  Mathematics 10~(19) (2022).
\newblock \href {https://doi.org/10.3390/math10193466}
  {\path{doi:10.3390/math10193466}}.

\bibitem{Sujoy:24}
S.~Barua, A.~Merabet, Lévy arithmetic algorithm: An enhanced metaheuristic
  algorithm and its application to engineering optimization, Expert Systems
  with Applications 241 (2024) 122335.
\newblock \href {https://doi.org/https://doi.org/10.1016/j.eswa.2023.122335}
  {\path{doi:https://doi.org/10.1016/j.eswa.2023.122335}}.

\bibitem{Velasco:23}
L.~Velasco, H.~Guerrero, A.~Hospitaler, A literature review and critical
  analysis of metaheuristics recently developed, Archives of Computational
  Methods in Engineering 31 (07 2023).
\newblock \href {https://doi.org/10.1007/s11831-023-09975-0}
  {\path{doi:10.1007/s11831-023-09975-0}}.

\bibitem{Rajwar:23}
K.~Rajwar, K.~Deep, S.~Das, An exhaustive review of the metaheuristic
  algorithms for search and optimization: taxonomy, applications, and open
  challenges, Artificial Intelligence Review 56 (04 2023).
\newblock \href {https://doi.org/10.1007/s10462-023-10470-y}
  {\path{doi:10.1007/s10462-023-10470-y}}.

\bibitem{Mohammadi:23}
A.~Mohammadi, F.~Sheikholeslam, Intelligent optimization: Literature review and
  state-of-the-art algorithms (1965–2022), Engineering Applications of
  Artificial Intelligence 126 (2023) 106959.
\newblock \href {https://doi.org/10.1016/j.engappai.2023.106959}
  {\path{doi:10.1016/j.engappai.2023.106959}}.

\bibitem{Wolpert:97}
D.~Wolpert, W.~Macready, No free lunch theorems for optimization, IEEE
  Transactions on Evolutionary Computation 1~(1) (1997) 67--82.
\newblock \href {https://doi.org/10.1109/4235.585893}
  {\path{doi:10.1109/4235.585893}}.

\bibitem{Zhong:23}
R.~Zhong, F.~Peng, E.~Zhang, J.~Yu, M.~Munetomo, Vegetation evolution with
  dynamic maturity strategy and diverse mutation strategy for solving
  optimization problems, Biomimetics 8~(6) (2023).
\newblock \href {https://doi.org/10.3390/biomimetics8060454}
  {\path{doi:10.3390/biomimetics8060454}}.

\bibitem{Xu:23}
Y.~Xu, R.~Zhong, C.~Zhang, J.~Yu, Multiplayer battle game-inspired optimizer
  for complex optimization problems (2023).
\newblock \href {http://arxiv.org/abs/2401.00401} {\path{arXiv:2401.00401}}.

\bibitem{Zoph:17}
B.~Zoph, Q.~V. Le, \href{https://arxiv.org/abs/1611.01578}{Neural architecture
  search with reinforcement learning} (2017).
\newline\urlprefix\url{https://arxiv.org/abs/1611.01578}

\bibitem{Tom:19}
T.~Véniat, O.~Schwander, L.~Denoyer, Stochastic adaptive neural architecture
  search for keyword spotting, in: ICASSP 2019 - 2019 IEEE International
  Conference on Acoustics, Speech and Signal Processing (ICASSP), 2019, pp.
  2842--2846.
\newblock \href {https://doi.org/10.1109/ICASSP.2019.8683305}
  {\path{doi:10.1109/ICASSP.2019.8683305}}.

\bibitem{Li:20}
L.~Li, A.~Talwalkar, Random search and reproducibility for neural architecture
  search, in: Proceedings of The 35th Uncertainty in Artificial Intelligence
  Conference, Vol. 115 of Proceedings of Machine Learning Research, PMLR, 2020,
  pp. 367--377.

\bibitem{Xie:20}
S.~Xie, H.~Zheng, C.~Liu, L.~Lin, Snas: Stochastic neural architecture search
  (2020).
\newblock \href {http://arxiv.org/abs/1812.09926} {\path{arXiv:1812.09926}}.

\bibitem{Hieu:18}
H.~Pham, M.~Y. Guan, B.~Zoph, Q.~V. Le, J.~Dean, Efficient neural architecture
  search via parameter sharing (2018).
\newblock \href {http://arxiv.org/abs/1802.03268} {\path{arXiv:1802.03268}}.

\bibitem{Ramakanth:19}
R.~Pasunuru, M.~Bansal, Continual and multi-task architecture search (2019).
\newblock \href {http://arxiv.org/abs/1906.05226} {\path{arXiv:1906.05226}}.

\bibitem{Cai:19}
H.~Cai, L.~Zhu, S.~Han, Proxylessnas: Direct neural architecture search on
  target task and hardware (2019).
\newblock \href {http://arxiv.org/abs/1812.00332} {\path{arXiv:1812.00332}}.

\bibitem{Gabriel:20}
G.~Bender, H.~Liu, B.~Chen, G.~Chu, S.~Cheng, P.-J. Kindermans, Q.~Le, Can
  weight sharing outperform random architecture search? an investigation with
  tunas (2020).
\newblock \href {http://arxiv.org/abs/2008.06120} {\path{arXiv:2008.06120}}.

\bibitem{Guo:20}
Y.~Guo, Y.~Chen, Y.~Zheng, P.~Zhao, J.~Chen, J.~Huang, M.~Tan, Breaking the
  curse of space explosion: Towards efficient nas with curriculum search
  (2020).
\newblock \href {http://arxiv.org/abs/2007.07197} {\path{arXiv:2007.07197}}.

\bibitem{Han:20}
H.~Shi, R.~Pi, H.~Xu, Z.~Li, J.~T. Kwok, T.~Zhang, Bridging the gap between
  sample-based and one-shot neural architecture search with bonas (2020).
\newblock \href {http://arxiv.org/abs/1911.09336} {\path{arXiv:1911.09336}}.

\bibitem{Chu:21}
X.~Chu, B.~Zhang, R.~Xu, Fairnas: Rethinking evaluation fairness of weight
  sharing neural architecture search (2021).
\newblock \href {http://arxiv.org/abs/1907.01845} {\path{arXiv:1907.01845}}.

\bibitem{Chen:21}
B.~Chen, P.~Li, C.~Li, B.~Li, L.~Bai, C.~Lin, M.~Sun, J.~yan, W.~Ouyang, Glit:
  Neural architecture search for global and local image transformer (2021).
\newblock \href {http://arxiv.org/abs/2107.02960} {\path{arXiv:2107.02960}}.

\bibitem{Xia:21}
X.~Xia, X.~Xiao, X.~Wang, M.~Zheng, Progressive automatic design of search
  space for one-shot neural architecture search (2021).
\newblock \href {http://arxiv.org/abs/2005.07564} {\path{arXiv:2005.07564}}.

\bibitem{Tong:22}
L.~Tong, B.~Du, Neural architecture search via reference point based
  multi-objective evolutionary algorithm, Pattern Recognition 132 (2022)
  108962.
\newblock \href {https://doi.org/https://doi.org/10.1016/j.patcog.2022.108962}
  {\path{doi:https://doi.org/10.1016/j.patcog.2022.108962}}.

\bibitem{Dong:19}
X.~Dong, Y.~Yang, Searching for a robust neural architecture in four gpu hours
  (2019).
\newblock \href {http://arxiv.org/abs/1910.04465} {\path{arXiv:1910.04465}}.

\bibitem{Yao:19}
Q.~Yao, J.~Xu, W.-W. Tu, Z.~Zhu, Efficient neural architecture search via
  proximal iterations (2019).
\newblock \href {http://arxiv.org/abs/1905.13577} {\path{arXiv:1905.13577}}.

\bibitem{Liu:19}
H.~Liu, K.~Simonyan, Y.~Yang, Darts: Differentiable architecture search (2019).
\newblock \href {http://arxiv.org/abs/1806.09055} {\path{arXiv:1806.09055}}.

\bibitem{Jiang:19}
Y.~Jiang, C.~Hu, T.~Xiao, C.~Zhang, J.~Zhu, Improved differentiable
  architecture search for language modeling and named entity recognition, in:
  Proceedings of the 2019 Conference on Empirical Methods in Natural Language
  Processing and the 9th International Joint Conference on Natural Language
  Processing (EMNLP-IJCNLP), Association for Computational Linguistics, Hong
  Kong, China, 2019, pp. 3585--3590.
\newblock \href {https://doi.org/10.18653/v1/D19-1367}
  {\path{doi:10.18653/v1/D19-1367}}.

\bibitem{Wang:23}
W.~Wang, X.~Zhang, H.~Cui, H.~Yin, Y.~Zhang, Fp-darts: Fast parallel
  differentiable neural architecture search for image classification, Pattern
  Recognition 136 (2023) 109193.
\newblock \href {https://doi.org/https://doi.org/10.1016/j.patcog.2022.109193}
  {\path{doi:https://doi.org/10.1016/j.patcog.2022.109193}}.

\bibitem{Ren:21}
P.~Ren, Y.~Xiao, X.~Chang, P.-y. Huang, Z.~Li, X.~Chen, X.~Wang, A
  comprehensive survey of neural architecture search: Challenges and solutions,
  ACM Comput. Surv. 54~(4) (may 2021).
\newblock \href {https://doi.org/10.1145/3447582} {\path{doi:10.1145/3447582}}.

\bibitem{Matt:24}
M.~Poyser, T.~P. Breckon, Neural architecture search: A contemporary literature
  review for computer vision applications, Pattern Recognition 147 (2024)
  110052.
\newblock \href {https://doi.org/https://doi.org/10.1016/j.patcog.2023.110052}
  {\path{doi:https://doi.org/10.1016/j.patcog.2023.110052}}.

\bibitem{Ramtin:20}
R.~Hosseini, X.~Yang, P.~Xie, Dsrna: Differentiable search of robust neural
  architectures (2020).
\newblock \href {http://arxiv.org/abs/2012.06122} {\path{arXiv:2012.06122}}.

\bibitem{Chaitanya:21}
C.~Devaguptapu, D.~Agarwal, G.~Mittal, P.~Gopalani, V.~N. Balasubramanian, On
  adversarial robustness: A neural architecture search perspective (2021).
\newblock \href {http://arxiv.org/abs/2007.08428} {\path{arXiv:2007.08428}}.

\bibitem{Xie:23}
B.~Xie, H.~Chang, Z.~Zhang, X.~Wang, D.~Wang, Z.~Zhang, R.~Ying, W.~Zhu,
  Adversarially robust neural architecture search for graph neural networks,
  in: Proceedings of the IEEE/CVF Conference on Computer Vision and Pattern
  Recognition (CVPR), 2023, pp. 8143--8152.

\bibitem{Jung:23}
S.~Jung, J.~Lukasik, M.~Keuper, Neural architecture design and robustness: A
  dataset, in: ICLR, 2023.

\bibitem{Dirk:20}
D.~Sudholt, The Benefits of Population Diversity in Evolutionary Algorithms: A
  Survey of Rigorous Runtime Analyses, Springer International Publishing, Cham,
  2020, pp. 359--404.
\newblock \href {https://doi.org/10.1007/978-3-030-29414-4_8}
  {\path{doi:10.1007/978-3-030-29414-4_8}}.

\bibitem{Holm:79}
S.~Holm, A simple sequentially rejective multiple test procedure, Scandinavian
  Journal of Statistics 6~(2) (1979) 65--70.

\bibitem{Min:20}
S.~Li, H.~Chen, M.~Wang, A.~A. Heidari, S.~Mirjalili, Slime mould algorithm: A
  new method for stochastic optimization, Future Generation Computer Systems
  111 (2020) 300--323.
\newblock \href {https://doi.org/https://doi.org/10.1016/j.future.2020.03.055}
  {\path{doi:https://doi.org/10.1016/j.future.2020.03.055}}.

\bibitem{Zhao:24}
W.~Zhao, L.~Wang, Z.~Zhang, H.~Fan, J.~Zhang, S.~Mirjalili, N.~Khodadadi,
  Q.~Cao, Electric eel foraging optimization: A new bio-inspired optimizer for
  engineering applications, Expert Systems with Applications 238 (2024) 122200.
\newblock \href {https://doi.org/https://doi.org/10.1016/j.eswa.2023.122200}
  {\path{doi:https://doi.org/10.1016/j.eswa.2023.122200}}.

\bibitem{Niu:24}
Y.~Niu, X.~Yan, Y.~Wang, Y.~Niu, An improved sand cat swarm optimization for
  moving target search by uav, Expert Systems with Applications 238 (2024)
  122189.
\newblock \href {https://doi.org/https://doi.org/10.1016/j.eswa.2023.122189}
  {\path{doi:https://doi.org/10.1016/j.eswa.2023.122189}}.

\bibitem{Faridmehr:23}
I.~Faridmehr, M.~L. Nehdi, I.~F. Davoudkhani, A.~Poolad, Mountaineering
  team-based optimization: A novel human-based metaheuristic algorithm,
  Mathematics 11~(5) (2023).
\newblock \href {https://doi.org/10.3390/math11051273}
  {\path{doi:10.3390/math11051273}}.

\bibitem{Lian:24}
J.~Lian, G.~Hui, Human evolutionary optimization algorithm, Expert Systems with
  Applications 241 (2024) 122638.
\newblock \href {https://doi.org/https://doi.org/10.1016/j.eswa.2023.122638}
  {\path{doi:https://doi.org/10.1016/j.eswa.2023.122638}}.

\bibitem{Yaguchi:11}
K.~Yaguchi, K.~Tamura, K.~Yasuda, A.~Ishigame, Basic study of proximate
  optimality principle based combinatorial optimization method, in: 2011 IEEE
  International Conference on Systems, Man, and Cybernetics, 2011, pp.
  1753--1758.
\newblock \href {https://doi.org/10.1109/ICSMC.2011.6083925}
  {\path{doi:10.1109/ICSMC.2011.6083925}}.

\bibitem{Wu:23}
W.~Lei, W.~Jiawei, M.~Zezhou, Enhancing grey wolf optimizer with levy flight
  for engineering applications, IEEE Access 11 (2023) 74865--74897.
\newblock \href {https://doi.org/10.1109/ACCESS.2023.3295242}
  {\path{doi:10.1109/ACCESS.2023.3295242}}.

\bibitem{Saravanan:23}
G.~Saravanan, S.~Neelakandan, P.~Ezhumalai, S.~Maurya, Improved wild horse
  optimization with levy flight algorithm for effective task scheduling in
  cloud computing, J. Cloud Comput. 12~(1) (feb 2023).
\newblock \href {https://doi.org/10.1186/s13677-023-00401-1}
  {\path{doi:10.1186/s13677-023-00401-1}}.

\bibitem{Chang:23}
C.~Zhong, G.~Li, Z.~Meng, W.~He, Opposition-based learning equilibrium
  optimizer with levy flight and evolutionary population dynamics for
  high-dimensional global optimization problems, Expert Systems with
  Applications 215 (2023) 119303.
\newblock \href {https://doi.org/https://doi.org/10.1016/j.eswa.2022.119303}
  {\path{doi:https://doi.org/10.1016/j.eswa.2022.119303}}.

\bibitem{Syama:23}
S.~Syama, J.~Ramprabhakar, R.~Anand, J.~M. Guerrero, A hybrid extreme learning
  machine model with lévy flight chaotic whale optimization algorithm for wind
  speed forecasting, Results in Engineering 19 (2023) 101274.
\newblock \href {https://doi.org/https://doi.org/10.1016/j.rineng.2023.101274}
  {\path{doi:https://doi.org/10.1016/j.rineng.2023.101274}}.

\bibitem{Nguyen:23}
N.~{Van Thieu}, S.~Mirjalili, Mealpy: An open-source library for latest
  meta-heuristic algorithms in python, Journal of Systems Architecture 139
  (2023) 102871.
\newblock \href {https://doi.org/https://doi.org/10.1016/j.sysarc.2023.102871}
  {\path{doi:https://doi.org/10.1016/j.sysarc.2023.102871}}.

\bibitem{Thieu:20}
T.~Nguyen, A framework of optimization functions using numpy (opfunu) for
  optimization problems (2020).
\newblock \href {https://doi.org/10.5281/zenodo.3620960}
  {\path{doi:10.5281/zenodo.3620960}}.

\bibitem{Zhong:24}
R.~Zhong, F.~Peng, J.~Yu, M.~Munetomo, Q-learning based vegetation evolution
  for numerical optimization and wireless sensor network coverage optimization,
  Alexandria Engineering Journal 87 (2024) 148--163.
\newblock \href {https://doi.org/https://doi.org/10.1016/j.aej.2023.12.028}
  {\path{doi:https://doi.org/10.1016/j.aej.2023.12.028}}.

\bibitem{Thieu:23}
N.~V. Thieu, Enoppy: A python library for engineering optimization problems
  (may 2023).
\newblock \href {https://doi.org/10.5281/zenodo.7953206}
  {\path{doi:10.5281/zenodo.7953206}}.

\bibitem{Srinivas:94}
M.~Srinivas, L.~Patnaik, Genetic algorithms: a survey, Computer 27~(6) (1994)
  17--26.
\newblock \href {https://doi.org/10.1109/2.294849}
  {\path{doi:10.1109/2.294849}}.

\bibitem{Kennedy:95}
J.~Kennedy, R.~Eberhart, Particle swarm optimization, in: Proceedings of
  ICNN'95 - International Conference on Neural Networks, Vol.~4, 1995, pp.
  1942--1948 vol.4.
\newblock \href {https://doi.org/10.1109/ICNN.1995.488968}
  {\path{doi:10.1109/ICNN.1995.488968}}.

\bibitem{Storn:97}
R.~Storn, K.~Price, Differential evolution - a simple and efficient heuristic
  for global optimization over continuous spaces, Journal of Global
  Optimization 11 (1997) 341--359.
\newblock \href {https://doi.org/10.1023/A:1008202821328}
  {\path{doi:10.1023/A:1008202821328}}.

\bibitem{Hansen:01}
N.~Hansen, A.~Ostermeier, Completely derandomized self-adaptation in evolution
  strategies, Evolutionary Computation 9~(2) (2001) 159--195.
\newblock \href {https://doi.org/10.1162/106365601750190398}
  {\path{doi:10.1162/106365601750190398}}.

\bibitem{Seyedali:14}
S.~Mirjalili, S.~M. Mirjalili, A.~Lewis, Grey wolf optimizer, Advances in
  Engineering Software 69 (2014) 46--61.
\newblock \href
  {https://doi.org/https://doi.org/10.1016/j.advengsoft.2013.12.007}
  {\path{doi:https://doi.org/10.1016/j.advengsoft.2013.12.007}}.

\bibitem{Seyedali:16}
S.~Mirjalili, Sca: A sine cosine algorithm for solving optimization problems,
  Knowledge-Based Systems 96 (2016) 120--133.
\newblock \href {https://doi.org/https://doi.org/10.1016/j.knosys.2015.12.022}
  {\path{doi:https://doi.org/10.1016/j.knosys.2015.12.022}}.

\bibitem{Andrew:16}
S.~Mirjalili, A.~Lewis, The whale optimization algorithm, Advances in
  Engineering Software 95 (2016) 51--67.
\newblock \href
  {https://doi.org/https://doi.org/10.1016/j.advengsoft.2016.01.008}
  {\path{doi:https://doi.org/10.1016/j.advengsoft.2016.01.008}}.

\bibitem{Ali:19}
A.~A. Heidari, S.~Mirjalili, H.~Faris, I.~Aljarah, M.~Mafarja, H.~Chen, Harris
  hawks optimization: Algorithm and applications, Future Generation Computer
  Systems 97 (2019) 849--872.
\newblock \href {https://doi.org/https://doi.org/10.1016/j.future.2019.02.028}
  {\path{doi:https://doi.org/10.1016/j.future.2019.02.028}}.

\bibitem{Qais:22}
M.~H. Qais, H.~M. Hasanien, R.~A. Turky, S.~Alghuwainem, M.~Tostado-Véliz,
  F.~Jurado, Circle search algorithm: A geometry-based metaheuristic
  optimization algorithm, Mathematics 10~(10) (2022).
\newblock \href {https://doi.org/10.3390/math10101626}
  {\path{doi:10.3390/math10101626}}.

\bibitem{Fatma:22}
F.~A. Hashim, E.~H. Houssein, K.~Hussain, M.~S. Mabrouk, W.~Al-Atabany, Honey
  badger algorithm: New metaheuristic algorithm for solving optimization
  problems, Mathematics and Computers in Simulation 192 (2022) 84--110.
\newblock \href {https://doi.org/https://doi.org/10.1016/j.matcom.2021.08.013}
  {\path{doi:https://doi.org/10.1016/j.matcom.2021.08.013}}.

\bibitem{Su:23}
H.~Su, D.~Zhao, A.~A. Heidari, L.~Liu, X.~Zhang, M.~Mafarja, H.~Chen,
  \href{https://www.sciencedirect.com/science/article/pii/S0925231223001480}{Rime:
  A physics-based optimization}, Neurocomputing 532 (2023) 183--214.
\newblock \href {https://doi.org/https://doi.org/10.1016/j.neucom.2023.02.010}
  {\path{doi:https://doi.org/10.1016/j.neucom.2023.02.010}}.
\newline\urlprefix\url{https://www.sciencedirect.com/science/article/pii/S0925231223001480}

\bibitem{Hadi:21}
H.~Bayzidi, S.~Talatahari, M.~Saraee, C.-P. Lamarche, Social network search for
  solving engineering optimization problems, Computational Intelligence and
  Neuroscience 2021 (2021) 1--32.
\newblock \href {https://doi.org/10.1155/2021/8548639}
  {\path{doi:10.1155/2021/8548639}}.

\bibitem{Dong:22}
X.~Dong, L.~Liu, K.~Musial, B.~Gabrys, Nats-bench: Benchmarking nas algorithms
  for architecture topology and size, IEEE Transactions on Pattern Analysis and
  Machine Intelligence 44~(7) (2022) 3634--3646.
\newblock \href {https://doi.org/10.1109/TPAMI.2021.3054824}
  {\path{doi:10.1109/TPAMI.2021.3054824}}.

\bibitem{Ian:15}
I.~J. Goodfellow, J.~Shlens, C.~Szegedy, Explaining and harnessing adversarial
  examples (2015).
\newblock \href {http://arxiv.org/abs/1412.6572} {\path{arXiv:1412.6572}}.

\bibitem{Alexey:17}
A.~Kurakin, I.~Goodfellow, S.~Bengio, Adversarial machine learning at scale
  (2017).
\newblock \href {http://arxiv.org/abs/1611.01236} {\path{arXiv:1611.01236}}.

\bibitem{Croce:20}
F.~Croce, M.~Hein, Reliable evaluation of adversarial robustness with an
  ensemble of diverse parameter-free attacks (2020).
\newblock \href {http://arxiv.org/abs/2003.01690} {\path{arXiv:2003.01690}}.

\bibitem{Maksym:20}
M.~Andriushchenko, F.~Croce, N.~Flammarion, M.~Hein, Square attack: a
  query-efficient black-box adversarial attack via random search (2020).
\newblock \href {http://arxiv.org/abs/1912.00049} {\path{arXiv:1912.00049}}.

\bibitem{Zhong:23_1}
R.~Zhong, E.~Zhang, M.~Munetomo, Cooperative coevolutionary surrogate
  ensemble-assisted differential evolution with efficient dual differential
  grouping for large-scale expensive optimization problems, Complex \&
  Intelligent System (2023) 1--21\href
  {https://doi.org/https://doi.org/10.1007/s40747-023-01262-6}
  {\path{doi:https://doi.org/10.1007/s40747-023-01262-6}}.

\end{thebibliography}
\end{document}